\pdfoutput=1
\documentclass[11pt]{article}

\usepackage[final]{acl}

\usepackage[main=USenglish]{babel}

\useshorthands*{"}
\defineshorthand{"-}{\babelhyphen{hard}}
\defineshorthand{"=}{\babelhyphen{–}}
\defineshorthand{"/}{\babelhyphen{---}}
\defineshorthand{"@}{\babelhyphen{---}}

\newcommand*{\EMDASH}{\unskip\babelhyphen{---}}

\usepackage{times}
\usepackage{latexsym}
\usepackage[T1]{fontenc}
\usepackage[utf8]{inputenc}
\usepackage{microtype}
\usepackage{inconsolata}
\usepackage{adjustbox}
\usepackage{graphicx}
\graphicspath{{figures/}}
\usepackage{booktabs}
\usepackage{placeins}
\usepackage[morefloats=200]{morefloats}
\usepackage{amsmath}
\usepackage{amssymb}
\usepackage{multirow}
\usepackage{xcolor}
\usepackage{array}
\usepackage{enumitem}

\usepackage[table]{xcolor}
\usepackage{booktabs}
\usepackage{xfp}

\definecolor{maxc}{HTML}{08306B}
\definecolor{softwhite}{HTML}{F2EFE8}
\newcommand{\hc}[1]{%
  \cellcolor{maxc!\fpeval{round(100*#1)}!white}%
  \ifdim #1pt>0.55pt \textcolor{softwhite}{#1}\else #1\fi
}

\newcommand*{\ANC}{ANC}
\newcommand*{\CKA}{CKA}
\newcommand*{\GMM}{GMM}
\newcommand*{\GMMLAMBDA}{\ensuremath{\lambda}}

\title{A Systematic Comparison of Multilingual Interpretability Methods\\Reveals Anisotropy-Driven Failures}

\author{Oskar Holmström \quad Marcel Bollmann \quad Marco Kuhlmann \\
  Linköping University \\
  \texttt{oskar.holmstrom@liu.se}\\}

\begin{document}
\maketitle

\begin{abstract}
Multilingual language models develop shared cross"-lingual representations, and various interpretability methods claim to quantify this sharing.
These methods have been developed largely in isolation, and when they disagree, it is unclear whether the disagreement reflects a property of the model or an artifact of the measurement.
We compare four sharing metrics \EMDASH CKA, ANC, GMM dominance per token, and ILO \EMDASH across 21 base models from five families (125M"=14B parameters) and correlate each with cross"-lingual transfer on five downstream tasks.
We find that the metrics differ in their quantification of cross"-lingual sharing in these models and suggest that the disagreement traces to \emph{anisotropy}, the tendency of representations to cluster in a narrow cone of the embedding space.
Only ILO's correlation with cross"-lingual transfer (Spearman's $\rho = 0.90$) survives controls for model size, family, and per-task variation.
We therefore recommend ILO as the primary sharing metric, to be reported alongside anisotropy diagnostics.
\end{abstract}

\begin{figure}[t]
    \centering
    \includegraphics[width=\columnwidth]{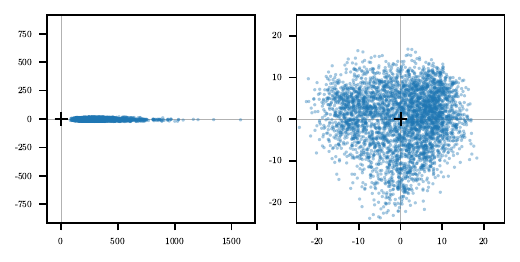}
    \caption{Sentence embeddings at layer~18 of Qwen3"-8B"-Base, projected to 2D via PCA. Left: Embeddings exhibit anisotropy in that they form a narrow cone offset from the origin. Right: Removing the top"-2 principal components yields a significantly more isotropic residual.
    We show that language identity is largely relocated to this residual (\S\ref{sec:why_disagree}), implying that sharing metrics that explicitly or implicitly focus on the principal subspace will be unreliable.}
    \label{fig:anisotropy-viz}
\end{figure}

\section{Introduction}
\label{sec:introduction}

Multilingual language models develop shared representations across languages without explicit alignment supervision \citep{pires-etal-2019-multilingual, wu-dredze-2019-beto}.
A growing set of interpretability methods now claims to measure this sharing: probing classifiers, similarity metrics, distributional approaches, and neuron-level analyses.
These methods have largely been developed in isolation, on one or two model families at a time.
When different metrics on different models yield different conclusions, it is unclear whether the disagreement reflects genuine differences between models or is an artifact of the measurement itself.

This ambiguity is especially concerning as contextual embeddings often exhibit \emph{anisotropy}: they occupy a narrow cone in the representation space, rather than spreading evenly in all directions (\emph{isotropy}), with variance concentrated in a few dimensions \citep{ethayarajh-2019-contextual, timkey-van-schijndel-2021-rogue}, as shown in Figure~\ref{fig:anisotropy-viz}.
Anisotropy poses at least two risks for any geometric measure of cross-lingual sharing:
First, it inflates apparent similarity, making any two vectors look close by default.
Second, it confines meaningful variation to a low"-dimensional subspace, so any metric sensitive to dominant directions of variance \EMDASH whether it explicitly projects onto them or implicitly up"-weights them \EMDASH will be driven by that subspace rather than by the full representational geometry.

Because conclusions about which architectures, training regimes, and scaling choices yield stronger cross"-lingual sharing are only as reliable as the metrics on which they rest, it matters whether metrics can be trusted to track sharing through anisotropy.
This paper therefore examines how this distortion affects the metrics under consideration, and assesses which, if any, can be relied on to reflect genuine cross"-lingual sharing.

Concretely, we address three questions through a systematic comparison of four sharing metrics across 21 base models from five families (BLOOM, OPT, XGLM, Qwen3, SmolLM3; 125M--14B parameters), supplemented by instruction-tuned variants and intermediate checkpoints:

\begin{itemize}[leftmargin=*, itemsep=0pt]
\item How do multilingual models share representations across languages?
\item When do existing metrics agree or disagree as to the degree of cross"-lingual sharing?
\item Which metric best predicts cross"-lingual transfer?
\end{itemize}

Our answers, anticipating \S\ref{sec:results}: models exhibit three sharing patterns rather than a binary split implied by metrics dominated by high"-variance directions, and the metric disagreement traces directly to anisotropy, with language probes showing that language identity is relocated to the directions such metrics discard or down"-weight.
Among the four metrics, ILO \citep{wilie-etal-2025} is the only one whose association with cross"-lingual transfer survives controls for size, family, and per"-task variation, reaching a Spearman rank correlation of $\rho = 0.90$; ANC \citep{del-fishel-2022} forms a clear second tier whose association rests on between"-family contrasts, while PCA"-based GMM dominance per token \citep{shani-basirat-2025} has a within"-family signal too weak to characterize as a relationship across matched architectures and CKA \citep{kornblith-etal-2019-similarity} shows no association at any level.
We therefore recommend ILO as the primary sharing metric, and further suggest that it should be reported alongside anisotropy diagnostics.

\section{Related Work}
\label{sec:related_work}

\paragraph{Measuring cross-lingual sharing}

Multilingual models develop shared representations between languages without explicit alignment supervision
\citep{pires-etal-2019-multilingual, wu-dredze-2019-beto}.
The contributing factors remain debated; shared positional embeddings, similar word order, parameter sharing across languages, and training"-data imbalance have all been proposed as drivers \citep{dufter-schutze-2020-identifying, conneau-etal-2020-emerging, schaefer-etal-2024-imbalance}.

A series of measures has been developed to quantify cross"-lingual sharing:
\emph{Similarity"-based metrics} compare activation patterns across languages on parallel input.
Centered Kernel Alignment (CKA; \citealp{kornblith-etal-2019-similarity}) measures global geometric similarity between cross"-lingual representation matrices.
Average Neuron Correlation (ANC; \citealp{del-fishel-2022}) captures per"-neuron alignment.
\emph{Lens-based methods} such as the logit lens \citep{nostalgebraist-2020-logitlens} project intermediate hidden states into the vocabulary. \citet{wendler-etal-2024-llamas} use this to argue that Llama-2 routes multilingual inputs through English"-aligned representations.
Subsequent work has questioned both the technique \citep{belrose-etal-2025-tuned-lens} and the claim itself: apparent reliance on English varies with training data balance, and representations often mix from several languages rather than being pivoted toward English \citep{zhong-etal-2024-beyond-english, schut-etal-2025-english, trinley-etal-2025-aya23, harrasse-etal-2026-transcoders}.
\citet{shani-basirat-2025} present a \emph{mixture model-based metric}: they fit Gaussian mixture models over PCA"-projected representations and use them to derive language dominance scores.
\citet{wilie-etal-2025} introduce Interlingual Local Overlap (ILO), a \emph{neighborhood"-based metric} that operates on the full hidden state.
Finally, \emph{neuron"-level analyses} identify units that activate selectively per language, using activation entropy \citep{tang-etal-2024-language-neurons, kojima-etal-2024-multilingual}, ternary classifications \citep{zhang-etal-2025-language-neurons}, sparse autoencoders \citep{deng-etal-2025-sae-features}, or transcoders \citep{harrasse-etal-2026-transcoders}.

A parallel line of work scores each language by its directed alignment with English and validates it against downstream performance, at the language level in MEXA \citep{kargaran-etal-2025-mexa} and at the instance level in DALI \citep{ravisankar-etal-2026-map}.
These primarily measure alignment with English rather than sharing among all languages, which is what the four metrics we compare quantify; we relate the two views in Appendix~\ref{sec:app_perlanguage}.
All of these methods have largely been developed and evaluated in isolation.
Comparisons typically restrict themselves to one or two model families, leaving open whether disagreements between metrics reflect substantive differences between models or measurement artifacts.
Filling this gap, we compare four representative metrics \EMDASH spanning similarity-based, distributional, and neighborhood approaches \EMDASH on the same set of models (\S\ref{sec:results}).

\paragraph{Representational anisotropy}

Contextual embeddings from transformer models tend to occupy a narrow cone in representation space rather than filling it isotropically, with most pairs of vectors having high cosine similarity regardless of their semantic content \citep{ethayarajh-2019-contextual, gao-etal-2019-degeneration}.
\citet{bis-etal-2021-too-much} argue that much of this apparent anisotropy stems from a common shift shared across embeddings: once the mean is subtracted, the residual distribution is closer to isotropic.
Other proposed causes of anisotropy include self"-attention dynamics \citep{godey-etal-2024-anisotropy} and optimizer"-level effects in Adam \citep{stollenwerk-stollenwerk-2025-coupled-adam}.
Regardless of cause, anisotropy distorts geometric analyses of representations. \citet{rajaee-pilehvar-2022-isotropy} and \citet{haemmerl-etal-2023-exploring} document anisotropy in multilingual models specifically, and show that post"-hoc isotropy enhancement improves cross"-lingual semantic similarity \EMDASH indicating
that anisotropy can mask genuine cross-lingual structure.
How isotropy should be measured is itself contested \citep{rudman-etal-2022-isoscore}, and anisotropy is not uniformly harmful: it has limited impact on sentence"-representation expressiveness for clustering \citep{ait-saada-nadif-2023-anisotropy}, is incompatible with cluster structure by construction \citep{mickus-etal-2024-isotropy}, and decreasing isotropy during finetuning can improve downstream performance \citep{rudman-eickhoff-2024-stable}.
Our concern is not whether anisotropy is desirable in a model, but what it does to the metrics used to measure cross"-lingual sharing (\S\ref{sec:why_disagree}); our own diagnosis there rests on effective dimensionality and probing rather than on the cosine diagnostic alone.

\section{Experimental Setup}
\label{sec:experimental_setup}

In our main experiments, we compute four cross"-lingual sharing metrics across 21 models.
This section presents these metrics and models along with the data we used and our statistical framework.

\subsection{Sharing Metrics}
\label{sec:methods}

We select one metric from each of the main methodological families of representation"-level cross"-lingual analysis, so that the comparison contrasts kinds of measurement rather than variants of one kind: CKA for global geometric similarity, ANC for neuron"-level correspondence, \GMMLAMBDA\ for distributional structure, and ILO for local neighborhood overlap.
The metrics are:

\begin{itemize}[leftmargin=*, itemsep=0pt]

\item \textbf{(Linear) CKA} (Centered Kernel Alignment; \citealp{kornblith-etal-2019-similarity}) is the established geometric"-similarity baseline.
It measures the similarity between two representation spaces by comparing their pairwise distance structures \EMDASH specifically, how consistently pairs of representations that are similar in one space are also similar in the other.
Values range from~0 (no shared structure) to~1 (identical up to isotropic scaling and orthogonal transformation).

\item \textbf{ANC} (Average Neuron"-Wise Correlation; \citealp{del-fishel-2022}) measures whether individual neurons carry consistent activation patterns across languages \EMDASH in contrast to CKA, which measures whether the overall pairwise similarity structure is preserved across representations.
For each neuron, ANC computes the absolute Pearson correlation between its activations on parallel sentences in two languages.
Averaging over neurons yields a value in $[0,1]$, where 1 indicates that every neuron responds identically (up to sign differences) across languages.

\item \textbf{GMM dominance per token.} \citet{shani-basirat-2025} fit a supervised Gaussian Mixture Model in a PCA subspace retaining 98\% of variance and report the fraction of tokens whose likelihood is higher in a foreign"-language space than in their own.
We refer to this quantity as $\GMMLAMBDA$ and report it by layer after subsampling each language to 10{,}000 tokens before PCA and GMM fitting.
This matches the subsampling used for ILO and removes tokenizer"-fertility bias as a confound when comparing across families with different scripts.
Unbalanced word"-level and sentence"-level variants appear in Appendix~\ref{sec:app_metrics}.

\item \textbf{ILO} (Interlingual Local Overlap; \citealp{wilie-etal-2025}) quantifies cross"-lingual sharing via $k$-nearest"-neighbor queries ($k=5$) in the full representation space, combining bridge scores (how often a token's neighbors include other languages) and reachability scores (the fraction of other languages that appear in the $k$"-nearest"-neighbor set of any token of that language) into a single index ranging from 0 (complete separation) to 1 (complete sharing).
We compute ILO on token"-level representations with the same 10{,}000-tokens-per-language subsample as \GMMLAMBDA, which controls tokenizer fertility. The released implementation of \citet{wilie-etal-2025} instead operates on mean"-pooled sentence embeddings; we report that variant in Appendix~\ref{sec:app_ilo_variant}.
We additionally report a centered variant (ILO$_c$), computed identically after subtracting the per"-layer mean hidden state across all tokens and languages, which removes anisotropy"-driven neighborhood structure that can inflate apparent sharing.

\end{itemize}

\subsection{Models}
\label{sec:models}

Our 21 base models come from five model families, spanning 125M to 14B parameters and varying in training data composition and scale.
We briefly introduce the model families here; see Appendix~\ref{sec:app_models_ckpts} for full details on models and checkpoints.

\begin{itemize}[leftmargin=*, itemsep=0pt]

\item \textbf{BLOOM} \cite{scao-etal-2022-bloom} (560M--7.1B) is trained on 350B tokens of the ROOTS corpus \cite{laurenccon-etal-2022-bigscience}.
It covers 46 natural languages with ca.\ 70\% non"-English data, and provides pretraining checkpoints.

\item \textbf{XGLM} \cite{lin-etal-2022-xglm} (564M--7.5B) is trained on 30 languages drawn from CC100-XL. We use both BLOOM and XGLM despite their age because the sharing metrics we evaluate were validated on these families: \citet{shani-basirat-2025} report \GMMLAMBDA\ on BLOOM and \citet{del-fishel-2022} report ANC on XGLM.

\item \textbf{OPT} \cite{zhang-etal-2022-opt} (125M--6.7B) serves as our non"-multilingual baseline.
It is trained on 180B predominantly English tokens from the Pile \cite{gao-etal-2020-pile}, BookCorpus, and CC-News, with only incidental non-English content from Common Crawl.

\item \textbf{Qwen3} \cite{qwen-2025-qwen3} (0.6B--14B) provides a modern high"-compute comparison.
It is trained on ca.\ 36T tokens spanning 119 languages and dialects, though exact proportions are undisclosed.

\item \textbf{SmolLM3} \cite{bakouch-2025-smollm3} (3B) is trained on 11.2T tokens drawn primarily from FineWeb"-Edu, FineWeb"-2, and DCLM, with ca.\ 12\% multilingual web data natively supporting six languages (English, French, Spanish, German, Italian, Portuguese) plus additional training on Arabic, Chinese, and Russian; it uniquely provides intermediate checkpoints across three training stages for studying multilingual representations during pre- and post"-training.

\end{itemize}

\noindent We additionally characterize instruction"-tuned variants \EMDASH BLOOMZ \cite{muennighoff-etal-2023-crosslingual}, Qwen3 non-Base, SmolLM3-Instruct \EMDASH and 59~intermediate checkpoints from BLOOM and SmolLM3, yielding a total of~32 model configurations plus checkpoints.
Instruction"-tuned variants contribute to representational characterization but are excluded from transfer analysis; all transfer correlations are based on the 21 base models.
This is because our transfer protocol finetunes on English data and evaluates zero"-shot (\S\ref{sec:transfer_eval}), measuring what the pretraining geometry makes possible; on instruction"-tuned variants it would entangle sharing with post"-training recipes that differ across families.
The question we \emph{do} ask about post"-training---whether it changes the sharing structure---is addressed in \S\ref{sec:additional}.

\subsection{Data}
\label{sec:data}

All interpretability analyses use the FLORES-200 \cite{nllb-team-2022-no} devtest set, providing 1,012 parallel sentences across ten languages: Arabic, Chinese, English, French, German, Japanese, Portuguese, Russian, Spanish, and Turkish.
These languages span four scripts (Latin, Cyrillic, Arabic, CJK), multiple typological families (Romance, Germanic, Turkic, Japonic, Sino"-Tibetan, Semitic, Slavic), and a range of morphological types (isolating, fusional, agglutinative).
Sentence"-level pairs are required by CKA and ANC, which compute cross-lingual similarity from aligned representations; the remaining methods operate on monolingual representations.

\subsection{Cross-Lingual Transfer Evaluation}
\label{sec:transfer_eval}

We use zero"-shot cross"-lingual transfer as a functional validation criterion, not as ground truth for sharing: transfer has drivers beyond representational alignment, among them task format, vocabulary overlap, and target"-language pretraining exposure, which is why we average over five tasks, require associations to survive a number of controls (\S\ref{sec:stats}), and keep the diagnosis of the disagreement between methods (\S\ref{sec:why_disagree}) independent of transfer.

We fine"-tune each of the 21 base models on English training data and evaluate zero"-shot on up to nine non"-English languages (coverage varies by task; Appendix~\ref{sec:app_xlang_datasets}) across five tasks: Belebele (reading comprehension; \citealp{bandarkar-etal-2024-belebele}), XNLI (natural language inference; \citealp{conneau-etal-2018-xnli}), XCSR (commonsense reasoning; \citealp{lin-etal-2021-xcsr}), SIB"-200 (topic classification; \citealp{adelani-etal-2024-sib200}), and XQuAD (extractive question answering; \citealp{artetxe-etal-2020-cross}).
As an aggregate per"-model metric, we compute, for each task, the mean accuracy over the non"-English target languages that task covers, and then take the unweighted mean of these five per"-task means.
We refer to this number as the \emph{transfer score}.

For each model"=task pair, we sweep over 3"=5 learning rates between $5\times10^{-6}$ and $10^{-4}$, with ranges adapted to the model family's stable regime.
We use AdamW with linear warmup (10\% of steps), weight decay 0.01, and gradient clipping at 1.0.
Task"-specific epoch counts range from 3 to 20.
The best learning rate is selected for each model"=task pair based on validation accuracy.
Full hyperparameters are in Appendix~\ref{sec:app_xlang_finetuning}.

Because the \S\ref{sec:transfer} correlations link base model metrics to finetuned model behavior, we re"-apply the full interpretability suite after finetuning to verify that the base model geometry survives.

\subsection{Statistical Framework}
\label{sec:stats}

Beyond characterizing representations, we test for functional relevance: whether sharing metrics correlate with cross"-lingual transfer, and whether any association survives controls for model size, family membership, and task choice.

\paragraph{Primary correlation}

We use Spearman rank correlations ($\rho$) for all tests.
The primary correlation is computed across all models relative to the transfer score defined in \S\ref{sec:transfer_eval}.

\paragraph{Decompositions}

We report three additional analyses to check whether the primary correlation is robust.
\emph{Per"-family} $\rho$ separates between- from within"-family contributions to the cross"-model association.
\emph{Per"-task} $\rho$ for each metric against each task (4 metrics $\times$ 5 tasks $=$ 20 tests) shows whether the primary correlation holds task by task.
Because running 20 tests simultaneously inflates the family-wise false-positive rate, we apply the \citet{bh-fdr} procedure, which bounds the expected proportion of false discoveries among reported associations at $\alpha = 0.05$. This ensures that a pattern such as a correlation being significant across all five tasks reflects a consistent relationship rather than accumulated chance.

\emph{Partial Spearman} $\rho$ controlling for $\log_{10}$ parameter size checks that the sharing"=transfer association is not reducible to a model"-size effect.
Within most families, parameter count correlates with both cross"-lingual sharing and transfer performance, raising the possibility that any observed association between sharing and transfer is a spurious consequence of larger models doing everything better rather than a genuine relationship between the two.
Partialling out $\log_{10}$ parameters removes the variance in both sharing and transfer that is attributable to size, so that the residual correlation reflects whether sharing predicts transfer over and above what model size alone would explain.

\paragraph{Permutation tests}

Given our small sample size, we verify significance with non"-parametric permutation tests. Each test recomputes $\rho$ on 10{,}000 random shuffles of the transfer score values to estimate the null distribution, from which we compute a one"-tailed $p$"-value as the proportion of recomputed $\rho$ values greater than or equal to the observed $\rho$. We run two designs: The first shuffles across all models, testing whether any monotonic association exists. The second shuffles within each family only, testing whether the association persists once between"-family contrasts are removed. Family membership is otherwise a confound: model families differ systematically in both architecture and training data, so the cross"-model correlation could be driven entirely by between"-family contrasts.

\subsection{Additional Diagnostics}

In addition to the sharing metrics, we report two diagnostics at each layer.
\emph{Effective dimensionality} ($n_\text{pca}$) is the number of PCA components needed to retain 98\% of variance; small values relative to the full number of dimensions indicate that representations occupy a low"-dimensional subspace.
\emph{Mean random cosine similarity} is the average cosine between 5{,}000 random pairs of hidden states; high values indicate the ``narrow cone'' geometry where any two vectors look similar by default \citep{ethayarajh-2019-contextual}, biasing similarity"-based measures upward.
Together, these metrics flag layers where sharing metrics should be interpreted with caution.

\section{Results and Discussion}
\label{sec:results}

This section presents our main results together with additional analyses, and discusses our findings.
We organize the section around four questions:

\subsection{How Do Models Share Representations Across Languages?}
\label{sec:how_mix}

\begin{figure*}[t]
\centering
\includegraphics[width=\textwidth]{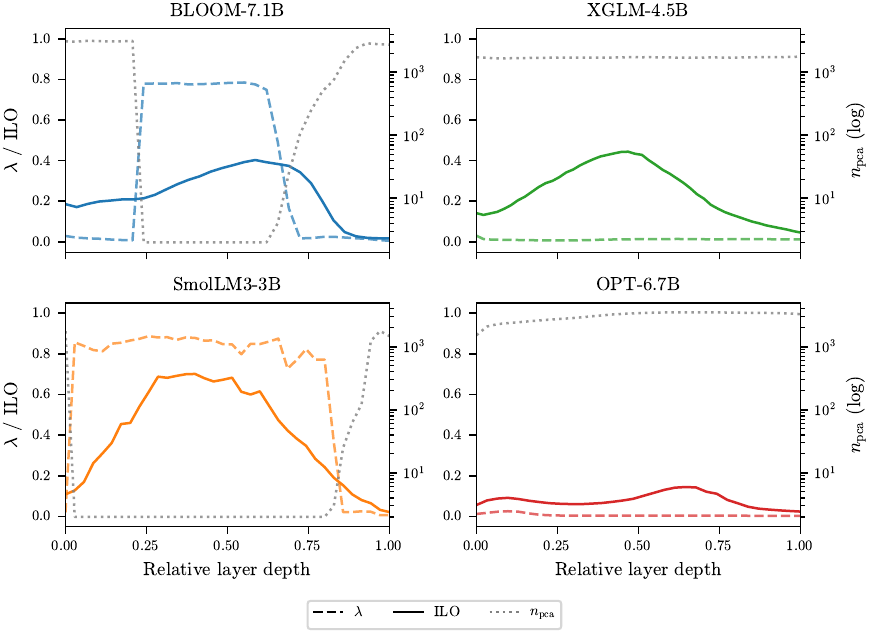}
\caption{Layer"-wise cross-lingual sharing profiles for four representative models (one per sharing pattern). Each panel shows GMM $\lambda$ (dashed) and ILO (solid) according to the scale on the left and effective PCA dimensionality $n_\text{pca}$ (dotted gray) according to the (log) scale on the right.}
\label{fig:layer_profiles}
\end{figure*}

When we apply the four metrics across our 21 base models, three distinct sharing patterns emerge
(cf.\ Table~\ref{tab:sharing_metrics}, Figure~\ref{fig:layer_profiles}).

\begin{itemize}[leftmargin=*]

\item \textbf{Pattern 1: Concentrated sharing} (BLOOM ${\geq}$1B, Qwen3, SmolLM3): high \GMMLAMBDA\ ($>0.75$), high ILO, and variance concentrated in just $n_\text{pca}=2$ dimensions at the peak \GMMLAMBDA\ layer.
The sharing metrics increase through the middle layers and fall toward the output, producing the inverted-U profile visible in Figure~\ref{fig:layer_profiles}.

\item \textbf{Pattern 2: Distributed sharing} (XGLM): high ILO ($0.44$ at 4.5B,
comparable to BLOOM-7.1B's $0.40$) but near-zero $\lambda$, with
variance distributed across thousands of dimensions ($n_\text{pca}$ in the thousands at every layer).

\item \textbf{Pattern 3: Minimal sharing} (OPT): low values on every metric,
consistent with OPT's predominantly English training data.

\end{itemize}

The critical case is XGLM.
GMM dominance per token used in isolation groups XGLM with OPT, despite XGLM's transfer score ($0.55$) approaching BLOOM-7.1B ($0.60$) and clearly exceeding OPT-6.7B ($0.41$).
The remaining metrics tell a coherent story: across all 21 models, ANC spans $0.34$--$0.68$ and tracks the sharing patterns, while CKA falls in a narrow band ($0.58$--$0.73$) and discriminates poorly across families.
Results for all models and metrics are presented in Appendix~\ref{sec:app_fullresults}.

\begin{table}[t]
\resizebox{\linewidth}{!}{%
\setlength{\tabcolsep}{3pt}%
\begin{tabular}{@{}l ccccc rc@{}}
\toprule
& \multicolumn{5}{c}{\textbf{Sharing}} & \multicolumn{2}{c}{\textbf{Anisotropy}} \\
\cmidrule(lr){2-6} \cmidrule(lr){7-8}
\textbf{Model} & $\lambda$ & \textbf{ILO} & \textbf{CKA} & \textbf{ANC} & \textbf{Xfer}
               & $n_\text{pca}$ & $\overline{\cos}$ \\
\midrule
BLOOM-560M  & \hc{.21} & \hc{.25} & \hc{.66} & \hc{.38} & \hc{.44} & 20   & .86 \\
BLOOM-7.1B  & \hc{.79} & \hc{.40} & \hc{.68} & \hc{.44} & \hc{.60} & 2    & .95 \\
OPT-6.7B    & \hc{.03} & \hc{.15} & \hc{.58} & \hc{.35} & \hc{.41} & 3477 & .99 \\
XGLM-4.5B   & \hc{.03} & \hc{.44} & \hc{.66} & \hc{.56} & \hc{.55} & 1702 & .90 \\
Qwen3-8B    & \hc{.90} & \hc{.63} & \hc{.63} & \hc{.66} & \hc{.67} & 2    & .99 \\
SmolLM3-3B  & \hc{.89} & \hc{.70} & \hc{.73} & \hc{.68} & \hc{.65} & 2    & .99 \\
\bottomrule
\end{tabular}
}
\caption{Cross-lingual sharing metrics for representative models (one per sharing pattern).
\emph{Sharing metrics:}
$\lambda$ = GMM dominance per token at peak layer;
ILO = peak interlingual local overlap;
CKA and ANC = mean across layers and language pairs;
Xfer = transfer score (\S\ref{sec:transfer_eval}).
\emph{Anisotropy diagnostics:}
$n_\text{pca}$ = PCA components retained at 98\% variance at the model's peak-sharing layer(Appendix~\ref{sec:app_anisotropy});
$\overline{\cos}$ = mean cosine similarity between random hidden states. Full sharing results are in Appendix~\ref{sec:app_fullresults} and anisotropy diagnostics are in Appendix~\ref{sec:app_anisotropy}.}
\label{tab:sharing_metrics}
\end{table}

\subsection{Why Do Methods Disagree?}
\label{sec:why_disagree}

The metric disagreement in \S\ref{sec:how_mix} is not arbitrary: each metric has a measurable sensitivity to representational anisotropy, and language probes locate where language identity is actually
encoded.

\paragraph{Anisotropy sensitivity differs across metrics}

Mean random cosine similarity ($\overline{\cos}$) at peak"-sharing layers ranges from $0.86$ (BLOOM-560M) to $0.99$ (several models), indicating that representations occupy a narrow cone where random vectors are similar by default \citep{ethayarajh-2019-contextual}.
The four metrics differ in how this geometry affects them.
\CKA\ measures absolute geometric similarity and is inflated by the shared cone direction, consistent with its narrow cross"-family range (Table~\ref{tab:sharing_metrics}).
\ANC\ computes per"-neuron Pearson correlations between parallel"-sentence activations; because Pearson subtracts each neuron's mean, \ANC\ is invariant to the common"-shift component of anisotropy \citep{bis-etal-2021-too-much}, consistent with its stronger cross"-family discrimination relative to \CKA.
ILO's cosine kNN is invariant to per"-vector scaling but sensitive to a shared mean shift.
The centered variant ILO$_c$ (\S\ref{sec:methods}) tests this directly: across all 21 models the mean $|\Delta|$ is $0.003$ (Appendix~\ref{sec:app_ilo_centering}), so this shift does not appreciably inflate the reported ILO scores.
These results are empirical, not architectural: a model with a stronger or differently"-shaped cone could show a larger centering shift, and the metric ranking could change accordingly.
The disagreement also does not reduce to the metrics' different input units: CKA and ANC compare the same mean"-pooled sentence embeddings yet correlate with transfer at $\rho = 0.14$ and $0.84$, and \GMMLAMBDA\ and ILO share the same token"-level representations yet give opposite verdicts on XGLM (Appendix~\ref{sec:app_fullresults}).
A more severe failure mode affects a different class of metrics: those that first reduce dimensionality by variance and then measure structure in whatever survives the reduction.
We examine this class next through \GMMLAMBDA.

\paragraph{Variance"-ranked projection collapses the measurement space}

The \GMMLAMBDA\ pipeline retains PCA components that cover 98\% of the variance before clustering.
This splits our 21 models into two clearly separated categories: every high"-\GMMLAMBDA\ model has $n_\text{pca}=2$ at its peak-\GMMLAMBDA\ layer, regardless of whether the full hidden dimensionality is $1{,}024$ or $5{,}120$, while every low"-\GMMLAMBDA\ model retains $\geq 20$ components.
The split is not an artifact of the threshold; it persists at 90\% retention (Appendix~\ref{sec:app_anisotropy}).
High \GMMLAMBDA\ therefore reports cluster structure in a 2"-dimensional plane while ignoring the remaining hundreds or thousands of potentially meaningful directions.
For models where $n_\text{pca}$ is larger (XGLM, OPT), no low"-dimensional clustering plane exists for the \GMM\ to fit, and \GMMLAMBDA\ stays near zero by construction.

\paragraph{Language identity is relocated, not destroyed}

To better understand the collapse, we test whether language identity persists outside the principal components: at each layer we train an $L_2$"-regularized logistic"-regression probe on the full hidden state, on the PCA principal subspace (top components covering 98\% of variance), and on its orthogonal complement (the residual subspace; setup in Appendix~\ref{sec:app_probes}).
Accuracy in the principal subspace reflects language information in the directions GMM dominance can ``see''; accuracy in the residual subspace reflects what it discards.

\begin{figure}[t]
\centering
\includegraphics[width=\columnwidth]{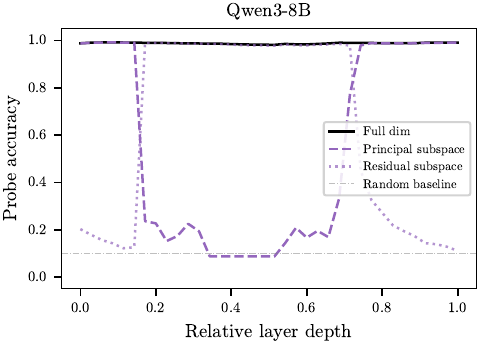}
\caption{Language probe accuracy in Qwen3-8B. The curves correspond to the full space, the principal subspace (top components covering 98\% of variance), the residual subspace, and a random baseline at $0.1$.}
\label{fig:probe_qwen3}
\end{figure}

At layers with high $\lambda$, principal"-subspace accuracy drops to chance levels,  while residual"-subspace accuracy reaches $96$--$99\%$  (Figure~\ref{fig:probe_qwen3}).
The pattern holds across BLOOM-7.1B (layers 7--19), Qwen3 (a contiguous band of 19--23 layers, depending on scale), and SmolLM3-3B (layers 1--28).
Language identity is not destroyed at sharing layers; it is relocated to the directions that a variance"-ranked projection discards.
This is problematic since the $\GMMLAMBDA$ procedure fits Gaussian components labeled by language and asks how often each token has higher likelihood under a foreign"-language component than its own, which requires language identity to be at least partially recoverable in the subspace it operates on.
Therefore, high $\GMMLAMBDA$ in this regime measures how a language"-labeled mixture model behaves in a subspace from which language has already been removed and does not measure cross"-lingual sharing in any geometric sense.
Probe results for all models are in Appendix~\ref{sec:app_probe_results}.

None of this shows that anisotropy is itself harmful to models: the two families with the strongest transfer, Qwen3 and SmolLM3, are also the most extreme on our anisotropy diagnostics ($n_\text{pca} = 2$ at their peak"-sharing layer, $\overline{\cos} = 0.99$; Table~\ref{tab:sharing_metrics}), so anisotropy is bad for certain measurements of sharing, not for transfer itself.

\subsection{Do Interpretability Methods Predict Cross-Lingual Transfer?}
\label{sec:transfer}

To see whether interpretability metrics predict cross"-lingual transfer, we correlate them with the transfer score defined in \S\ref{sec:transfer_eval}.
Per"-task and per"-language results are in Appendix~\ref{sec:app_xlang_results}, per"-language correlations in Appendix~\ref{sec:app_perlanguage}, and all statistical results are in Appendix~\ref{sec:app_regression}.

\begin{figure}[t]
\centering
\vspace*{0.08cm}
\includegraphics[width=0.95\columnwidth]{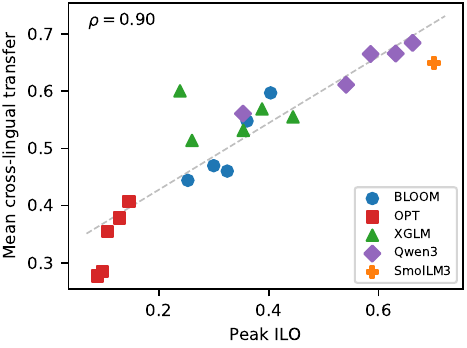}
\caption{Peak ILO predicts mean cross"-lingual transfer
across five tasks for all 21 base models. Spearman $\rho = 0.90$
($p<0.001$); partial $\rho = 0.84$ controlling for $\log_{10}$
parameters.}
\label{fig:mixing_transfer}
\end{figure}

\paragraph{ILO predicts transfer}

Figure~\ref{fig:mixing_transfer} shows the primary association: peak ILO and transfer score are monotonically related across all models.
Each family occupies a distinct range of ILO values, so the cross-model correlation reflects both between- and within"-family contributions; we disentangle these in the robustness analyses below.
Table~\ref{tab:correlations} summarizes all four metrics.
ILO is the strongest predictor ($\rho = 0.90$), ANC forms a second tier ($\rho = 0.84$), \GMMLAMBDA\ shows a moderate cross"-model association ($\rho = 0.52$), and CKA is near zero ($\rho = 0.14$).

\paragraph{The association is robust for ILO only}

We perform three checks to separate metrics that capture a sharing"-transfer relationship from those that capture confounds:
First, we control for $\log_{10}$ parameters via partial correlation; this leaves $\lambda$, ILO, and ANC largely unchanged (Table~\ref{tab:correlations}), so model size does not account for the associations. Second, we perform Benjamini"=Hochberg false discovery rate correction; this is survived by ILO and ANC on all five tasks, by $\lambda$ on two of them (Belebele and XQuAD), and by CKA on none.
Finally, we do a permutation test that shuffles transfer only within families; this confirms ILO ($p = .014$) but not CKA ($p = 0.93$), ANC ($p = 0.40$), or $\lambda$ ($p = 0.056$).
Excluding the non"-multilingual OPT baseline gives the same picture: ILO remains the strongest predictor ($\rho = 0.78$), while ANC drops ($\rho = 0.64$; per"-family and per"-task breakdowns in Appendix~\ref{sec:app_partial_spearman}).
Thus ILO is the only metric whose association survives all three controls; ANC's rests on between"-family contrasts, and CKA shows none at any level.

Note that these results are associational, not causal, as training data, architecture, and scale co"-vary across families.
However, ILO and zero"-shot transfer both depend on the local mixing of languages in representation space, so one could claim that ILO's stronger correlation partly reflects that it is the metric closest in form to our criterion.
But the same can be argued for \GMMLAMBDA, which shows the opposite result.
We therefore recommend ILO as the metric that withstands our controls and the transfer"-independent anisotropy centering check (\S\ref{sec:why_disagree}), not as a validated measure of sharing.

\paragraph{Additional checks}

The ranking in Table~\ref{tab:correlations} is unchanged when all four metrics are summarized under a uniform peak or mean layer rule (Appendix~\ref{sec:app_aggregation}); it also holds per language, where per"-language ILO correlates positively with per"-language transfer in 20 of 21 models, robust to model"-level and subword"-overlap controls (Appendix~\ref{sec:app_perlanguage}); and it is not an artifact of finetuning, which shifts the measured metrics by a median of less than 0.025 across our 105 model"=task pairs (Appendix~\ref{sec:app_finetuning_effects}).

\begin{table}[t]
\centering
\small
\setlength{\tabcolsep}{5pt}
\begin{tabular}{lcccc}
\toprule
\textbf{Metric} & $\rho$ & \textbf{Partial $\rho$}
& \textbf{Tasks} & \textbf{Within family} \\
\midrule
$\lambda$ & .52$^{*\hphantom{**}}$  & .50 & 2/5 & n.s. \\
ILO       & .90$^{***}$ & .84 & 5/5 & $p=.014$ \\
CKA       & .14$^{\hphantom{***}}$ & .29 & 0/5 & n.s. \\
ANC       & .84$^{***}$ & .84 & 5/5 & n.s. \\
\bottomrule
\end{tabular}
\caption{Spearman correlations with mean cross-lingual transfer
($n=21$). \emph{Partial $\rho$} controls for $\log_{10}$ parameters.
\emph{Tasks}: per-task correlations surviving Benjamini--Hochberg
correction at $\alpha=0.05$ over the 20-test family.
\emph{Within family}: significance under permutation tests that
shuffle transfer only within families (10{,}000 shuffles).
Significance levels: $^*p<.05$, $^{***}p<.001$.}
\label{tab:correlations}
\end{table}

\subsection{How Does Sharing Emerge During and After Pretraining?}
\label{sec:additional}

\paragraph{Sharing emerges through abrupt phase transitions}
Using intermediate checkpoints from BLOOM (30 across 5 sizes) and SmolLM3 (29 across 3 stages), $\lambda$ and ILO undergo coincident jumps at specific training steps rather than rising gradually. In BLOOM-1b1, both metrics jump in lockstep between steps 10K and 100K
($\lambda$: $0.13 \to 0.79$); BLOOM-560M never makes this transition (peak $\lambda = 0.26$). The capacity threshold for BLOOM sits between 560M and 1.1B, but it is architecture- and not size-specific, since the similarly sized Qwen3-0.6B exhibits strong sharing ($\lambda = 0.92$). SmolLM3 establishes sharing at its earliest checkpoint ($\lambda = 0.81$ at step 40K) and then the late Stage~3 reasoning training expands the high-$\lambda$ zone from 22 to 30 layers. Per-checkpoint results are in Appendix~\ref{sec:app_training_dynamics}.

\paragraph{Instruction tuning can induce sharing}
\label{sec:instruction_tuning}

Because our metrics describe each model at a single training snapshot, instruction"-tuned variants, a second snapshot of the same model, let us test whether the three sharing patterns of \S\ref{sec:how_mix} are stable model properties and whether post"-training itself reshapes sharing.
BLOOM-560M, for which our metrics never indicate multilingual sharing, develops a 10-layer collapsed"-dimensionality zone after instruction tuning (BLOOMZ-560m), where peak $\lambda$ rises from $0.21$ to
$0.84$ and peak ILO from $0.25$ to $0.37$, with $n_\text{pca}=2$ at the new peak layer. The effect then attenuates at scale, where BLOOMZ-7.1B and Qwen3-8B-Instruct show only modest shifts. Full before/after metric results are in Appendix~\ref{sec:app_instruction_effects}. They suggest that sharing is not fixed at the end of pretraining: it can emerge through instruction tuning in small models and reorganize through midtraining in larger ones.

\section{Conclusion}
\label{sec:conclusion}

Cross"-lingual sharing metrics have been developed and evaluated largely in isolation, leaving open whether disagreements between them reflect genuine differences between models or artifacts of measurement. Our systematic comparison across 21 base models from five families shows that the latter dominates: representational anisotropy makes GMM dominance per token, reliant on PCA, report sharing where probes show language identity has merely been relocated to discarded dimensions, and miss sharing where it is distributed across thousands of directions. This conflates three distinct sharing patterns"/concentrated, distributed, and minimal"/by collapsing the latter two into a single near-zero score. Among the four metrics tested, only ILO's association with cross-lingual transfer survives controls for model size, family membership, and per-task decomposition.

The practical implication for interpretability practice is that geometric sharing scores cannot be read independently of the geometry they sit in. We recommend reporting metrics alongside anisotropy diagnostics, such as effective dimensionality and mean random cosine similarity at a minimum, so that a reader can tell whether a high score reflects cross-lingual sharing or anisotropic collapse.
These results are corrective rather than dismissive of global sharing metrics: a single global score remains the right instrument for quantifying and comparing sharing across models, while explaining how sharing is implemented calls for mechanistic methods \citep{tang-etal-2024-language-neurons, kojima-etal-2024-multilingual, deng-etal-2025-sae-features, harrasse-etal-2026-transcoders}, which, insofar as they too read structure off anisotropic geometry, likewise benefit from anisotropy diagnostics.
Ultimately, these metrics are instruments for understanding models and guiding the development of new models that improve sharing of representations across languages, a role they can play only when they measure what we mean to track.

\paragraph{Future work}

Concentrated sharing always coincides with low $n_\text{pca}$ at the
peak layer, but our observational design cannot tell whether this bottleneck actively forces language-agnostic computation or is incidental. Optimizers that suppress anisotropy, such as Coupled
Adam \citep{stollenwerk-stollenwerk-2025-coupled-adam}, offer a clean intervention to test both this question and whether our metric reliability ranking holds under different anisotropy profiles.
Controlled training with matched architectures and varied data composition would further isolate the factors driving the three sharing patterns.
Future work also includes applying purpose"-built isotropy measures such as IsoScore \citep{rudman-etal-2022-isoscore} and extending the comparison to multilingual encoders.

\section*{Limitations}
\label{sec:limitations}

\paragraph{Small and non-independent sample}

Our 21 models from five families share architecture, data, and tokenizer within each family, making the effective between"-family sample size closer to five.
We address this through within"-family analyses, partial correlations, and within"-family permutation tests (\S\ref{sec:transfer}), but statistical power remains limited.
SmolLM3 is represented by a single model size, and detailed per"-language training data proportions are publicly available only for BLOOM, preventing direct analysis of how data composition relates to sharing patterns.

\paragraph{Generalization of metric reliability}

The reliability ranking we report ($\lambda$ heavily anisotropy"-sensitive, CKA weakly discriminative, ILO and ANC more robust) holds across the 21 models in this study.
Models with substantially higher anisotropy, or with anisotropy modes we did not encounter, may shift this ranking.
The takeaway is therefore not that ILO and ANC are universally safe but that anisotropy diagnostics should be reported alongside any geometric sharing metric.

\paragraph{Decoder"-only models}

All 21 models we study are decoder"-only language models, whereas much of the cross"-lingual sharing literature was built on multilingual encoders \citep{pires-etal-2019-multilingual, conneau-etal-2020-unsupervised}, which differ in training objective and attention pattern and in which anisotropy has been documented independently \citep{rajaee-pilehvar-2022-isotropy}.
Whether the reliability ranking we report transfers to encoders is untested, and we make no claim about them.

\paragraph{Language selection and coverage}

The ten evaluation languages are those reasonably covered by all five transfer benchmarks and by all five model families, and the restriction is deliberate: admitting languages that only some benchmarks or families support would make each model's transfer score depend on which languages happened to be available for it, and the cross"-model comparison would no longer be like"-for"-like.
Even within this set coverage is not complete (XCSR omits Turkish, XNLI omits Japanese and Portuguese, and XQuAD omits Japanese, French, and Portuguese; Appendix~\ref{sec:app_xlang_results}), which is one reason we average over the languages a task does cover.
Two ramifications follow.
First, the ten languages are comparatively high"-resource and three are Romance, so per"-language patterns may be biased toward Romance, and our conclusions concern sharing among well"-represented languages.
Second, we cannot say from these data whether ILO remains the most reliable metric, or whether \GMMLAMBDA\ fails in the same way, for languages thinly represented in pretraining, the regime where a practitioner would most want a trustworthy sharing metric; testing this requires benchmarks with wider coverage and is left to future work.

\paragraph{Observational design}

All reported correlations are associational.
Common causes such as training data diversity, which promotes both sharing and transfer simultaneously, remain plausible.
Controlled interventions would be needed to isolate causal mechanisms.

\paragraph{Hyperparameter sensitivity and confounds}

We follow \citet{shani-basirat-2025} for the 98\% PCA variance threshold and \citet{wilie-etal-2025} for ILO parameters ($k=5$, $\tau=3$) without systematic sensitivity analysis.
ILO is also the most computationally expensive of the four metrics; asymptotic time and memory costs for all four are given in Appendix~\ref{sec:app_cost}.
All sharing metrics show some correlation with tokenizer vocabulary overlap ($\rho \approx 0.33$--$0.41$).
Future work should explore vocabulary"-controlled metrics or language pairs matched on tokenizer overlap.

\section*{Acknowledgments}

We thank the anonymous reviewers for their constructive feedback and insightful suggestions.
This research was supported by TrustLLM funded by Horizon Europe GA 101135671.
We also acknowledge support from the Graduate School in Computer Science (CUGS).
Computational resources were provided on the Berzelius system funded by the Knut and Alice Wallenberg foundation and operated by NAISS. 

\bibliography{custom}

\appendix

\section{Method Details}
\label{sec:app_methods}

\subsection{Models and Checkpoints}
\label{sec:app_models_ckpts}

Table~\ref{tab:app_models} lists every base model and instruction-tuned variant used in the analyses reported in the main paper.
Architectural details are taken from the corresponding HuggingFace model cards.
The 21 base models constitute the sample for all transfer correlations
(\S\ref{sec:transfer}); the instruction-tuned variants and intermediate
checkpoints are used for representational characterization only
(\S\ref{sec:additional}).

\begin{table}[!b]
\centering
\small
\setlength{\tabcolsep}{6pt}
\begin{tabular}{llrr}
\toprule
\textbf{Family} & \textbf{Model} & \textbf{Layers} & \textbf{$d$} \\
\midrule
\multirow{5}{*}{BLOOM}
  & BLOOM-560M  & 24 & 1024 \\
  & BLOOM-1b1   & 24 & 1536 \\
  & BLOOM-1b7   & 24 & 2048 \\
  & BLOOM-3B    & 30 & 2560 \\
  & BLOOM-7b1   & 30 & 4096 \\
\midrule
\multirow{5}{*}{BLOOMZ}
  & bloomz-560m & 24 & 1024 \\
  & bloomz-1b1  & 24 & 1536 \\
  & bloomz-1b7  & 24 & 2048 \\
  & bloomz-3b   & 30 & 2560 \\
  & bloomz-7b1  & 30 & 4096 \\
\midrule
\multirow{5}{*}{OPT}
  & OPT-125M  & 12 & 768  \\
  & OPT-350M  & 24 & 1024 \\
  & OPT-1.3B  & 24 & 2048 \\
  & OPT-2.7B  & 32 & 2560 \\
  & OPT-6.7B  & 32 & 4096 \\
\midrule
\multirow{5}{*}{XGLM}
  & XGLM-564M  & 24 & 1024 \\
  & XGLM-1.7B  & 24 & 2048 \\
  & XGLM-2.9B  & 48 & 2048 \\
  & XGLM-4.5B  & 48 & 2048 \\
  & XGLM-7.5B  & 32 & 4096 \\
\midrule
\multirow{5}{*}{Qwen3-Base}
  & Qwen3-0.6B-Base & 28 & 1024 \\
  & Qwen3-1.7B-Base & 28 & 2048 \\
  & Qwen3-4B-Base   & 36 & 2560 \\
  & Qwen3-8B-Base   & 36 & 4096 \\
  & Qwen3-14B-Base  & 40 & 5120 \\
\midrule
\multirow{5}{*}{Qwen3-Instruct}
  & Qwen3-0.6B & 28 & 1024 \\
  & Qwen3-1.7B & 28 & 2048 \\
  & Qwen3-4B   & 36 & 2560 \\
  & Qwen3-8B   & 36 & 4096 \\
  & Qwen3-14B  & 40 & 5120 \\
\midrule
\multirow{2}{*}{SmolLM3}
  & SmolLM3-3B-Base     & 36 & 2048 \\
  & SmolLM3-3B          & 36 & 2048 \\
\bottomrule
\end{tabular}
\caption{\textbf{Models used in the study.} Layer depth and hidden size
$d$ are taken from the published model cards; parameter counts are
indicated by the model names.}
\label{tab:app_models}
\end{table}

Table~\ref{tab:app_checkpoints} enumerates the intermediate checkpoints used in the training-dynamics analysis. For BLOOM, 10 of the 40 checkpoints
are byte"-identical and are excluded as apparent duplicates; both members of each pair are dropped because we cannot determine which step is correct. SmolLM3-3B-Base checkpoints are partitioned into the three training stages documented in the model release.

\begin{table}[!t]
\centering
\small
\setlength{\tabcolsep}{5pt}
\begin{tabular}{lp{0.62\columnwidth}}
\toprule
\textbf{Model / Stage} & \textbf{Steps used} \\
\midrule
\multicolumn{2}{l}{\textbf{BLOOM}} \\
BLOOM-560M (6/8) & 1K, 100K, 200K, 300K, 400K, 600K \\
BLOOM-1b1 (8/8)  & 1K, 10K, 100K, 200K, 300K, 400K, 500K, 600K \\
BLOOM-1b7 (4/8)  & 50K, 100K, 150K, 200K \\
BLOOM-3B  (6/8)  & 1K, 10K, 50K, 100K, 150K, 300K \\
BLOOM-7b1 (6/8)  & 1K, 10K, 50K, 150K, 250K, 300K \\
\midrule
\multicolumn{2}{l}{\textbf{SmolLM3-3B-Base}} \\
Stage 1 (20) & 40K, 200K, 400K, 560K, 760K, 920K, 1120K, 1280K, 1480K, 1640K, 1840K, 2000K, 2200K, 2360K, 2560K, 2720K, 2920K, 3080K, 3280K, 3440K \\
Stage 2 (5)  & 3480K, 3600K, 3800K, 4000K, 4200K \\
Stage 3 (4)  & 4240K, 4400K, 4560K, 4720K \\
\bottomrule
\end{tabular}
\caption{\textbf{Intermediate checkpoints used in the training-dynamics
analysis.}  The excluded pairs are
(BLOOM-560M: 10K, 500K),
(BLOOM-1b7: 1K, 10K),
(BLOOM-1b7: 250K, 300K),
(BLOOM-3B: 200K, 250K), and
(BLOOM-7b1: 100K, 200K). SmolLM3 stage boundaries follow the model
release: Stage~1 $\to$ Stage~2 at $\sim$3.46M steps, Stage~2 $\to$ Stage~3
at $\sim$4.22M steps.}
\label{tab:app_checkpoints}
\end{table}

\FloatBarrier
\subsection{Metrics}
\label{sec:app_metrics}

Let $h_\ell^{(i)} \in \mathbb{R}^d$ denote the hidden state at layer $\ell$ for
token (or sentence) $i$, and let $\ell(i) \in \mathcal{L}$ be the language label
of $i$ over a fixed set of $L = |\mathcal{L}|$ languages. All metrics are
computed independently per layer $\ell$ on the FLORES-200 \texttt{devtest} split.

\paragraph{GMM dominance per token ($\lambda$).}
Our metric builds on the GMM construction of \citet{shani-basirat-2025}. At each layer, embeddings $H_\ell \in \mathbb{R}^{N \times d}$ are projected by PCA onto the smallest number of components $n_\text{pca}(\ell)$ retaining $\geq 98\%$ of the variance, with a floor of $n_\text{pca}(\ell) \geq 2$ (matching the original implementation). A supervised Gaussian mixture is then fit with one full-covariance component per language; mean and covariance are estimated from per-language sample statistics rather than via expectation--maximization, and mixing weights are set proportional to per-language token counts:
\begin{align}
\hat{\mu}_k &= \frac{1}{|S_k|}\sum_{i \in S_k} z_\ell^{(i)}, \\
\hat{\Sigma}_k &= \frac{1}{|S_k| - 1}\sum_{i \in S_k}(z_\ell^{(i)} - \hat{\mu}_k)(z_\ell^{(i)} - \hat{\mu}_k)^\top + \epsilon I, \\
\hat{\pi}_k &= |S_k| / N,
\end{align}
where $z_\ell^{(i)} = \mathrm{PCA}_\ell(h_\ell^{(i)})$ and $S_k = \{i : \ell(i) = k\}$. The ridge $\epsilon = 10^{-4}$ is our addition, ensuring positive-definiteness for near-singular per-language covariances.

Given the fitted mixture, every hidden state $h$ has a posterior $P(\ell \mid h)$ under each language Gaussian. For a token $i$ from source language $s(i)$, we form the source-against-best-competitor ratio
\begin{equation}
\Lambda_{\ell}(i) \;=\; \frac{P\!\bigl(s(i) \mid h_\ell^{(i)}\bigr)}{\max_{t \neq s(i)} P\!\bigl(t \mid h_\ell^{(i)}\bigr)},
\end{equation}
which is a token-level instance of the pairwise likelihood ratio of \citet{shani-basirat-2025} (their Eq.~4), aggregated over foreign languages by retaining only the most competitive one. The layer-wise dominance fraction is the proportion of tokens for which this ratio falls below one,
\begin{equation}
\lambda(\ell) \;=\; \frac{1}{N}\,\bigl|\{i : \Lambda_{\ell}(i) < 1\}\bigr|,
\end{equation}
computed jointly across all source languages. The per-model summary reported in Table~\ref{tab:sharing_metrics} is the peak-layer value $\lambda^\star = \max_\ell \lambda(\ell)$.

\paragraph{Relation to the per-language dominance score.} Our $\lambda$ corresponds to the per-token analysis of \citet[\S5.2]{shani-basirat-2025} and is distinct from the per-language \emph{dominance score} $\hat{D}_i(T) = \mathbb{E}_{S \neq T}\,\mathbb{E}_{h \sim S}[P(T \mid h)]$ they develop in their \S5.1 to test for a single mediating language. The dominance score is a scalar per target language. $\lambda$ instead asks, for every token, whether the source-language Gaussian retains the largest posterior, and reports the population fraction for which it does not. The two metrics answer complementary questions and need not move together: $\hat{D}_i(T)$ tests whether one language acts as a representational magnet for the others, whereas $\lambda$ measures the extent to which per-language regions of the embedding space remain linearly separable, irrespective of whether any single language dominates the rest.

We compute three variants of $\lambda$ that differ only in what counts as a token:
\begin{itemize}
\item \textbf{Word-level} $\lambda$: tokens are kept at their natural per-language frequencies.
\item \textbf{Sample-balanced} $\lambda$ (default reported in the main paper): each language is subsampled to $10{,}000$ tokens before PCA fitting and GMM evaluation, matching the ILO subsampling protocol and removing tokenizer-fertility bias.
\item \textbf{Sentence-level} $\lambda$: $h_\ell^{(i)}$ is replaced by the mean of token hidden states over sentence $i$, giving one embedding per sentence.
\end{itemize}
PCA hyperparameters are identical across the three variants.

\paragraph{Effective dimensionality and mean random cosine similarity.}
At each layer we report two anisotropy diagnostics computed on the same pooled
hidden states used by $\lambda$:
\begin{align}
n_\text{pca}(\ell) &= \min\Bigl\{k \,:\, \tfrac{\sum_{j \leq k}\sigma_j^2(\ell)}{\sum_j \sigma_j^2(\ell)} \geq 0.98\Bigr\}, \\
\overline{\cos}(\ell) &= \frac{1}{M}\sum_{m=1}^{M} \cos\bigl(h_\ell^{(i_m)},\, h_\ell^{(j_m)}\bigr),
\end{align}
where $\{\sigma_j(\ell)\}$ are the singular values of the centered hidden-state
matrix and $\{(i_m, j_m)\}_{m=1}^{M}$ are $M = 5{,}000$ pairs of distinct
indices sampled uniformly at random with a fixed seed.

\paragraph{Interlingual Local Overlap (ILO).}
We follow the implementation of \citet{wilie-etal-2025}. At layer $\ell$, all
hidden states across languages are pooled into one matrix $H_\ell$, and for
every point $i$ we retrieve its $k = 5$ nearest neighbors under cosine
similarity, $\mathcal{N}_k(i)$, excluding self-similarity. Let $L_k(i) =
\{\ell(j) : j \in \mathcal{N}_k(i)\} \setminus \{\ell(i)\}$ be the set of
\emph{other} languages appearing among $i$'s neighbors. The bridge score for
language $a$ is
\begin{equation}
\mathrm{Bridge}_a(\ell) \;=\; \frac{|\{i \in S_a : |L_k(i)| \geq \tau\}|}{|S_a|},
\end{equation}
with threshold $\tau = 3$. The reachability score counts the distinct other
languages appearing in any neighbor set of any point of $a$ and is normalized
by $L - 1$:
\begin{equation}
\mathrm{Reach}_a(\ell) \;=\; \frac{1}{L-1}\,\Bigl|\bigcup_{i \in S_a} L_k(i)\Bigr|.
\end{equation}
The per-language ILO score is the harmonic mean of the two,
\begin{equation}
\mathrm{ILO}_a(\ell) \;=\; \frac{2\,\mathrm{Bridge}_a(\ell)\,\mathrm{Reach}_a(\ell)}{\mathrm{Bridge}_a(\ell) + \mathrm{Reach}_a(\ell)},
\end{equation}
(set to $0$ if either term is $0$), and the layer score is the mean over
languages, $\mathrm{ILO}(\ell) = L^{-1}\sum_a \mathrm{ILO}_a(\ell)$.
All ILO results in the paper use this word-level computation, subsampled to $10{,}000$ tokens per language to equalize representation in the kNN graph (using a fixed seed); the sentence-level variant of the released implementation is reported in Appendix~\ref{sec:app_ilo_variant}. We use $k = 5$, $\tau = 3$, and cosine similarity throughout, matching the defaults of the released implementation of \citet{wilie-etal-2025} (described there as the best settings).

\paragraph{Centered ILO ($\mathrm{ILO}_c$).}
We define $\mathrm{ILO}_c$ identically to $\mathrm{ILO}$ but on
mean-subtracted hidden states. Concretely, at each layer $\ell$ we compute the
sample-weighted grand mean over the pooled (post-subsampling) embeddings,
\begin{equation}
\bar{h}_\ell \;=\; \frac{1}{N}\sum_{i=1}^{N} h_\ell^{(i)},
\end{equation}
and replace every $h_\ell^{(i)}$ with $h_\ell^{(i)} - \bar{h}_\ell$ before
running kNN. Subtracting the same vector from every language preserves
cross-lingual offsets but removes the shared anisotropic ``common direction''
\citep{ethayarajh-2019-contextual}.

\paragraph{Centered Kernel Alignment (CKA).}
For a language pair $(a, b)$ and layer $\ell$, let $X \in \mathbb{R}^{n \times
d}$ and $Y \in \mathbb{R}^{n \times d}$ be the matrices of mean-pooled
sentence embeddings for the $n$ FLORES sentence pairs. Linear CKA
\citep{kornblith-etal-2019-similarity} is
\begin{equation}
\mathrm{CKA}(X, Y) \;=\; \frac{\|Y^\top X\|_F^2}{\|X^\top X\|_F\,\|Y^\top Y\|_F},
\end{equation}
computed after centering each column of $X$ and $Y$. We aggregate over the
$\binom{L}{2}$ language pairs and over layers by simple averaging; the
per-model summary in Table~\ref{tab:sharing_metrics} is the layer-mean of the
language-pair mean.

\paragraph{Average Neuron Correlation (ANC).}
For the same $X, Y$, ANC \citep{del-fishel-2022} measures alignment at the level of individual neurons:
\begin{equation}
\mathrm{ANC}(X, Y) \;=\; \frac{1}{d}\sum_{j=1}^{d} \bigl| \rho\bigl(X_{:,j}, Y_{:,j}\bigr)\bigr|,
\end{equation}
where $\rho$ is the Pearson correlation across the $n$ aligned sentence pairs and the absolute value accounts for sign-flips that are immaterial to neuron
function. We aggregate over language pairs and layers as for CKA.

\FloatBarrier
\subsection{Computational Cost}
\label{sec:app_cost}

We state asymptotic per-layer costs in the notation of \S\ref{sec:app_metrics} ($d$: hidden size; $n = 1{,}012$ parallel sentences for the sentence-level metrics; $N = 10^{5}$ pooled tokens for the token-level metrics).
ANC computes $d$ Pearson correlations over $n$ aligned samples per language pair: $O(nd)$ time.
CKA additionally requires a matrix product: $O(nd \cdot \min(n, d))$ per pair.
$\lambda$ is dominated by the PCA of the $N \times d$ matrix of pooled hidden states: $O(Nd^2)$ for $N > d$.
ILO retrieves exact nearest neighbors over all $N$ pooled points, an all-pairs computation of $O(N^2 d)$; since $N$ is two orders of magnitude larger than $n$, ILO is the most expensive of the four.
Memory is dominated by the pooled hidden states, $O(Nd)$, about $2$~GB per layer at single precision for $N = 10^{5}$ and $d = 5{,}120$.

\FloatBarrier
\subsection{Language Probe Setup}
\label{sec:app_probes}

The language probe is a multinomial logistic
regression over $L = 10$ FLORES languages, trained per layer in three feature
spaces:
\begin{itemize}
\item \textbf{Full}: the raw hidden state $h_\ell \in \mathbb{R}^d$.
\item \textbf{PCA principal}: projection onto the top $n_\text{pca}(\ell)$
  PCA components retaining $\geq 98\%$ of variance (the same components $\lambda$
  operates on, with the same floor of $2$).
\item \textbf{PCA residual}: projection onto the orthogonal complement, the
  remaining $d - n_\text{pca}(\ell)$ components.
\end{itemize}

\paragraph{Hyperparameters.}
We use a stratified 80/20 split over the $L \times 1{,}012$ FLORES-200 sentences. The classifier is multinomial
logistic regression with $L_2$ regularization, trained by L-BFGS with strong Wolfe line search, $\text{tol\_grad} = 10^{-5}$, $\text{tol\_change} = 10^{-9}$,
and at most $1{,}000$ iterations. Features are not standardized; using raw PCA scores keeps the three subspaces directly comparable in scale (the PCA principal subspace already contains the high-variance directions). We report test-set accuracy.

\FloatBarrier

\subsection{Cross-lingual Finetuning Datasets}
\label{sec:app_xlang_datasets}

We use five established benchmarks for the transfer evaluation
(\S\ref{sec:transfer}); per-language coverage and split sizes are given in
Table~\ref{tab:app_xfer_data}.
\begin{itemize}
\item \textbf{Belebele} \citep{bandarkar-etal-2024-belebele}: multiple-choice
  reading comprehension, 4-way. We train on the
  \texttt{belebele\_training\_set} (English, $\approx$71\,K examples) and
  evaluate on the test set in 10 languages.
\item \textbf{XNLI} \citep{conneau-etal-2018-xnli}: 3-way natural language
  inference, parallel across 15 languages. We train on the English
  \texttt{train} split ($\approx$392\,K examples) and evaluate on the
  \texttt{test} split for the 8 of our 10 languages present in XNLI (Japanese
  and Portuguese excluded).
\item \textbf{XCSR} \citep{lin-etal-2021-xcsr}: 5-way multilingual
  commonsense reasoning. We train on the English X-CSQA \texttt{train} split
  ($\approx$9.7\,K examples) and evaluate on the \texttt{validation} split for
  the 9 covered target languages.
\item \textbf{SIB-200} \citep{adelani-etal-2024-sib200}: 7-way topic
  classification on FLORES-200 sentences. We train on the per-language
  \texttt{train} split (701 examples) and evaluate on the \texttt{test} split
  in all 10 languages.
\item \textbf{XQuAD} \citep{artetxe-etal-2020-cross}: extractive QA. Following
  standard practice, we train on the English SQuAD~v1.1 \texttt{train} split
  ($\approx$87.6\,K examples) and evaluate F1 on the XQuAD test sets for the
  7 covered target languages.
\end{itemize}
The transfer score for each model is the unweighted mean of the five per-task means, each computed over that task's non-English target languages. Per-task language counts and overall coverage are summarized in Table~\ref{tab:app_xfer_data}.

\begin{table}[!htbp]
\centering
\small
\setlength{\tabcolsep}{4pt}
\begin{tabular}{lllr}
\toprule
\textbf{Task} & \textbf{Train} & \textbf{Eval split} & \textbf{Eval langs} \\
\midrule
Belebele  & EN ($\approx$71\,K)    & test       & 10 \\
XNLI      & EN ($\approx$392\,K)   & test       & 8  \\
XCSR      & EN ($\approx$9.7\,K)   & validation & 9  \\
SIB-200   & EN (701)               & test       & 10 \\
XQuAD     & EN-SQuAD ($\approx$88\,K) & test    & 7  \\
\bottomrule
\end{tabular}
\caption{Cross-lingual transfer benchmarks. Training is always English-only
(``EN''). Eval-language counts cover only those languages in our 10-language
set that are also present in each benchmark.}
\label{tab:app_xfer_data}
\end{table}

\FloatBarrier

\subsection{Cross-lingual Finetuning Setup}
\label{sec:app_xlang_finetuning}

\paragraph{Optimizer and schedule.}
For every (model, task) pair we finetune with AdamW (PyTorch defaults
$\beta_1 = 0.9$, $\beta_2 = 0.999$, $\epsilon = 10^{-8}$), weight decay
$0.01$, gradient clipping at $\ell_2$ norm $1.0$, linear warmup over the
first $10\%$ of training steps followed by linear decay to zero, and maximum
sequence length $512$. Models are trained in bfloat16 mixed precision on
A100/H200 GPUs. The classification
head is a fresh task-specific linear layer on top of the last-layer hidden
state of the final non-padding token; QA training uses standard span prediction heads. All other model parameters are unfrozen.

\paragraph{Per-task budget.}
Effective batch size and epoch count per task are: XNLI, XCSR, XQuAD
(eff.\ batch $32$, 3 epochs for the larger tasks; 20 epochs for XCSR);
SIB-200 (eff.\ batch $32$, 20 epochs); Belebele (eff.\ batch $16$, 3
epochs). For tasks with a validation split (XNLI, XCSR, SIB-200) we select the epoch with the lowest validation cross-entropy and use that checkpoint for downstream evaluation; for tasks without a validation split (Belebele, XQuAD) we use the final-epoch model.

\paragraph{Learning-rate sweep.}
For each (model, task) pair we sweep over 3--5 learning rates drawn from
$\{2\!\times\!10^{-6},\, 5\!\times\!10^{-6},\, 10^{-5},\, 2\!\times\!10^{-5},\,
5\!\times\!10^{-5},\, 10^{-4},\, 2\!\times\!10^{-4},\, 5\!\times\!10^{-4},\,
10^{-3},\, 2\!\times\!10^{-3}\}$, with the range adapted to each family's
stable regime (BLOOM, OPT and Qwen3 stable at $10^{-5}$--$10^{-4}$; XGLM
stable at $5\!\times\!10^{-5}$--$10^{-3}$). When the best LR sat at a sweep
boundary we extended the range until either an interior optimum was found or
the boundary remained stable.

\paragraph{Best-LR selection.}
For each (model, task) pair we select the learning rate that maximizes mean
target-language accuracy (or F1, for XQuAD) averaged across all evaluation
languages of that task. The selected LR is then used for the corresponding
representational (interpretability) re-analysis on the finetuned model.
Selected values per (model, task) are listed in
Table~\ref{tab:app_selected_lr}.

\begin{table*}[!htbp]
\centering
\small
\setlength{\tabcolsep}{6pt}
\begin{tabular}{llrrrrr}
\toprule
\textbf{Family} & \textbf{Model} & \textbf{Belebele} & \textbf{XNLI} & \textbf{XCSR} & \textbf{SIB-200} & \textbf{XQuAD} \\
\midrule
\multirow{5}{*}{BLOOM}
  & BLOOM-560M  & 2e-5 & 5e-5 & 5e-5 & 2e-4 & 5e-5 \\
  & BLOOM-1b1   & 2e-5 & 5e-5 & 5e-5 & 2e-5 & 2e-5 \\
  & BLOOM-1b7   & 2e-5 & 1e-4 & 5e-5 & 2e-5 & 2e-5 \\
  & BLOOM-3B    & 2e-5 & 5e-5 & 5e-5 & 2e-5 & 1e-5 \\
  & BLOOM-7b1   & 2e-5 & 5e-5 & 5e-5 & 5e-5 & 2e-5 \\
\midrule
\multirow{5}{*}{OPT}
  & OPT-125M  & 1e-5 & 5e-5 & 2e-4 & 1e-4 & 2e-5 \\
  & OPT-350M  & 5e-5 & 1e-4 & 5e-5 & 5e-5 & 1e-4 \\
  & OPT-1.3B  & 2e-5 & 5e-5 & 2e-5 & 2e-5 & 5e-5 \\
  & OPT-2.7B  & 1e-5 & 5e-5 & 2e-5 & 2e-5 & 5e-5 \\
  & OPT-6.7B  & 1e-5 & 2e-5 & 2e-5 & 1e-5 & 2e-5 \\
\midrule
\multirow{5}{*}{XGLM}
  & XGLM-564M  & 1e-4 & 5e-4 & 5e-4 & 1e-3 & 2e-4 \\
  & XGLM-1.7B  & 2e-4 & 5e-4 & 2e-4 & 2e-4 & 2e-4 \\
  & XGLM-2.9B  & 1e-4 & 2e-4 & 1e-4 & 1e-4 & 1e-4 \\
  & XGLM-4.5B  & 2e-5 & 5e-5 & 5e-5 & 5e-5 & 5e-5 \\
  & XGLM-7.5B  & 5e-5 & 1e-4 & 5e-5 & 1e-4 & 5e-5 \\
\midrule
\multirow{5}{*}{Qwen3-Base}
  & Qwen3-0.6B-Base & 5e-5 & 5e-5 & 5e-5 & 5e-5 & 1e-5 \\
  & Qwen3-1.7B-Base & 5e-5 & 5e-5 & 1e-4 & 5e-5 & 1e-5 \\
  & Qwen3-4B-Base   & 2e-5 & 5e-5 & 5e-5 & 2e-5 & 2e-6 \\
  & Qwen3-8B-Base   & 2e-5 & 5e-5 & 5e-5 & 2e-4 & 1e-5 \\
  & Qwen3-14B-Base  & 2e-5 & 5e-5 & 5e-5 & 2e-5 & 1e-5 \\
\midrule
SmolLM3 & SmolLM3-3B-Base & 2e-5 & 5e-5 & 5e-5 & 1e-5 & 5e-6 \\
\bottomrule
\end{tabular}
\caption{\textbf{Selected learning rate per (model, task) pair} for the
21 base models used in the cross-lingual transfer analysis. For each pair
we sweep over 3--5 learning rates and select the one that maximises mean
target-language accuracy (F1 for XQuAD).}
\label{tab:app_selected_lr}
\end{table*}

\FloatBarrier

\section{Results}

\subsection{Full Metric Results}
\label{sec:app_fullresults}

\begin{figure*}[!htbp]
\centering
\includegraphics[width=\textwidth]{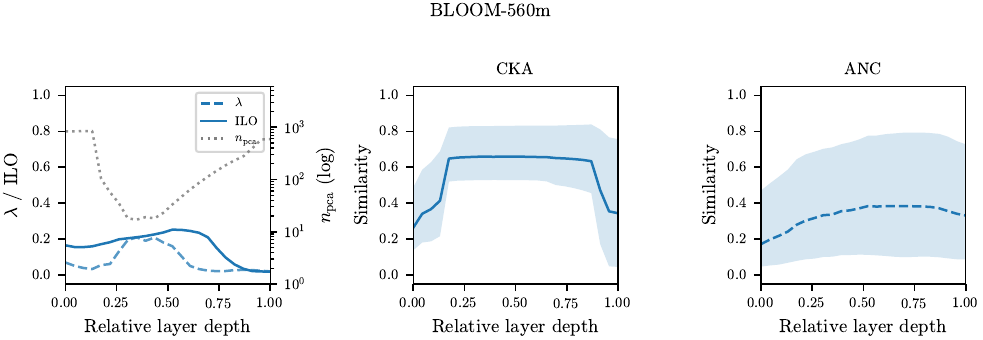}
\caption{\textbf{BLOOM-560M.} Sample-balanced $\lambda$, ILO, and $n_\text{pca}$ (left); CKA mean with min--max band over pairs (middle); ANC with the same convention (right).}
\label{fig:app_metrics_BLOOM-560M}
\end{figure*}

\begin{figure*}[!htbp]
\centering
\includegraphics[width=\textwidth]{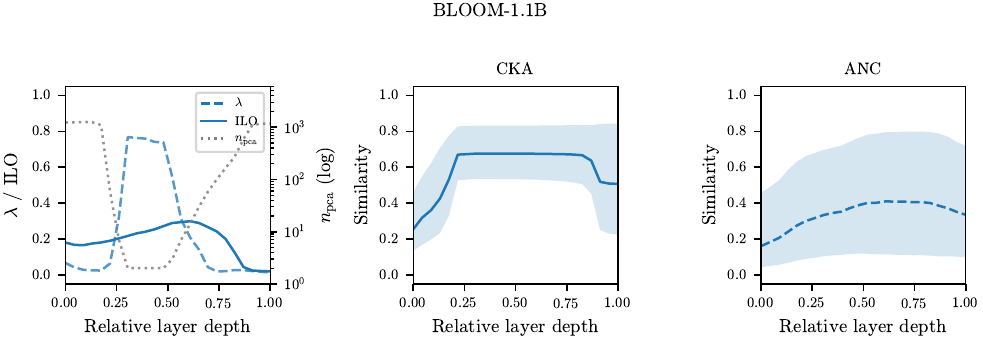}
\caption{\textbf{BLOOM-1.1B.} Panels as in Fig.~\ref{fig:app_metrics_BLOOM-560M}.}
\label{fig:app_metrics_BLOOM-1-1b}
\end{figure*}

\begin{figure*}[!htbp]
\centering
\includegraphics[width=\textwidth]{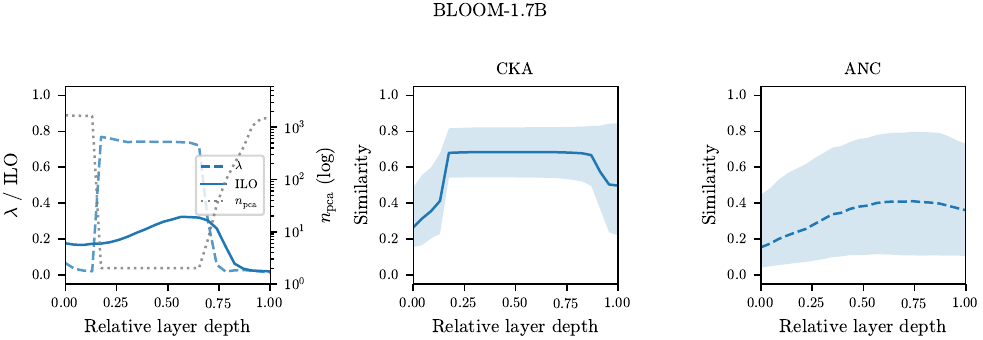}
\caption{\textbf{BLOOM-1.7B.} Panels as in Fig.~\ref{fig:app_metrics_BLOOM-560M}.}
\label{fig:app_metrics_BLOOM-1-7b}
\end{figure*}

\begin{figure*}[!htbp]
\centering
\includegraphics[width=\textwidth]{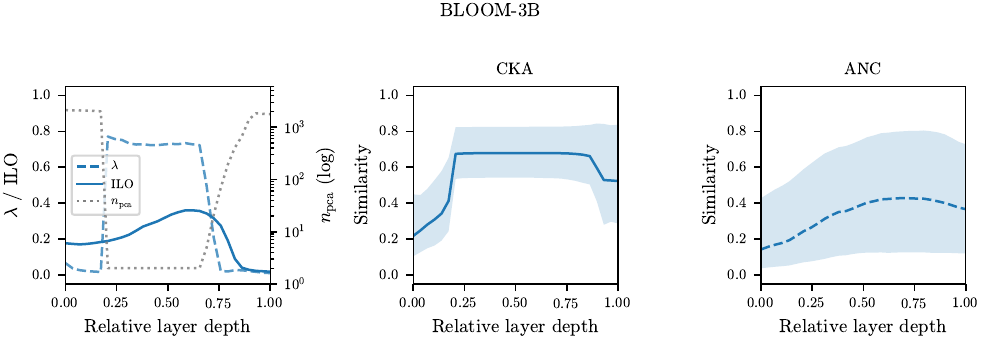}
\caption{\textbf{BLOOM-3B.} Panels as in Fig.~\ref{fig:app_metrics_BLOOM-560M}.}
\label{fig:app_metrics_BLOOM-3B}
\end{figure*}

\begin{figure*}[!htbp]
\centering
\includegraphics[width=\textwidth]{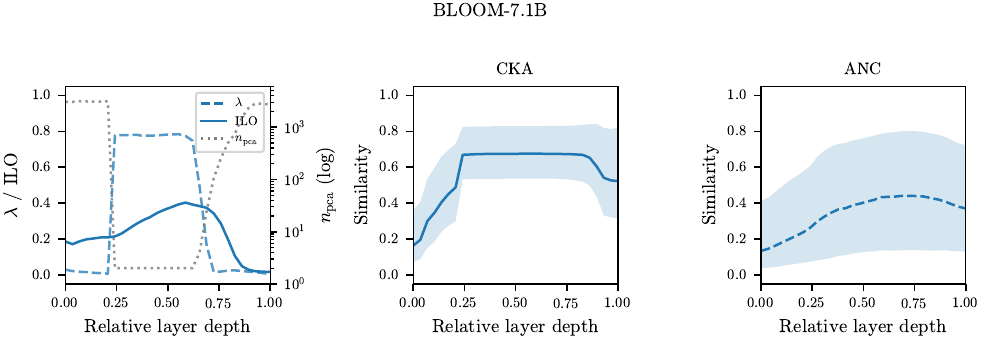}
\caption{\textbf{BLOOM-7.1B.} Panels as in Fig.~\ref{fig:app_metrics_BLOOM-560M}.}
\label{fig:app_metrics_BLOOM-7-1b}
\end{figure*}

\begin{figure*}[!htbp]
\centering
\includegraphics[width=\textwidth]{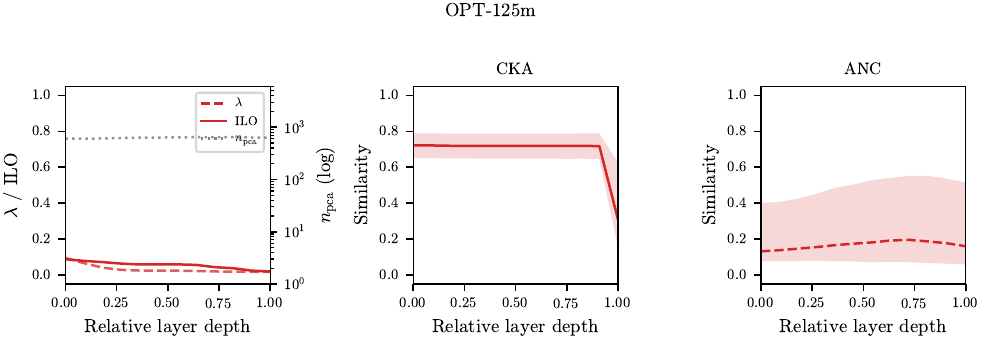}
\caption{\textbf{OPT-125M.} Panels as in Fig.~\ref{fig:app_metrics_BLOOM-560M}.}
\label{fig:app_metrics_OPT-125M}
\end{figure*}

\begin{figure*}[!htbp]
\centering
\includegraphics[width=\textwidth]{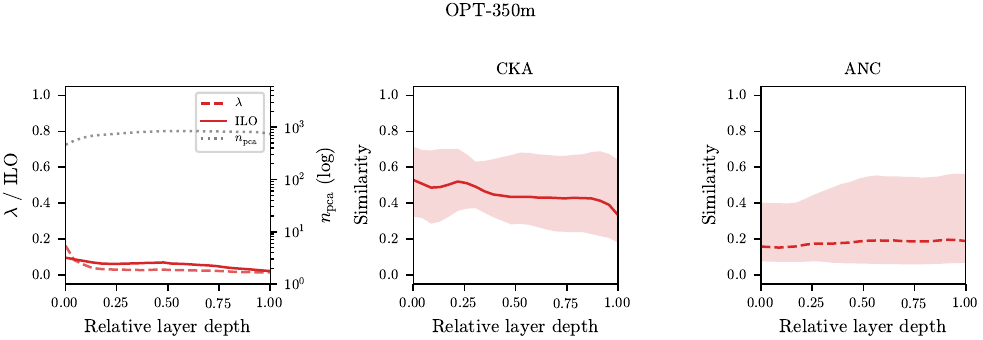}
\caption{\textbf{OPT-350m.} Panels as in Fig.~\ref{fig:app_metrics_BLOOM-560M}.}
\label{fig:app_metrics_OPT-350M}
\end{figure*}

\begin{figure*}[!htbp]
\centering
\includegraphics[width=\textwidth]{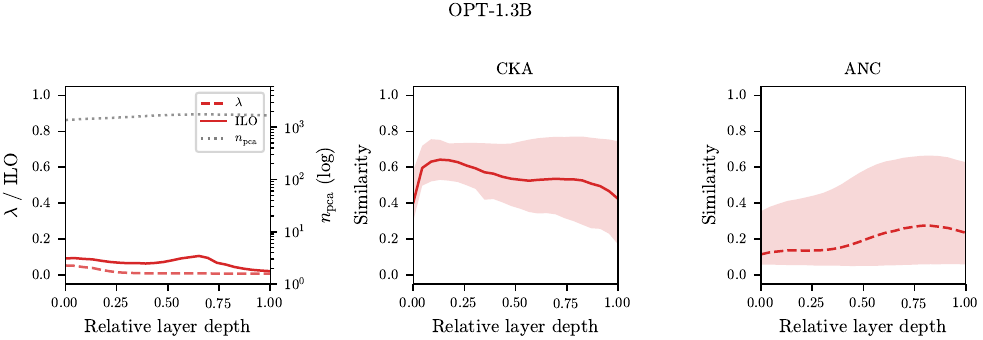}
\caption{\textbf{OPT-1.3B.} Panels as in Fig.~\ref{fig:app_metrics_BLOOM-560M}.}
\label{fig:app_metrics_opt-1-3b}
\end{figure*}

\begin{figure*}[!htbp]
\centering
\includegraphics[width=\textwidth]{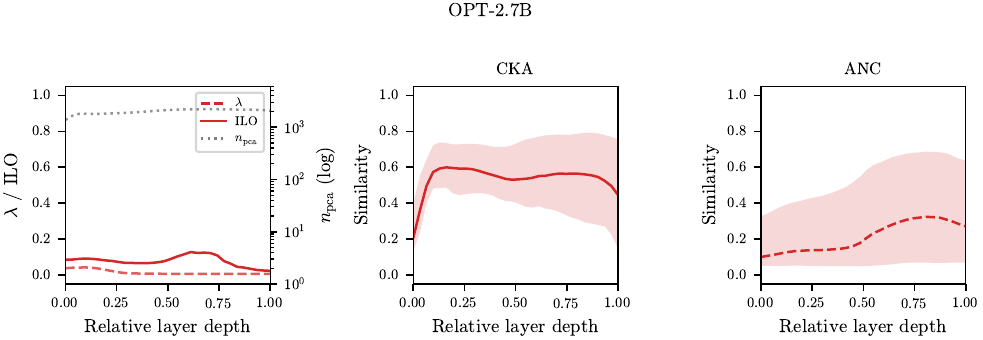}
\caption{\textbf{OPT-2.7B.} Panels as in Fig.~\ref{fig:app_metrics_BLOOM-560M}.}
\label{fig:app_metrics_opt-2-7b}
\end{figure*}

\begin{figure*}[!htbp]
\centering
\includegraphics[width=\textwidth]{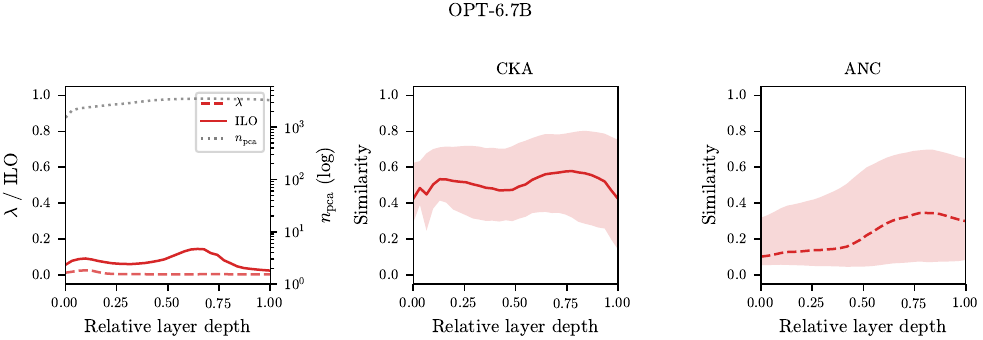}
\caption{\textbf{OPT-6.7B.} Panels as in Fig.~\ref{fig:app_metrics_BLOOM-560M}.}
\label{fig:app_metrics_opt-6-7b}
\end{figure*}

\begin{figure*}[!htbp]
\centering
\includegraphics[width=\textwidth]{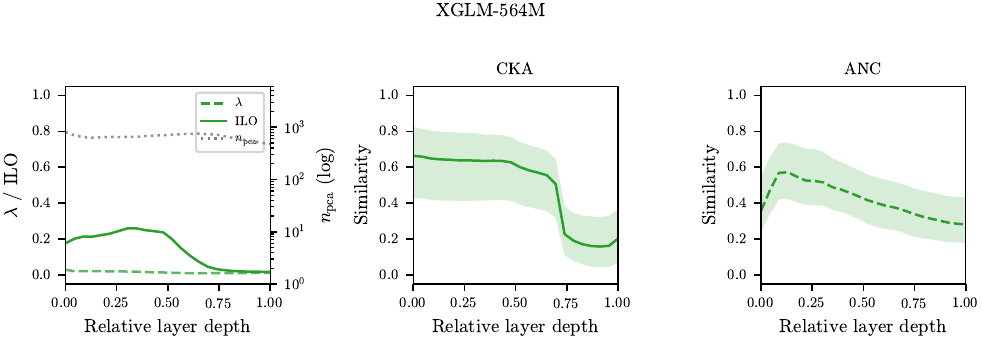}
\caption{\textbf{XGLM-564M.} Panels as in Fig.~\ref{fig:app_metrics_BLOOM-560M}.}
\label{fig:app_metrics_XGLM-564M}
\end{figure*}

\begin{figure*}[!htbp]
\centering
\includegraphics[width=\textwidth]{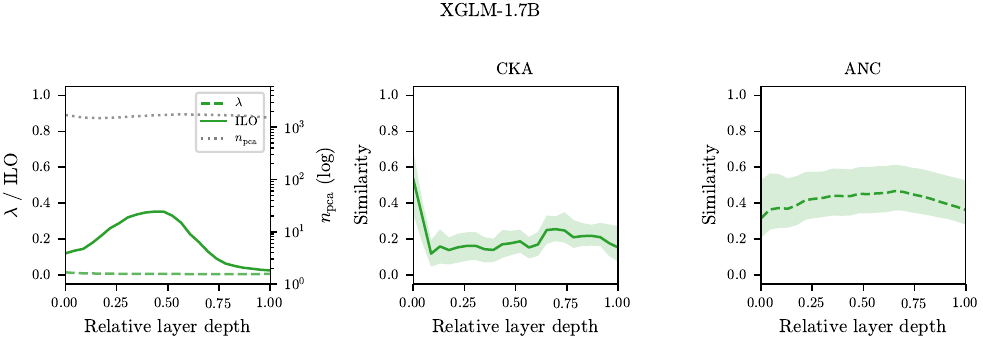}
\caption{\textbf{XGLM-1.7B.} Panels as in Fig.~\ref{fig:app_metrics_BLOOM-560M}.}
\label{fig:app_metrics_XGLM-1-7b}
\end{figure*}

\begin{figure*}[!htbp]
\centering
\includegraphics[width=\textwidth]{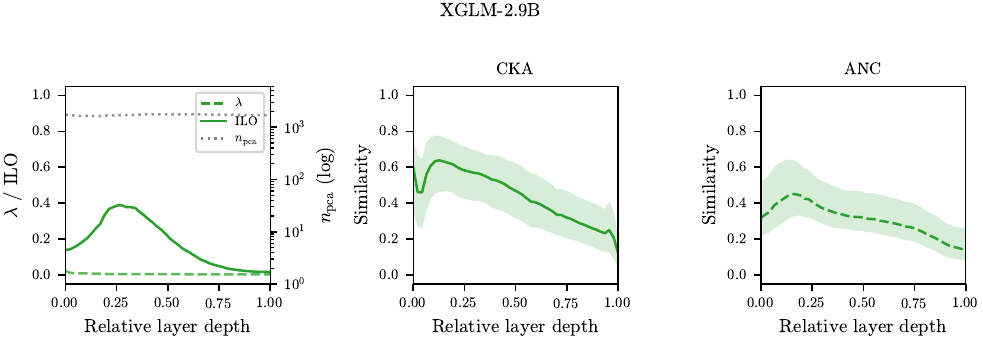}
\caption{\textbf{XGLM-2.9B.} Panels as in Fig.~\ref{fig:app_metrics_BLOOM-560M}.}
\label{fig:app_metrics_XGLM-2-9b}
\end{figure*}

\begin{figure*}[!htbp]
\centering
\includegraphics[width=\textwidth]{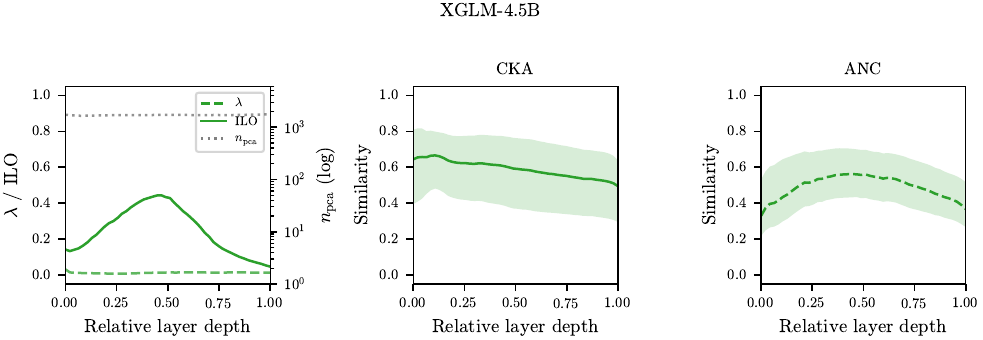}
\caption{\textbf{XGLM-4.5B.} Panels as in Fig.~\ref{fig:app_metrics_BLOOM-560M}.}
\label{fig:app_metrics_XGLM-4-5b}
\end{figure*}

\begin{figure*}[!htbp]
\centering
\includegraphics[width=\textwidth]{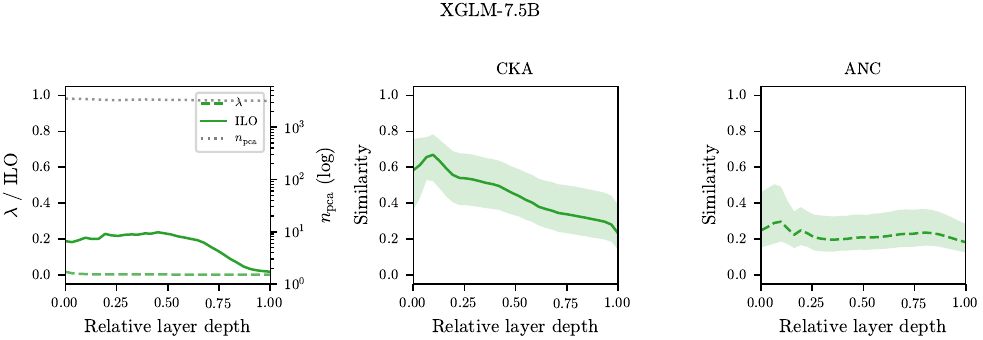}
\caption{\textbf{XGLM-7.5B.} Panels as in Fig.~\ref{fig:app_metrics_BLOOM-560M}.}
\label{fig:app_metrics_XGLM-7-5b}
\end{figure*}

\begin{figure*}[!htbp]
\centering
\includegraphics[width=\textwidth]{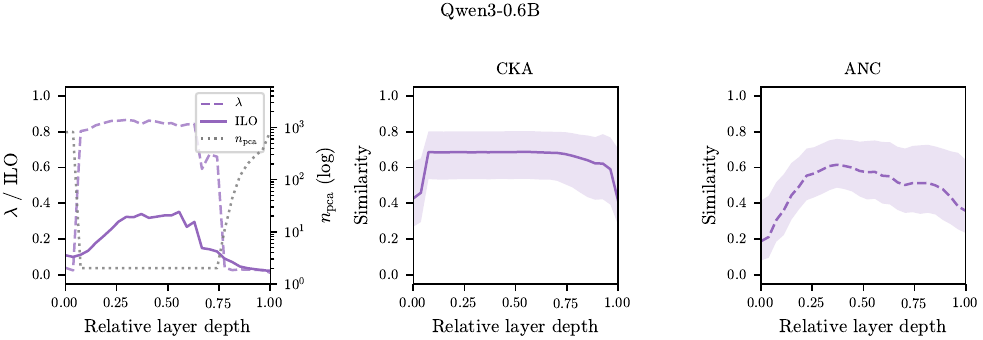}
\caption{\textbf{Qwen3-0.6B.} Panels as in Fig.~\ref{fig:app_metrics_BLOOM-560M}.}
\label{fig:app_metrics_qwen3-0-6b}
\end{figure*}

\begin{figure*}[!htbp]
\centering
\includegraphics[width=\textwidth]{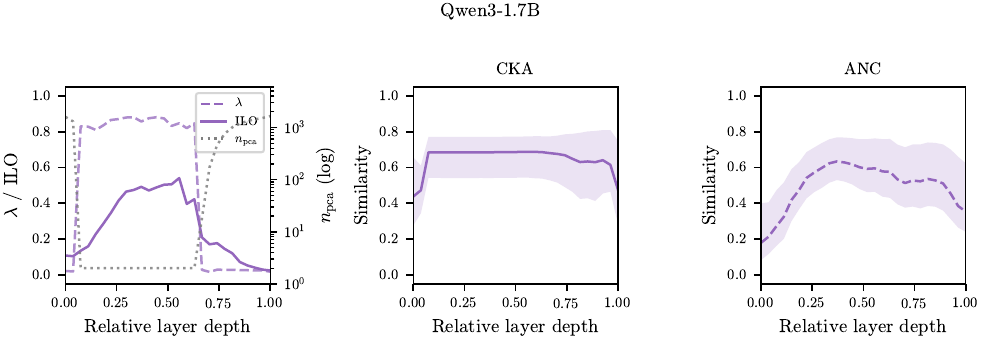}
\caption{\textbf{Qwen3-1.7B.} Panels as in Fig.~\ref{fig:app_metrics_BLOOM-560M}.}
\label{fig:app_metrics_qwen3-1-7b}
\end{figure*}

\begin{figure*}[!htbp]
\centering
\includegraphics[width=\textwidth]{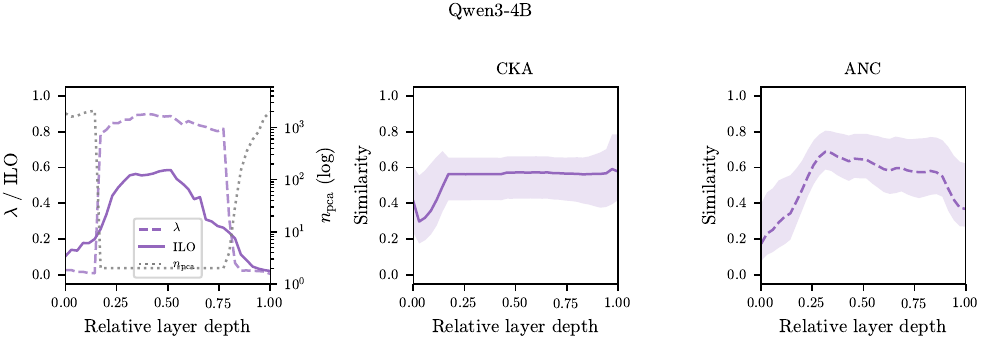}
\caption{\textbf{Qwen3-4B.} Panels as in Fig.~\ref{fig:app_metrics_BLOOM-560M}.}
\label{fig:app_metrics_qwen3-4b}
\end{figure*}

\begin{figure*}[!htbp]
\centering
\includegraphics[width=\textwidth]{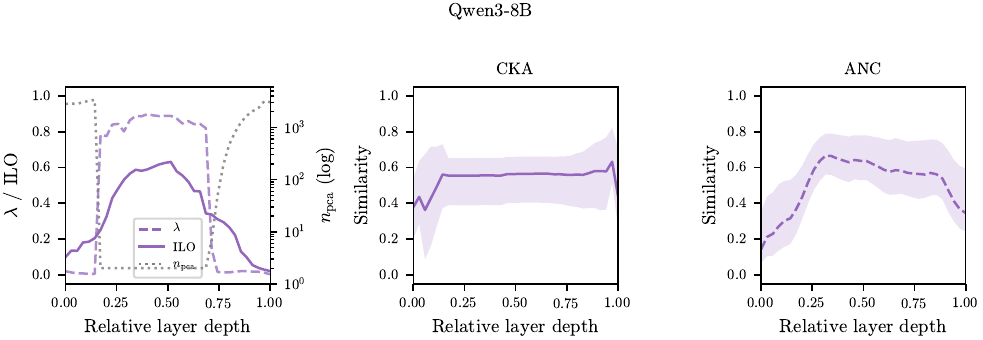}
\caption{\textbf{Qwen3-8B.} Panels as in Fig.~\ref{fig:app_metrics_BLOOM-560M}.}
\label{fig:app_metrics_qwen3-8b}
\end{figure*}

\begin{figure*}[!htbp]
\centering
\includegraphics[width=\textwidth]{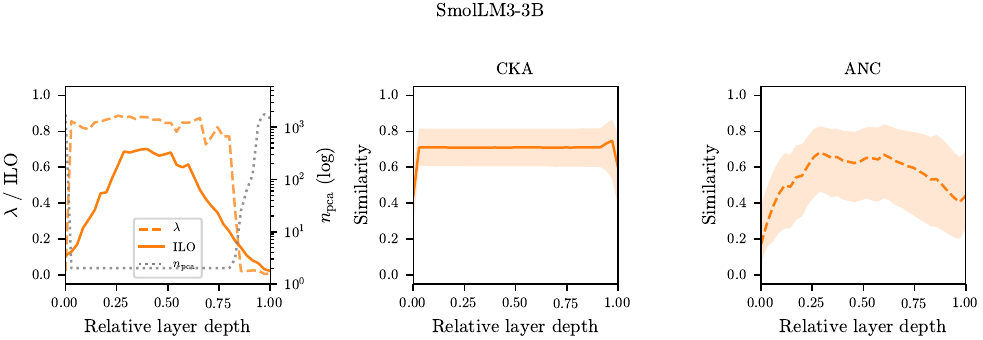}
\caption{\textbf{SmolLM3-3B.} Panels as in Fig.~\ref{fig:app_metrics_BLOOM-560M}.}
\label{fig:app_metrics_smollm3-3b}
\end{figure*}

\begin{figure*}[!htbp]
\centering
\includegraphics[width=\textwidth]{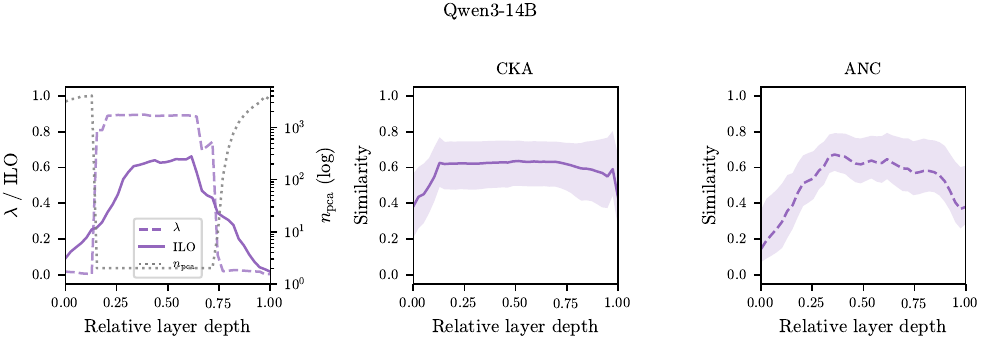}
\caption{\textbf{Qwen3-14B.} Panels as in Fig.~\ref{fig:app_metrics_BLOOM-560M}.}
\label{fig:app_metrics_qwen3-14b}
\end{figure*}

We report the full layer-wise metric profile for each of the 21 base models. Figures~\ref{fig:app_metrics_BLOOM-560M}--\ref{fig:app_metrics_qwen3-14b} show three panels per model: (left) sample-balanced GMM $\lambda$ (dashed), ILO (solid), and effective PCA dimensionality $n_\text{pca}$ (dotted gray, log right axis); (middle) CKA averaged across language pairs, with the shaded band giving the min--max range over pairs; (right) ANC with the same band convention. Splitting CKA and ANC keeps the per-pair spread readable, which a single overlaid panel does not.

\begin{figure*}[!htbp]
\centering
\includegraphics[width=\textwidth]{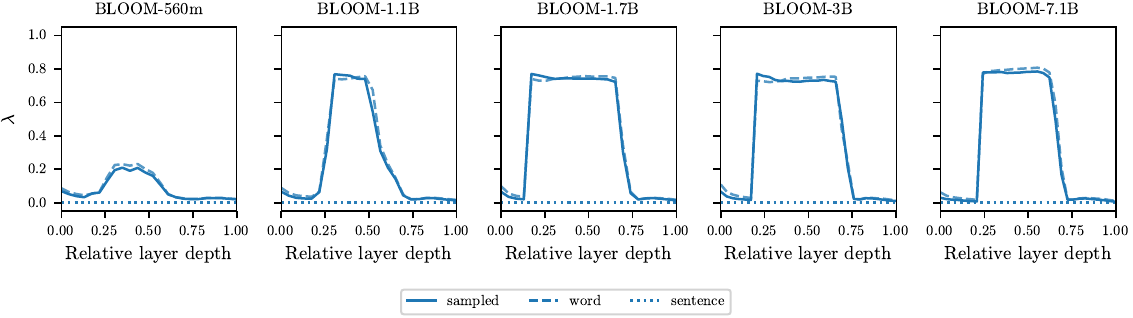}
\caption{\textbf{BLOOM family: $\lambda$ at three granularities.} Sample-balanced (solid), word-level (dashed), and sentence-level (dotted).}
\label{fig:app_lambda_grid_bloom}
\end{figure*}

\paragraph{Granularity of $\lambda$.}
Figures~\ref{fig:app_lambda_grid_bloom}--\ref{fig:app_lambda_grid_smollm3} compare three computations of $\lambda$ within each family: sample-balanced (10{,}000 tokens per language; used elsewhere in the paper), word-level (all token positions on the same GMM), and sentence-level (mean-pooled sentence embeddings). Sample-balanced and word-level $\lambda$ track each other closely across all 21 models, validating the sampling choice used in the main analysis. Sentence-level $\lambda$ diverges along family lines: it is zero throughout all layers for BLOOM, OPT, and XGLM, indicating that sentence-mean pooling collapses the cluster structure that token-level inputs preserve. In Qwen3 and SmolLM3 it instead shows a sharp early-layer peak (relative depth $\leq 0.37$) reaching $0.41$--$0.56$, well below the sample-balanced and word-level peaks ($\approx 0.85$--$0.91$) but clearly non-zero; the peak narrows as Qwen3 scales from 0.6B to 14B.
These granularities also address whether the metrics' different input units explain the cross-metric disagreement of \S\ref{sec:why_disagree}: the disagreements occur within each input class, since CKA and ANC are computed from identical sentence embeddings yet correlate with transfer at $\rho = 0.14$ and $0.84$, while $\lambda$ and ILO are computed from identical token samples yet give opposite verdicts on XGLM ($\lambda = 0.03$, ILO $= 0.44$).
If the input unit were the cause, metrics sharing a unit would agree; instead, the metrics that fail are those exposed to the common-shift geometry (CKA) or confined to the collapsed principal subspace ($\lambda$), irrespective of unit.

\begin{figure*}[!htbp]
\centering
\includegraphics[width=\textwidth]{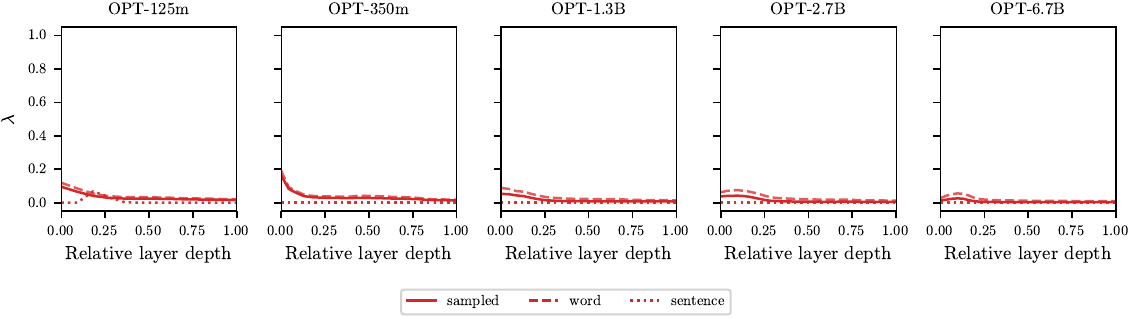}
\caption{\textbf{OPT family.} Lines as in Fig.~\ref{fig:app_lambda_grid_bloom}.}
\label{fig:app_lambda_grid_opt}
\end{figure*}

\begin{figure*}[!htbp]
\centering
\includegraphics[width=\textwidth]{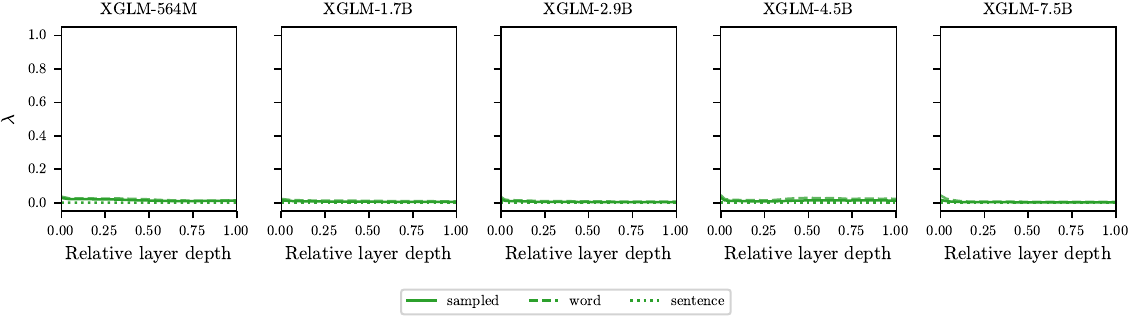}
\caption{\textbf{XGLM family.} Lines as in Fig.~\ref{fig:app_lambda_grid_bloom}.}
\label{fig:app_lambda_grid_xglm}
\end{figure*}

\begin{figure*}[!htbp]
\centering
\includegraphics[width=\textwidth]{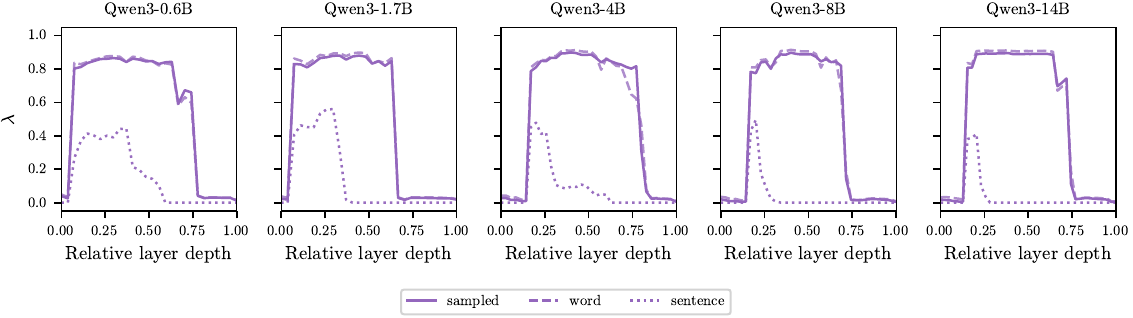}
\caption{\textbf{Qwen3 family.} Lines as in Fig.~\ref{fig:app_lambda_grid_bloom}.}
\label{fig:app_lambda_grid_qwen3}
\end{figure*}

\FloatBarrier

\begin{figure*}[!htbp]
\centering
\includegraphics[width=\textwidth]{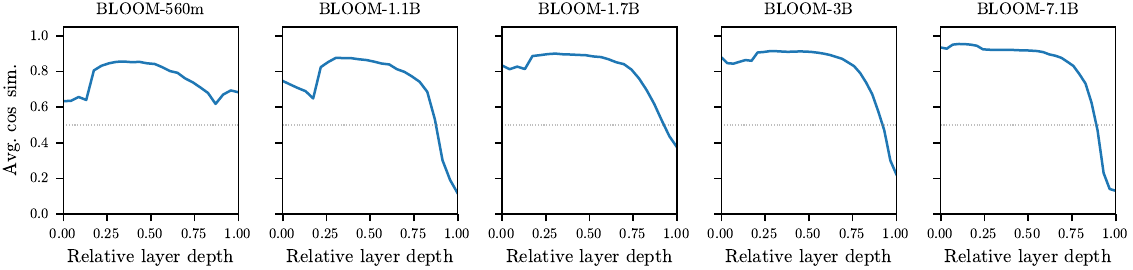}
\caption{\textbf{BLOOM family: average cosine similarity between random hidden-state pairs} across layers. The dotted gray line marks $0.5$.}
\label{fig:app_avgcos_bloom}
\end{figure*}

\begin{figure*}[!htbp]
\centering
\includegraphics[width=\textwidth]{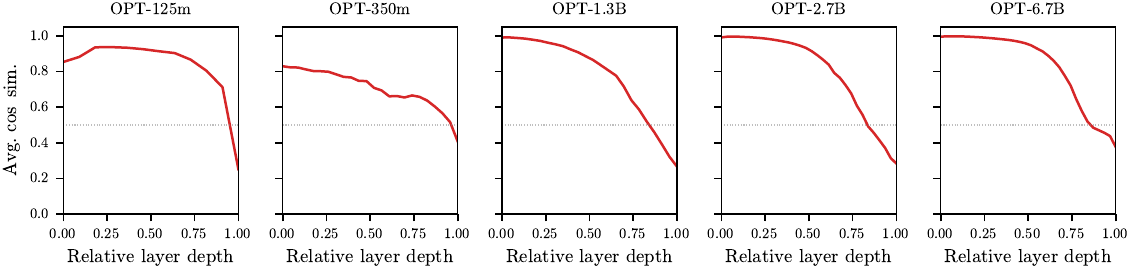}
\caption{\textbf{OPT family.} Layout as in Fig.~\ref{fig:app_avgcos_bloom}.}
\label{fig:app_avgcos_opt}
\end{figure*}

\begin{figure*}[!htbp]
\centering
\includegraphics[width=\textwidth]{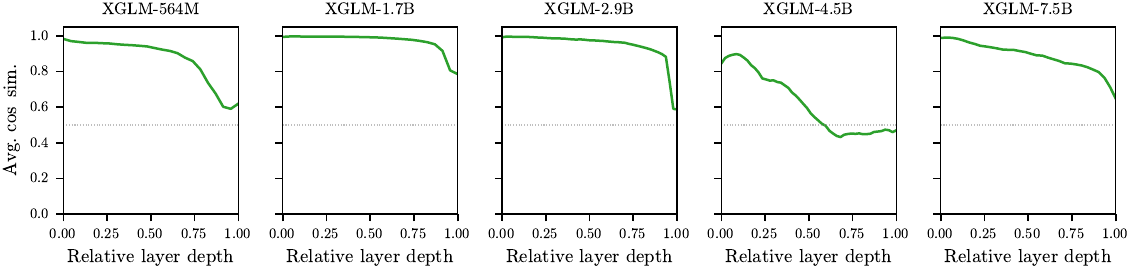}
\caption{\textbf{XGLM family.} Layout as in Fig.~\ref{fig:app_avgcos_bloom}.}
\label{fig:app_avgcos_xglm}
\end{figure*}

\begin{figure*}[!htbp]
\centering
\includegraphics[width=\textwidth]{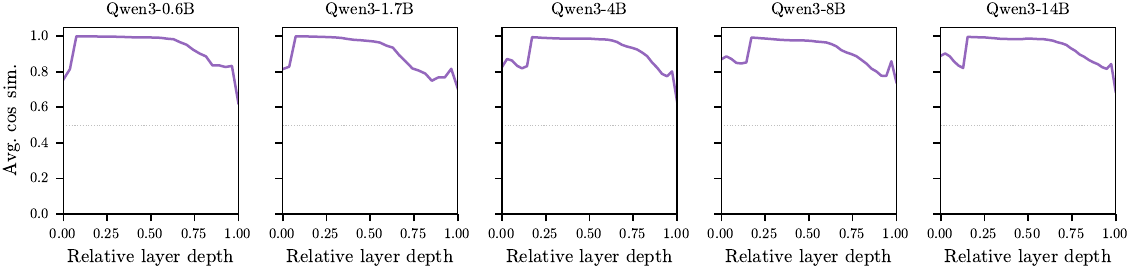}
\caption{\textbf{Qwen3 family.} Layout as in Fig.~\ref{fig:app_avgcos_bloom}.}
\label{fig:app_avgcos_qwen3}
\end{figure*}

\begin{figure}[!htbp]
\centering
\includegraphics[width=\columnwidth]{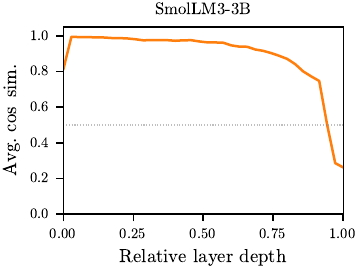}
\caption{\textbf{SmolLM3 family} (single model). Layout as in Fig.~\ref{fig:app_avgcos_bloom}.}
\label{fig:app_avgcos_smollm3}
\end{figure}

\noindent
\begin{minipage}[t]{0.47\textwidth}
\vspace{0pt}
\subsection{Anisotropy Analysis}
\label{sec:app_anisotropy}

\paragraph{Effective dimensionality at the peak-sharing layer.}
Table~\ref{tab:appendix-anisotropy-peak} reports $n_\text{pca}$ at the $98\%$ threshold used by the $\lambda$ pipeline and at a more relaxed $90\%$ alternative, computed on the same balanced subsample dataset (10{,}000 tokens per language). Both thresholds apply the pipeline's floor of $2$, following the \citet{shani-basirat-2025} implementation.

\paragraph{Cosine similarity profiles.}
Effective PCA dimensionality $n_\text{pca}$ is reported per layer for every model in Appendix~\ref{sec:app_fullresults}. Here we complement it with the canonical angular diagnostic: the average cosine similarity ($\overline{\cos}$) between random pairs of hidden states \citep{ethayarajh-2019-contextual}. Figures~\ref{fig:app_avgcos_bloom}--\ref{fig:app_avgcos_smollm3} show this quantity across layers, grouped by family.

All five families exhibit pronounced middle-layer anisotropy: $\overline{\cos}$ exceeds $0.85$ at mid-depth in every model; BLOOM-560M is the weakest case, capping at $0.86$ without ever plateauing. OPT and Qwen3 reach extreme values ($\geq 0.99$ in some layers), and BLOOM, OPT, Qwen3, and SmolLM3 all show the same overall shape: rapid rise from input layers, a long high-similarity plateau, and a sharp drop in the final two layers. XGLM is the family with the most variation: XGLM-4.5B in particular shows a substantial mid-network dip ($0.57$) absent in the other XGLM models. High anisotropy is therefore a generic property of the families we study, not a property of the high-$\lambda$ subset alone.

\end{minipage}

\hfill

\begin{minipage}[t]{0.50\textwidth}
\vspace{0pt}
\centering

\includegraphics[width=\linewidth]{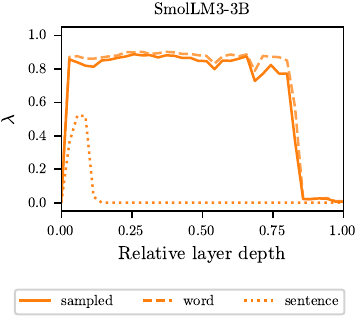}
\captionof{figure}{%
  \textbf{SmolLM3 family} (single model).
  Lines as in Fig.~\ref{fig:app_lambda_grid_bloom}.%
}
\label{fig:app_lambda_grid_smollm3}

\vspace{1em}

\scriptsize
\begin{tabular}{llrrrr}
\toprule
Family & Model & Peak~L &
$n_\text{pca}^{98}$ & $n_\text{pca}^{90}$ & $d$ \\
\midrule
BLOOM  & BLOOM-560M & 10 & 20 & 2 & 1024 \\
       & BLOOM-1.1B & 11 & 2 & 2 & 1536 \\
       & BLOOM-1.7B & 11 & 2 & 2 & 2048 \\
       & BLOOM-3B   & 18 & 2 & 2 & 2560 \\
       & BLOOM-7.1B & 16 & 2 & 2 & 4096 \\
\midrule
BLOOMZ & BLOOMZ-560M & 11 & 2 & 2 & 1024 \\
       & BLOOMZ-1.1B & 12 & 2 & 2 & 1536 \\
       & BLOOMZ-1.7B & 16 & 2 & 2 & 2048 \\
       & BLOOMZ-3B   & 23 & 2 & 2 & 2560 \\
       & BLOOMZ-7.1B & 23 & 2 & 2 & 4096 \\
\midrule
Qwen3  & Qwen3-0.6B & 8  & 2 & 2 & 1024 \\
       & Qwen3-1.7B & 12 & 2 & 2 & 2048 \\
       & Qwen3-4B   & 15 & 2 & 2 & 2560 \\
       & Qwen3-8B   & 14 & 2 & 2 & 4096 \\
       & Qwen3-14B  & 15 & 2 & 2 & 5120 \\
\midrule
SmolLM3 & SmolLM3-3B & 10 & 2 & 2 & 2048 \\
\midrule
OPT & OPT-125M & $\phantom{0}0^\dagger$ & 609 & 348 & 768 \\
    & OPT-350M & $\phantom{0}0^\dagger$ & 464 & 282 & 1024 \\
    & OPT-1.3B & $15^\dagger$ & 1753 & 1103 & 2048 \\
    & OPT-2.7B & $19^\dagger$ & 2189 & 1374 & 2560 \\
    & OPT-6.7B & $19^\dagger$ & 3477 & 2144 & 4096 \\
\midrule
XGLM & XGLM-564M & $11^\dagger$ & 699 & 297 & 1024 \\
     & XGLM-1.7B & $11^\dagger$ & 1713 & 1014 & 2048 \\
     & XGLM-2.9B & $11^\dagger$ & 1680 & 968 & 2048 \\
     & XGLM-4.5B & $19^\dagger$ & 1702 & 1015 & 2048 \\
     & XGLM-7.5B & $17^\dagger$ & 3306 & 1690 & 4096 \\
\bottomrule
\end{tabular}
\captionof{table}{%
  \textbf{Effective PCA dimensionality at each model's
  peak-sharing layer.} Peak layer maximises $\lambda$; for OPT
  and XGLM ($\lambda$ near zero throughout) we use the peak
  mean-ILO layer instead ($^\dagger$). $d$ is the hidden dimensionality.%
}
\label{tab:appendix-anisotropy-peak}

\end{minipage}

\FloatBarrier

\subsection{Probe Results}
\label{sec:app_probe_results}

\begin{figure*}[!htbp]
\centering
\includegraphics[width=\textwidth]{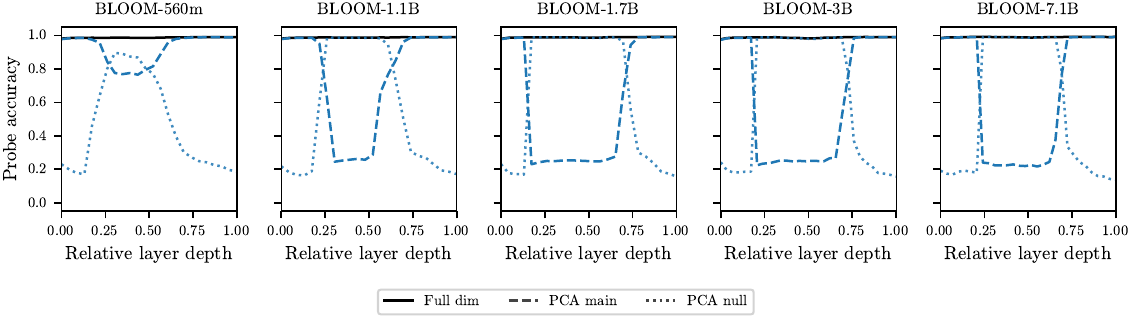}
\caption{\textbf{BLOOM family: word-level language probe accuracy} across layers, trained on the full hidden state, the PCA main subspace, or the PCA null complement. Random baseline is $0.1$.}
\label{fig:app_probe_bloom}
\end{figure*}

Figures~\ref{fig:app_probe_bloom}--\ref{fig:app_probe_smollm3} show word-level language-probe accuracy per model: linear probes trained on (i) the full hidden state, (ii) the PCA principal subspace retained at 98\% variance, and (iii) the PCA residual subspace. Chance accuracy for the 10-language classification is $0.1$.

The full-dimension probe achieves $\geq 0.96$ accuracy at every layer of every model, including layers where $\lambda$ is high---language identity is never destroyed, only relocated. The main--null relationship splits along the same family lines as in $\lambda$: BLOOM (from 1.1B onward), every Qwen3 model, and SmolLM3 show a clean inversion in middle layers where main-subspace accuracy collapses to near chance ($\approx 0.09$) and null-subspace accuracy rises to $\geq 0.99$. The width and depth of this inversion track $\lambda$: BLOOM-560M, the subthreshold model in its family, shows only a partial dip (main bottoms at $0.77$). In OPT and XGLM no inversion occurs at any layer: main-subspace accuracy tracks full-dimension accuracy throughout, and null-subspace accuracy stays low. The inversion is therefore not a generic anisotropy artifact---it requires the PCA-collapse pattern documented in \S\ref{sec:why_disagree}.

\begin{figure*}[!htbp]
\centering
\includegraphics[width=\textwidth]{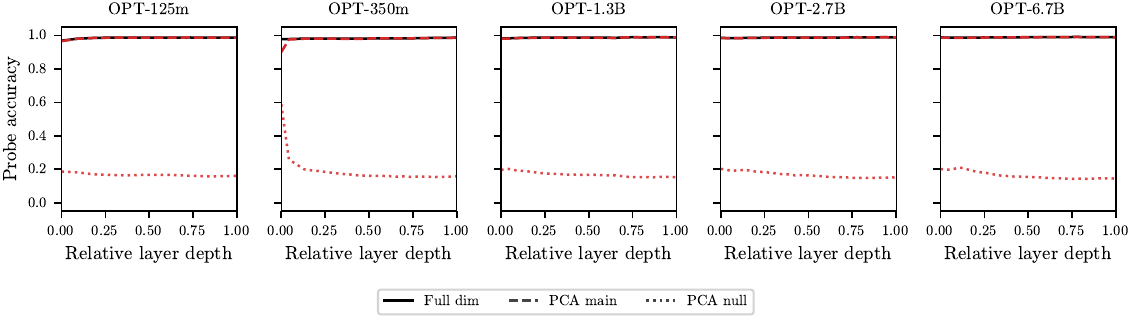}
\caption{\textbf{OPT family.} Layout as in Fig.~\ref{fig:app_probe_bloom}.}
\label{fig:app_probe_opt}
\end{figure*}

\begin{figure*}[!htbp]
\centering
\includegraphics[width=\textwidth]{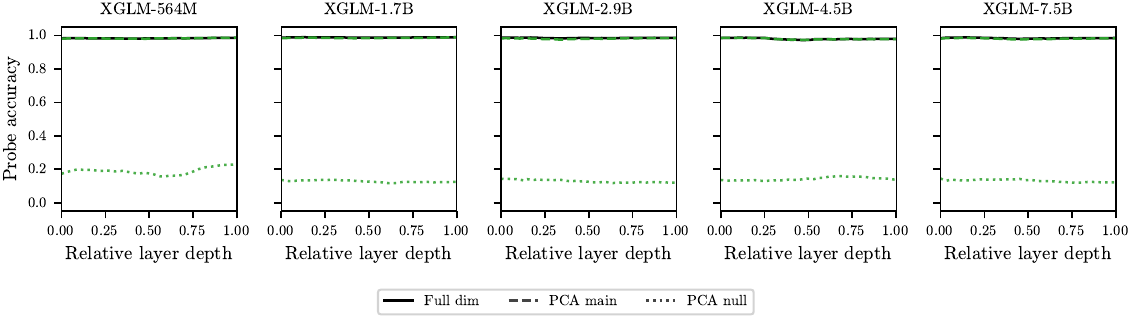}
\caption{\textbf{XGLM family.} Layout as in Fig.~\ref{fig:app_probe_bloom}.}
\label{fig:app_probe_xglm}
\end{figure*}

\begin{figure*}[!htbp]
\centering
\includegraphics[width=\textwidth]{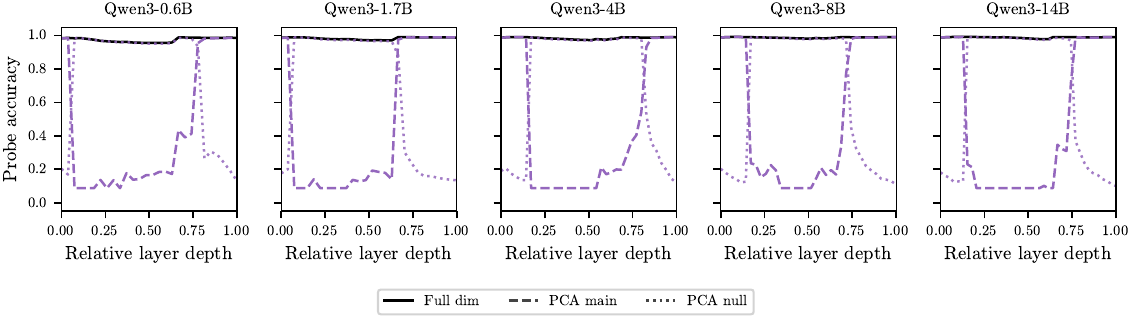}
\caption{\textbf{Qwen3 family.} Layout as in Fig.~\ref{fig:app_probe_bloom}.}
\label{fig:app_probe_qwen3}
\end{figure*}

\FloatBarrier

\noindent

\begin{minipage}[t]{0.47\textwidth}
\vspace{0pt}
\subsection{ILO Centering}
\label{sec:app_ilo_centering}
To check whether ILO captures genuine cross-lingual neighborhood structure
rather than anisotropic mean offsets, we recompute the metric after
subtracting the per-layer grand mean activation (over all tokens and
languages) from each hidden state before forming neighborhoods
(Table~\ref{tab:app_ilo_centering}). Across all 21 base models, the centered
peak ILO differs from the uncentered peak by at most $0.010$ and by less
than $0.005$ for 17 of 21 models. The sign of the shift varies and the
within-family ordering is preserved.

\subsection{Sentence-Level ILO Variant}

\label{sec:app_ilo_variant}

The released implementation of \citet{wilie-etal-2025} computes ILO on
mean-pooled sentence embeddings rather than on token-level representations.
For comparability with that implementation, we also computed this
sentence-level variant for all 21 models.
Its transfer correlation is close to the word-level result reported in the
body ($\rho = 0.93$ vs.\ $0.90$; the two variants agree on model ranks at
$\rho = 0.95$), and no conclusion in the paper changes under either variant.
We nevertheless present the word-level variant throughout, for two reasons.
First, it shares $\lambda$'s balanced $10{,}000$-token-per-language
subsample, so the two metrics are computed on identical inputs.
Second, the two variants differ sharply in their exposure to anisotropy:
recomputing each after subtracting the per-layer grand mean
(\S\ref{sec:app_ilo_centering}) shifts word-level peak ILO by at most $0.010$
(mean $|\Delta| = 0.003$), whereas sentence-level peak ILO shifts by up to
$0.127$ (mean $|\Delta| = 0.043$), with the largest shifts in the most
anisotropic models (Qwen3-14B, SmolLM3).
The sentence-level variant thus illustrates the paper's central point: the  same metric, applied to representations more exposed to the anisotropic common direction, absorbs measurably more of that direction into its score, and anisotropy diagnostics are needed to see it.

\subsection{Per-language perplexity}
Table~\ref{tab:app_perplexity} reports mean FLORES-200 devtest perplexity
per language for each base model. Cross-language variance is substantial
even for high-ILO models, confirming that ILO measures cross-lingual
representational \emph{overlap} rather than uniformly low surprise across
languages.

\end{minipage}
\hfill
\begin{minipage}[t]{0.50\textwidth}
\vspace{0pt}
\centering

\includegraphics[width=\linewidth]{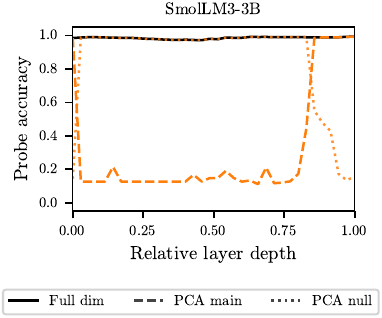}

\captionof{figure}{%
  \textbf{SmolLM3 family} (single model).
  Layout as in Fig.~\ref{fig:app_probe_bloom}.%
}
\label{fig:app_probe_smollm3}
\vspace{1em}

\scriptsize
\setlength{\tabcolsep}{3pt}
\begin{tabular}{llccc}
\toprule
\textbf{Family} & \textbf{Model} &
\textbf{Peak ILO} & \textbf{Peak ILO}$_c$ & $\Delta$ \\
\midrule
BLOOM & BLOOM-560M & 0.252 & 0.251 & $-$0.001 \\
      & BLOOM-1.1B & 0.299 & 0.303 & +0.003 \\
      & BLOOM-1.7B & 0.324 & 0.325 & +0.001 \\
      & BLOOM-3B   & 0.360 & 0.364 & +0.004 \\
      & BLOOM-7.1B & 0.403 & 0.406 & +0.003 \\
\midrule
OPT   & OPT-125M   & 0.088 & 0.092 & +0.004 \\
      & OPT-350m   & 0.097 & 0.094 & $-$0.002 \\
      & OPT-1.3B   & 0.106 & 0.107 & +0.001 \\
      & OPT-2.7B   & 0.128 & 0.129 & +0.001 \\
      & OPT-6.7B   & 0.145 & 0.145 & $-$0.000 \\
\midrule
XGLM  & XGLM-564M  & 0.260 & 0.252 & $-$0.008 \\
      & XGLM-1.7B  & 0.353 & 0.352 & $-$0.002 \\
      & XGLM-2.9B  & 0.388 & 0.390 & +0.002 \\
      & XGLM-4.5B  & 0.444 & 0.446 & +0.002 \\
      & XGLM-7.5B  & 0.238 & 0.229 & $-$0.009 \\
\midrule
Qwen3 & Qwen3-0.6B & 0.353 & 0.363 & +0.010 \\
      & Qwen3-1.7B & 0.541 & 0.545 & +0.004 \\
      & Qwen3-4B   & 0.586 & 0.585 & $-$0.001 \\
      & Qwen3-8B   & 0.631 & 0.628 & $-$0.003 \\
      & Qwen3-14B  & 0.662 & 0.658 & $-$0.004 \\
\midrule
SmolLM3 & SmolLM3-3B & 0.701 & 0.707 & +0.005 \\
\bottomrule
\end{tabular}
\captionof{table}{%
  \textbf{ILO peak before and after grand-mean centering.}
  Centered ILO subtracts the per-layer grand mean (over all tokens and
  languages) from each hidden state before computing neighborhood overlap.
  Small $\Delta$ values indicate that ILO measures genuine cross-lingual
  neighborhood structure rather than the shared anisotropic direction
  common to all languages.%
}
\label{tab:app_ilo_centering}
\end{minipage}
\par\medskip

\begin{table*}[!htbp]
\centering
\small
\setlength{\tabcolsep}{3pt}
\begin{tabular}{llrrrrrrrrrr}
\toprule
\textbf{Family} & \textbf{Model} & \textbf{ar} & \textbf{zh} & \textbf{en} & \textbf{fr} & \textbf{de} & \textbf{ja} & \textbf{pt} & \textbf{ru} & \textbf{es} & \textbf{tr} \\
\midrule
  BLOOM     & BLOOM-560M     & 77.1 & 108.2 & 51.5 & 29.5 & 166.6 & 85.2 & 45.7 & 39.0 & 36.0 & 328.9 \\
            & BLOOM-1.1B     & 59.9 & 75.7 & 36.7 & 22.0 & 93.2 & 58.6 & 33.0 & 26.6 & 27.2 & 147.6 \\
            & BLOOM-1.7B     & 51.7 & 65.0 & 32.5 & 19.7 & 66.0 & 42.7 & 28.9 & 20.5 & 24.4 & 100.7 \\
            & BLOOM-3B       & 45.9 & 59.4 & 29.6 & 18.1 & 51.8 & 36.4 & 26.7 & 17.2 & 22.8 & 78.9 \\
            & BLOOM-7.1B     & 39.2 & 50.6 & 26.0 & 16.5 & 36.0 & 28.8 & 23.6 & 13.5 & 20.8 & 58.3 \\
\midrule
  OPT       & OPT-125M       & 9.7 & 33.1 & 52.5 & 33.1 & 31.5 & 14.5 & 34.7 & 4.5 & 32.2 & 34.5 \\
            & OPT-350m       & 8.0 & 27.8 & 44.0 & 24.4 & 23.7 & 11.4 & 27.1 & 3.8 & 24.5 & 26.0 \\
            & OPT-1.3B       & 6.7 & 19.3 & 32.6 & 13.4 & 13.7 & 8.5 & 15.4 & 3.3 & 14.4 & 16.3 \\
            & OPT-2.7B       & 6.2 & 15.9 & 30.4 & 11.5 & 11.7 & 7.5 & 13.3 & 3.1 & 12.8 & 14.0 \\
            & OPT-6.7B       & 5.6 & 13.3 & 28.0 & 10.0 & 10.0 & 6.6 & 11.5 & 2.9 & 11.1 & 12.0 \\
\midrule
  XGLM      & XGLM-564M      & 57.5 & 51.0 & 38.1 & 21.5 & 35.8 & 33.9 & 35.6 & 32.4 & 32.5 & 73.8 \\
            & XGLM-1.7B      & 45.0 & 42.7 & 33.2 & 18.4 & 28.7 & 28.5 & 29.8 & 26.4 & 27.6 & 56.3 \\
            & XGLM-2.9B      & 35.0 & 34.4 & 27.6 & 15.6 & 23.9 & 23.1 & 25.1 & 22.7 & 23.4 & 46.0 \\
            & XGLM-4.5B      & 36.4 & 31.1 & 25.2 & 14.1 & 21.4 & 21.6 & 24.4 & 20.9 & 21.6 & 47.1 \\
            & XGLM-7.5B      & 32.5 & 30.5 & 24.6 & 13.9 & 20.9 & 20.7 & 22.3 & 20.6 & 21.0 & 39.7 \\
\midrule
  Qwen3     & Qwen3-0.6B     & 26.7 & 46.3 & 31.9 & 16.0 & 21.2 & 29.1 & 22.4 & 13.5 & 20.9 & 34.9 \\
            & Qwen3-1.7B     & 17.7 & 34.9 & 25.2 & 11.4 & 13.8 & 19.4 & 16.0 & 9.9 & 15.2 & 21.4 \\
            & Qwen3-4B       & 14.8 & 29.9 & 22.3 & 9.7 & 11.5 & 16.1 & 13.6 & 8.6 & 13.1 & 17.0 \\
            & Qwen3-8B       & 12.8 & 27.4 & 20.5 & 8.8 & 10.2 & 14.2 & 12.2 & 7.8 & 11.9 & 14.6 \\
            & Qwen3-14B      & 11.7 & 25.4 & 19.5 & 8.4 & 9.5 & 13.1 & 11.4 & 7.3 & 11.1 & 13.3 \\
\midrule
  SmolLM3   & SmolLM3-3B     & -- & -- & 20.8 & 8.1 & 8.8 & -- & 10.9 & -- & 10.6 & -- \\
\bottomrule
\end{tabular}
\caption{\textbf{Mean FLORES-200 devtest perplexity per language, per model.}
Lower is better. Languages: ar=Arabic, zh=Chinese, en=English, fr=French,
de=German, ja=Japanese, pt=Portuguese, ru=Russian, es=Spanish, tr=Turkish.}
\label{tab:app_perplexity}
\end{table*}

\clearpage
\FloatBarrier

\subsection{Cross-Lingual Transfer Results}
\label{sec:app_xlang_results}

We report full cross-lingual transfer results. Table~\ref{tab:app_transfer_summary} summarizes mean accuracy (F1 for XQuAD) on non-English target languages after English-only finetuning at the best learning rate per (model, task) pair; per-(model, task) learning rates are in Table~\ref{tab:app_selected_lr}. Tables~\ref{tab:app_transfer_belebele}--\ref{tab:app_transfer_xquad} give the per-language breakdown for each task. Language coverage varies by task: Belebele and SIB-200 cover all 10 evaluation languages, XCSR drops Turkish, XNLI drops Japanese and Portuguese, and XQuAD drops Japanese, French, and Portuguese.

\begin{table*}[!htbp]
\centering
\small
\begin{tabular}{llccccc}
\toprule
\textbf{Family} & \textbf{Model} & \textbf{Belebele} & \textbf{XNLI} & \textbf{XCSR} & \textbf{SIB-200} & \textbf{XQuAD} \\
\midrule
  BLOOM     & BLOOM-560M     & 0.295 & 0.606 & 0.391 & 0.594 & 0.334 \\
            & BLOOM-1.1B     & 0.320 & 0.646 & 0.406 & 0.547 & 0.430 \\
            & BLOOM-1.7B     & 0.321 & 0.672 & 0.445 & 0.469 & 0.394 \\
            & BLOOM-3B       & 0.375 & 0.691 & 0.460 & 0.747 & 0.466 \\
            & BLOOM-7.1B     & 0.405 & 0.729 & 0.498 & 0.771 & 0.581 \\
\midrule
  OPT       & OPT-125M       & 0.277 & 0.422 & 0.207 & 0.406 & 0.072 \\
            & OPT-350m       & 0.284 & 0.432 & 0.204 & 0.412 & 0.091 \\
            & OPT-1.3B       & 0.301 & 0.489 & 0.257 & 0.574 & 0.154 \\
            & OPT-2.7B       & 0.299 & 0.499 & 0.276 & 0.613 & 0.204 \\
            & OPT-6.7B       & 0.321 & 0.529 & 0.307 & 0.653 & 0.228 \\
\midrule
  XGLM      & XGLM-564M      & 0.285 & 0.666 & 0.421 & 0.660 & 0.537 \\
            & XGLM-1.7B      & 0.312 & 0.708 & 0.466 & 0.819 & 0.351 \\
            & XGLM-2.9B      & 0.351 & 0.718 & 0.513 & 0.881 & 0.380 \\
            & XGLM-4.5B      & 0.324 & 0.745 & 0.547 & 0.747 & 0.412 \\
            & XGLM-7.5B      & 0.363 & 0.759 & 0.566 & 0.842 & 0.473 \\
\midrule
  Qwen3     & Qwen3-0.6B     & 0.376 & 0.712 & 0.399 & 0.798 & 0.519 \\
            & Qwen3-1.7B     & 0.443 & 0.694 & 0.490 & 0.810 & 0.618 \\
            & Qwen3-4B       & 0.478 & 0.828 & 0.530 & 0.814 & 0.674 \\
            & Qwen3-8B       & 0.483 & 0.823 & 0.535 & 0.853 & 0.634 \\
            & Qwen3-14B      & 0.479 & 0.850 & 0.554 & 0.849 & 0.690 \\
\midrule
  SmolLM3   & SmolLM3-3B     & 0.458 & 0.798 & 0.614 & 0.801 & 0.577 \\
\bottomrule
\end{tabular}
\caption{\textbf{Cross-lingual transfer summary.} Mean accuracy
(F1 for XQuAD) on non-English target languages after English-only
finetuning at the best learning rate per (model, task) pair.}
\label{tab:app_transfer_summary}
\end{table*}

\begin{table*}[!htbp]
\centering
\small
\setlength{\tabcolsep}{4pt}
\begin{tabular}{llcccccccccc}
\toprule
\textbf{Family} & \textbf{Model} & \textbf{ar} & \textbf{zh} & \textbf{en} & \textbf{fr} & \textbf{de} & \textbf{ja} & \textbf{pt} & \textbf{ru} & \textbf{es} & \textbf{tr} \\
\midrule
  BLOOM     & BLOOM-560M     & 0.273 & 0.288 & 0.316 & 0.323 & 0.300 & 0.281 & 0.300 & 0.309 & 0.310 & 0.274 \\
            & BLOOM-1.1B     & 0.323 & 0.311 & 0.388 & 0.364 & 0.311 & 0.284 & 0.368 & 0.288 & 0.344 & 0.286 \\
            & BLOOM-1.7B     & 0.329 & 0.322 & 0.392 & 0.369 & 0.276 & 0.292 & 0.350 & 0.293 & 0.364 & 0.296 \\
            & BLOOM-3B       & 0.384 & 0.389 & 0.462 & 0.461 & 0.338 & 0.331 & 0.419 & 0.319 & 0.444 & 0.286 \\
            & BLOOM-7.1B     & 0.433 & 0.428 & 0.517 & 0.508 & 0.367 & 0.323 & 0.467 & 0.352 & 0.476 & 0.289 \\
\midrule
  OPT       & OPT-125M       & 0.293 & 0.259 & 0.290 & 0.291 & 0.293 & 0.261 & 0.281 & 0.286 & 0.269 & 0.260 \\
            & OPT-350m       & 0.282 & 0.266 & 0.311 & 0.274 & 0.290 & 0.300 & 0.281 & 0.300 & 0.298 & 0.261 \\
            & OPT-1.3B       & 0.297 & 0.270 & 0.441 & 0.337 & 0.311 & 0.279 & 0.322 & 0.292 & 0.332 & 0.270 \\
            & OPT-2.7B       & 0.286 & 0.276 & 0.452 & 0.344 & 0.306 & 0.278 & 0.313 & 0.307 & 0.326 & 0.260 \\
            & OPT-6.7B       & 0.289 & 0.294 & 0.538 & 0.378 & 0.347 & 0.271 & 0.347 & 0.308 & 0.387 & 0.272 \\
\midrule
  XGLM      & XGLM-564M      & 0.271 & 0.297 & 0.313 & 0.304 & 0.291 & 0.289 & 0.288 & 0.263 & 0.287 & 0.278 \\
            & XGLM-1.7B      & 0.278 & 0.320 & 0.338 & 0.336 & 0.312 & 0.304 & 0.322 & 0.319 & 0.338 & 0.282 \\
            & XGLM-2.9B      & 0.308 & 0.346 & 0.419 & 0.384 & 0.376 & 0.334 & 0.370 & 0.342 & 0.378 & 0.319 \\
            & XGLM-4.5B      & 0.301 & 0.299 & 0.354 & 0.348 & 0.336 & 0.326 & 0.332 & 0.349 & 0.343 & 0.279 \\
            & XGLM-7.5B      & 0.357 & 0.359 & 0.399 & 0.393 & 0.374 & 0.337 & 0.372 & 0.359 & 0.393 & 0.323 \\
\midrule
  Qwen3     & Qwen3-0.6B     & 0.357 & 0.412 & 0.488 & 0.392 & 0.374 & 0.343 & 0.401 & 0.377 & 0.407 & 0.318 \\
            & Qwen3-1.7B     & 0.444 & 0.462 & 0.584 & 0.482 & 0.466 & 0.384 & 0.457 & 0.440 & 0.468 & 0.381 \\
            & Qwen3-4B       & 0.487 & 0.521 & 0.586 & 0.506 & 0.497 & 0.430 & 0.481 & 0.474 & 0.484 & 0.419 \\
            & Qwen3-8B       & 0.487 & 0.510 & 0.611 & 0.498 & 0.501 & 0.432 & 0.503 & 0.491 & 0.499 & 0.429 \\
            & Qwen3-14B      & 0.473 & 0.498 & 0.599 & 0.471 & 0.503 & 0.460 & 0.484 & 0.507 & 0.492 & 0.418 \\
\midrule
  SmolLM3   & SmolLM3-3B     & 0.472 & 0.439 & 0.604 & 0.500 & 0.496 & 0.421 & 0.486 & 0.482 & 0.488 & 0.337 \\
\bottomrule
\end{tabular}
\caption{\textbf{Per-language accuracy on Belebele after English-only finetuning.} Languages absent for this task are omitted.}
\label{tab:app_transfer_belebele}
\end{table*}

\begin{table*}[!htbp]
\centering
\small
\setlength{\tabcolsep}{4pt}
\begin{tabular}{llcccccccc}
\toprule
\textbf{Family} & \textbf{Model} & \textbf{ar} & \textbf{zh} & \textbf{en} & \textbf{fr} & \textbf{de} & \textbf{ru} & \textbf{es} & \textbf{tr} \\
\midrule
  BLOOM     & BLOOM-560M     & 0.697 & 0.625 & 0.807 & 0.756 & 0.480 & 0.477 & 0.793 & 0.412 \\
            & BLOOM-1.1B     & 0.702 & 0.692 & 0.840 & 0.798 & 0.584 & 0.510 & 0.809 & 0.425 \\
            & BLOOM-1.7B     & 0.756 & 0.763 & 0.843 & 0.808 & 0.549 & 0.621 & 0.807 & 0.401 \\
            & BLOOM-3B       & 0.760 & 0.752 & 0.878 & 0.824 & 0.607 & 0.603 & 0.845 & 0.448 \\
            & BLOOM-7.1B     & 0.788 & 0.790 & 0.886 & 0.839 & 0.698 & 0.694 & 0.857 & 0.440 \\
\midrule
  OPT       & OPT-125M       & 0.350 & 0.338 & 0.819 & 0.537 & 0.447 & 0.334 & 0.574 & 0.375 \\
            & OPT-350m       & 0.362 & 0.371 & 0.840 & 0.485 & 0.454 & 0.389 & 0.589 & 0.374 \\
            & OPT-1.3B       & 0.356 & 0.341 & 0.878 & 0.661 & 0.598 & 0.380 & 0.686 & 0.398 \\
            & OPT-2.7B       & 0.370 & 0.346 & 0.891 & 0.677 & 0.642 & 0.398 & 0.709 & 0.352 \\
            & OPT-6.7B       & 0.344 & 0.408 & 0.908 & 0.725 & 0.610 & 0.489 & 0.722 & 0.401 \\
\midrule
  XGLM      & XGLM-564M      & 0.645 & 0.660 & 0.821 & 0.710 & 0.728 & 0.701 & 0.577 & 0.641 \\
            & XGLM-1.7B      & 0.699 & 0.714 & 0.844 & 0.739 & 0.743 & 0.751 & 0.630 & 0.679 \\
            & XGLM-2.9B      & 0.733 & 0.761 & 0.873 & 0.632 & 0.798 & 0.786 & 0.629 & 0.688 \\
            & XGLM-4.5B      & 0.714 & 0.723 & 0.880 & 0.793 & 0.794 & 0.770 & 0.698 & 0.724 \\
            & XGLM-7.5B      & 0.729 & 0.734 & 0.874 & 0.807 & 0.786 & 0.776 & 0.810 & 0.669 \\
\midrule
  Qwen3     & Qwen3-0.6B     & 0.686 & 0.685 & 0.870 & 0.742 & 0.762 & 0.672 & 0.801 & 0.633 \\
            & Qwen3-1.7B     & 0.607 & 0.743 & 0.912 & 0.814 & 0.684 & 0.708 & 0.837 & 0.461 \\
            & Qwen3-4B       & 0.780 & 0.822 & 0.918 & 0.857 & 0.845 & 0.837 & 0.863 & 0.794 \\
            & Qwen3-8B       & 0.820 & 0.844 & 0.927 & 0.857 & 0.844 & 0.821 & 0.871 & 0.706 \\
            & Qwen3-14B      & 0.846 & 0.834 & 0.921 & 0.870 & 0.856 & 0.857 & 0.874 & 0.817 \\
\midrule
  SmolLM3   & SmolLM3-3B     & 0.819 & 0.822 & 0.911 & 0.860 & 0.853 & 0.819 & 0.871 & 0.541 \\
\bottomrule
\end{tabular}
\caption{\textbf{Per-language accuracy on XNLI after English-only finetuning.} Languages absent for this task are omitted.}
\label{tab:app_transfer_xnli}
\end{table*}

\begin{table*}[!htbp]
\centering
\small
\setlength{\tabcolsep}{4pt}
\begin{tabular}{llccccccccc}
\toprule
\textbf{Family} & \textbf{Model} & \textbf{ar} & \textbf{zh} & \textbf{en} & \textbf{fr} & \textbf{de} & \textbf{ja} & \textbf{pt} & \textbf{ru} & \textbf{es} \\
\midrule
  BLOOM     & BLOOM-560M     & 0.378 & 0.493 & 0.935 & 0.559 & 0.205 & 0.236 & 0.508 & 0.210 & 0.536 \\
            & BLOOM-1.1B     & 0.418 & 0.544 & 0.940 & 0.534 & 0.212 & 0.230 & 0.517 & 0.226 & 0.564 \\
            & BLOOM-1.7B     & 0.468 & 0.581 & 0.949 & 0.595 & 0.240 & 0.269 & 0.573 & 0.242 & 0.591 \\
            & BLOOM-3B       & 0.489 & 0.567 & 0.943 & 0.628 & 0.259 & 0.254 & 0.613 & 0.239 & 0.629 \\
            & BLOOM-7.1B     & 0.539 & 0.567 & 0.958 & 0.670 & 0.299 & 0.310 & 0.658 & 0.271 & 0.671 \\
\midrule
  OPT       & OPT-125M       & 0.165 & 0.233 & 0.938 & 0.235 & 0.212 & 0.192 & 0.199 & 0.217 & 0.200 \\
            & OPT-350m       & 0.205 & 0.203 & 0.933 & 0.211 & 0.202 & 0.213 & 0.207 & 0.195 & 0.197 \\
            & OPT-1.3B       & 0.206 & 0.225 & 0.943 & 0.302 & 0.269 & 0.221 & 0.293 & 0.223 & 0.315 \\
            & OPT-2.7B       & 0.198 & 0.246 & 0.961 & 0.332 & 0.291 & 0.236 & 0.342 & 0.216 & 0.350 \\
            & OPT-6.7B       & 0.201 & 0.240 & 0.970 & 0.383 & 0.354 & 0.250 & 0.363 & 0.231 & 0.431 \\
\midrule
  XGLM      & XGLM-564M      & 0.308 & 0.443 & 0.933 & 0.461 & 0.468 & 0.347 & 0.434 & 0.445 & 0.464 \\
            & XGLM-1.7B      & 0.379 & 0.525 & 0.943 & 0.487 & 0.472 & 0.402 & 0.459 & 0.482 & 0.524 \\
            & XGLM-2.9B      & 0.448 & 0.567 & 0.955 & 0.521 & 0.525 & 0.438 & 0.516 & 0.534 & 0.556 \\
            & XGLM-4.5B      & 0.461 & 0.559 & 0.953 & 0.592 & 0.587 & 0.419 & 0.563 & 0.579 & 0.615 \\
            & XGLM-7.5B      & 0.500 & 0.582 & 0.957 & 0.602 & 0.603 & 0.441 & 0.578 & 0.588 & 0.630 \\
\midrule
  Qwen3     & Qwen3-0.6B     & 0.337 & 0.465 & 0.954 & 0.440 & 0.376 & 0.299 & 0.398 & 0.435 & 0.439 \\
            & Qwen3-1.7B     & 0.455 & 0.555 & 0.961 & 0.538 & 0.456 & 0.363 & 0.480 & 0.520 & 0.557 \\
            & Qwen3-4B       & 0.485 & 0.588 & 0.962 & 0.579 & 0.517 & 0.404 & 0.554 & 0.536 & 0.579 \\
            & Qwen3-8B       & 0.516 & 0.574 & 0.955 & 0.561 & 0.491 & 0.391 & 0.573 & 0.573 & 0.599 \\
            & Qwen3-14B      & 0.547 & 0.612 & 0.958 & 0.578 & 0.495 & 0.444 & 0.569 & 0.592 & 0.592 \\
\midrule
  SmolLM3   & SmolLM3-3B     & 0.602 & 0.604 & 0.966 & 0.677 & 0.628 & 0.438 & 0.625 & 0.644 & 0.693 \\
\bottomrule
\end{tabular}
\caption{\textbf{Per-language accuracy on XCSR after English-only finetuning.} Languages absent for this task are omitted.}
\label{tab:app_transfer_xcsr}
\end{table*}

\begin{table*}[!htbp]
\centering
\small
\setlength{\tabcolsep}{4pt}
\begin{tabular}{llcccccccccc}
\toprule
\textbf{Family} & \textbf{Model} & \textbf{ar} & \textbf{zh} & \textbf{en} & \textbf{fr} & \textbf{de} & \textbf{ja} & \textbf{pt} & \textbf{ru} & \textbf{es} & \textbf{tr} \\
\midrule
  BLOOM     & BLOOM-560M     & 0.770 & 0.725 & 0.735 & 0.750 & 0.451 & 0.632 & 0.701 & 0.338 & 0.721 & 0.260 \\
            & BLOOM-1.1B     & 0.784 & 0.760 & 0.897 & 0.819 & 0.417 & 0.431 & 0.647 & 0.221 & 0.681 & 0.162 \\
            & BLOOM-1.7B     & 0.745 & 0.407 & 0.848 & 0.647 & 0.495 & 0.446 & 0.510 & 0.275 & 0.559 & 0.142 \\
            & BLOOM-3B       & 0.863 & 0.902 & 0.907 & 0.843 & 0.696 & 0.725 & 0.824 & 0.534 & 0.868 & 0.466 \\
            & BLOOM-7.1B     & 0.828 & 0.868 & 0.848 & 0.804 & 0.770 & 0.799 & 0.848 & 0.632 & 0.858 & 0.534 \\
\midrule
  OPT       & OPT-125M       & 0.142 & 0.304 & 0.848 & 0.613 & 0.539 & 0.299 & 0.529 & 0.289 & 0.559 & 0.382 \\
            & OPT-350m       & 0.137 & 0.279 & 0.892 & 0.618 & 0.544 & 0.299 & 0.564 & 0.279 & 0.623 & 0.368 \\
            & OPT-1.3B       & 0.255 & 0.387 & 0.882 & 0.814 & 0.804 & 0.373 & 0.824 & 0.284 & 0.819 & 0.603 \\
            & OPT-2.7B       & 0.270 & 0.578 & 0.882 & 0.828 & 0.814 & 0.456 & 0.828 & 0.304 & 0.848 & 0.593 \\
            & OPT-6.7B       & 0.279 & 0.701 & 0.873 & 0.706 & 0.716 & 0.696 & 0.760 & 0.578 & 0.770 & 0.667 \\
\midrule
  XGLM      & XGLM-564M      & 0.711 & 0.740 & 0.843 & 0.627 & 0.755 & 0.765 & 0.721 & 0.716 & 0.593 & 0.309 \\
            & XGLM-1.7B      & 0.828 & 0.868 & 0.902 & 0.819 & 0.892 & 0.853 & 0.828 & 0.843 & 0.765 & 0.676 \\
            & XGLM-2.9B      & 0.873 & 0.912 & 0.902 & 0.863 & 0.902 & 0.897 & 0.853 & 0.912 & 0.863 & 0.858 \\
            & XGLM-4.5B      & 0.804 & 0.838 & 0.877 & 0.760 & 0.765 & 0.877 & 0.593 & 0.843 & 0.588 & 0.657 \\
            & XGLM-7.5B      & 0.804 & 0.843 & 0.882 & 0.804 & 0.873 & 0.853 & 0.848 & 0.853 & 0.863 & 0.833 \\
\midrule
  Qwen3     & Qwen3-0.6B     & 0.775 & 0.838 & 0.853 & 0.608 & 0.824 & 0.843 & 0.824 & 0.848 & 0.824 & 0.799 \\
            & Qwen3-1.7B     & 0.838 & 0.838 & 0.853 & 0.735 & 0.784 & 0.814 & 0.814 & 0.863 & 0.828 & 0.779 \\
            & Qwen3-4B       & 0.750 & 0.868 & 0.907 & 0.716 & 0.838 & 0.779 & 0.828 & 0.833 & 0.868 & 0.843 \\
            & Qwen3-8B       & 0.794 & 0.882 & 0.882 & 0.745 & 0.892 & 0.848 & 0.882 & 0.868 & 0.902 & 0.863 \\
            & Qwen3-14B      & 0.765 & 0.892 & 0.882 & 0.755 & 0.858 & 0.897 & 0.897 & 0.863 & 0.882 & 0.833 \\
\midrule
  SmolLM3   & SmolLM3-3B     & 0.706 & 0.824 & 0.907 & 0.750 & 0.892 & 0.779 & 0.848 & 0.863 & 0.882 & 0.662 \\
\bottomrule
\end{tabular}
\caption{\textbf{Per-language accuracy on SIB-200 after English-only finetuning.} Languages absent for this task are omitted.}
\label{tab:app_transfer_sib200}
\end{table*}

\begin{table*}[!htbp]
\centering
\small
\setlength{\tabcolsep}{4pt}
\begin{tabular}{llccccccc}
\toprule
\textbf{Family} & \textbf{Model} & \textbf{ar} & \textbf{zh} & \textbf{en} & \textbf{de} & \textbf{ru} & \textbf{es} & \textbf{tr} \\
\midrule
  BLOOM     & BLOOM-560M     & 0.398 & 0.402 & 0.751 & 0.235 & 0.243 & 0.628 & 0.099 \\
            & BLOOM-1.1B     & 0.555 & 0.443 & 0.768 & 0.404 & 0.318 & 0.713 & 0.149 \\
            & BLOOM-1.7B     & 0.353 & 0.413 & 0.799 & 0.439 & 0.354 & 0.706 & 0.096 \\
            & BLOOM-3B       & 0.481 & 0.429 & 0.820 & 0.513 & 0.453 & 0.747 & 0.175 \\
            & BLOOM-7.1B     & 0.682 & 0.503 & 0.823 & 0.630 & 0.546 & 0.776 & 0.349 \\
\midrule
  OPT       & OPT-125M       & 0.051 & 0.058 & 0.625 & 0.053 & 0.047 & 0.179 & 0.044 \\
            & OPT-350m       & 0.035 & 0.069 & 0.610 & 0.122 & 0.021 & 0.246 & 0.052 \\
            & OPT-1.3B       & 0.050 & 0.077 & 0.595 & 0.309 & 0.045 & 0.332 & 0.114 \\
            & OPT-2.7B       & 0.060 & 0.073 & 0.802 & 0.346 & 0.061 & 0.583 & 0.103 \\
            & OPT-6.7B       & 0.048 & 0.087 & 0.785 & 0.398 & 0.149 & 0.516 & 0.166 \\
\midrule
  XGLM      & XGLM-564M      & 0.516 & 0.509 & 0.717 & 0.576 & 0.563 & 0.622 & 0.434 \\
            & XGLM-1.7B      & 0.277 & 0.453 & 0.383 & 0.346 & 0.320 & 0.394 & 0.319 \\
            & XGLM-2.9B      & 0.313 & 0.496 & 0.443 & 0.363 & 0.356 & 0.396 & 0.359 \\
            & XGLM-4.5B      & 0.322 & 0.497 & 0.527 & 0.402 & 0.399 & 0.477 & 0.371 \\
            & XGLM-7.5B      & 0.366 & 0.535 & 0.583 & 0.479 & 0.455 & 0.538 & 0.465 \\
\midrule
  Qwen3     & Qwen3-0.6B     & 0.501 & 0.403 & 0.706 & 0.576 & 0.585 & 0.646 & 0.402 \\
            & Qwen3-1.7B     & 0.615 & 0.526 & 0.775 & 0.673 & 0.679 & 0.695 & 0.518 \\
            & Qwen3-4B       & 0.692 & 0.562 & 0.793 & 0.720 & 0.691 & 0.741 & 0.641 \\
            & Qwen3-8B       & 0.683 & 0.431 & 0.599 & 0.673 & 0.688 & 0.722 & 0.607 \\
            & Qwen3-14B      & 0.730 & 0.538 & 0.636 & 0.755 & 0.727 & 0.732 & 0.656 \\
\midrule
  SmolLM3   & SmolLM3-3B     & 0.632 & 0.465 & 0.585 & 0.640 & 0.635 & 0.698 & 0.389 \\
\bottomrule
\end{tabular}
\caption{\textbf{Per-language F1 on XQuAD after English-only finetuning.} Languages absent for this task are omitted.}
\label{tab:app_transfer_xquad}
\end{table*}

\FloatBarrier

\subsection{Training Dynamics}
\label{sec:app_training_dynamics}

\begin{figure*}[!htbp]
\centering
\includegraphics[width=\textwidth]{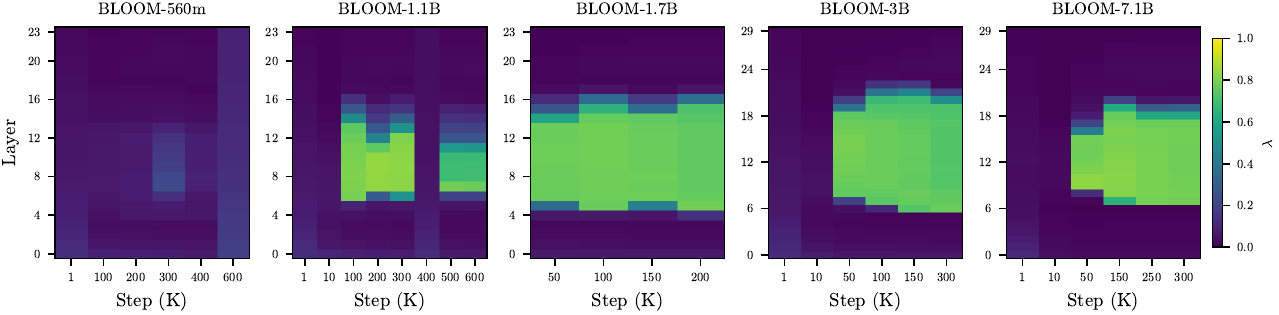}
\caption{\textbf{BLOOM training dynamics: sampled $\lambda$} per layer (vertical) across intermediate checkpoints (horizontal). Columns at non-equal step intervals reflect the available BLOOM checkpoints after excluding duplicate uploads.}
\label{fig:app_td_bloom_lambda}
\end{figure*}

\begin{figure*}[!htbp]
\centering
\includegraphics[width=\textwidth]{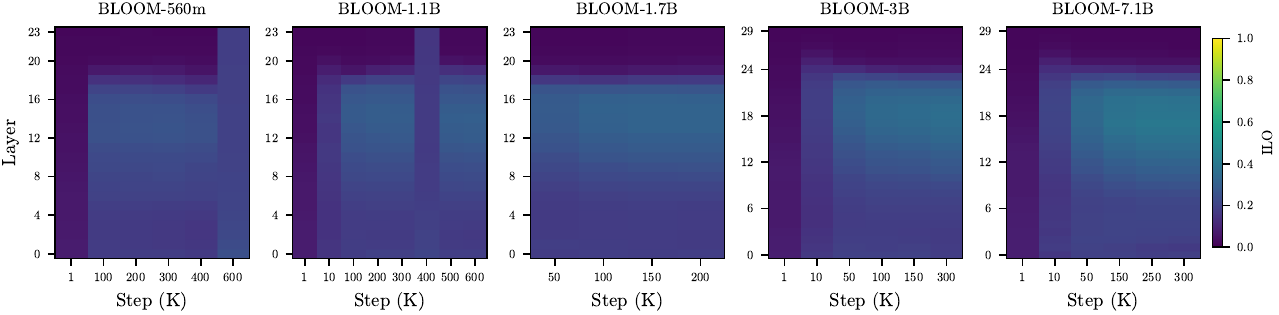}
\caption{\textbf{BLOOM training dynamics: ILO} (word-level). Layout as in Fig.~\ref{fig:app_td_bloom_lambda}.}
\label{fig:app_td_bloom_ilo}
\end{figure*}

\begin{figure*}[!htbp]
\centering
\includegraphics[width=\textwidth]{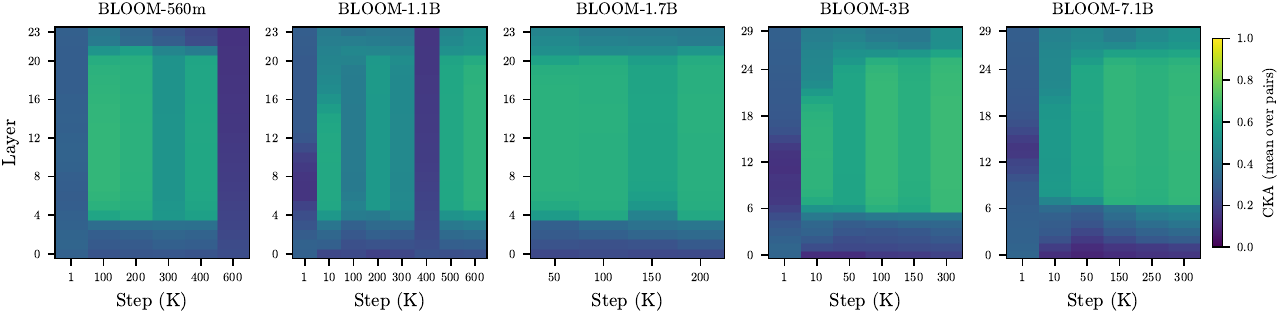}
\caption{\textbf{BLOOM training dynamics: CKA} averaged across language pairs. Layout as in Fig.~\ref{fig:app_td_bloom_lambda}.}
\label{fig:app_td_bloom_cka}
\end{figure*}

\begin{figure*}[!htbp]
\centering
\includegraphics[width=\textwidth]{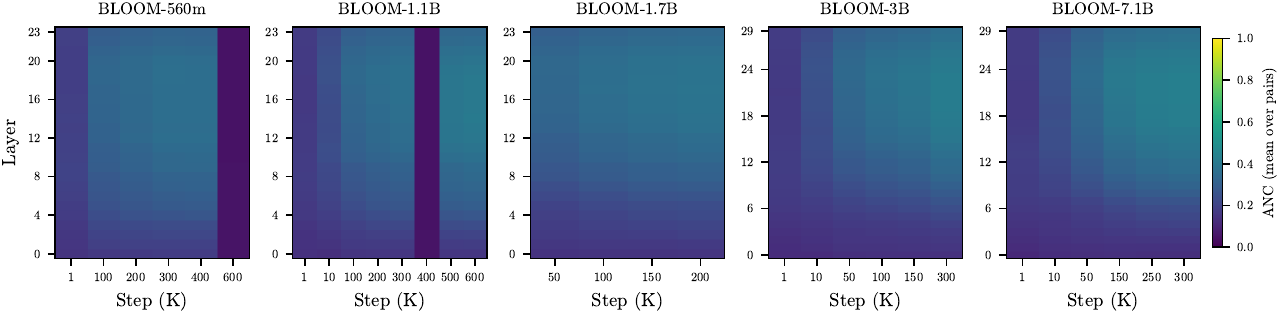}
\caption{\textbf{BLOOM training dynamics: ANC} averaged across language pairs. Layout as in Fig.~\ref{fig:app_td_bloom_lambda}.}
\label{fig:app_td_bloom_anc}
\end{figure*}

\begin{figure*}[!htbp]
\centering
\includegraphics[width=\textwidth]{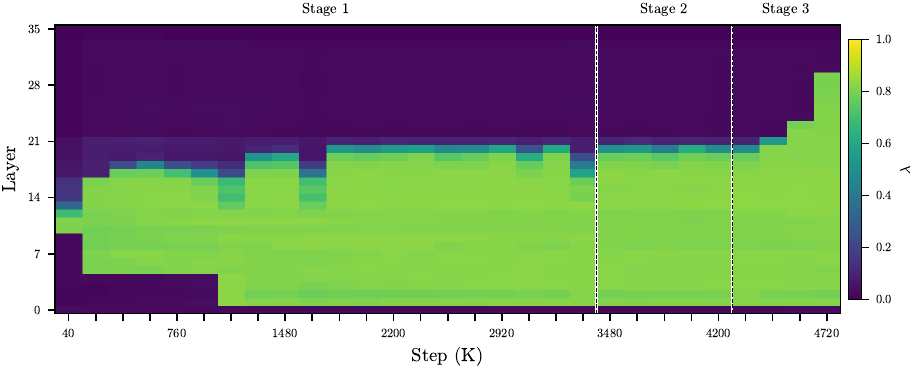}
\caption{\textbf{SmolLM3-3B training dynamics: sampled $\lambda$.} Vertical dashed lines mark the Stage~1$\to$2 and Stage~2$\to$3 boundaries.}
\label{fig:app_td_smollm3_lambda}
\end{figure*}

\begin{figure*}[!htbp]
\centering
\includegraphics[width=\textwidth]{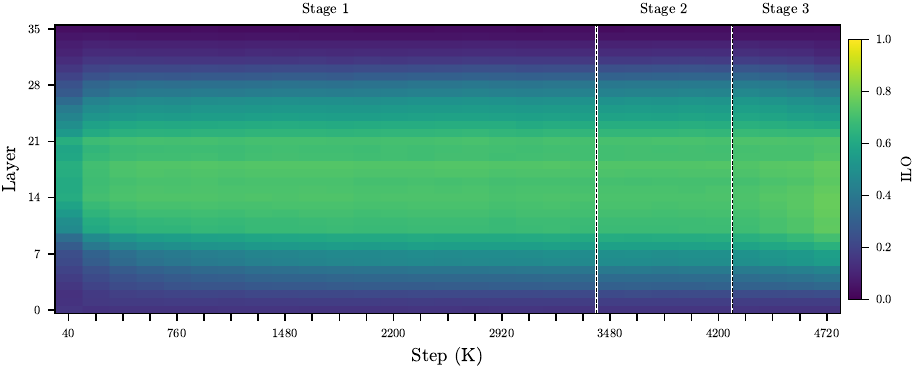}
\caption{\textbf{SmolLM3-3B training dynamics: ILO.} Layout as in Fig.~\ref{fig:app_td_smollm3_lambda}.}
\label{fig:app_td_smollm3_ilo}
\end{figure*}

\begin{figure*}[!htbp]
\centering
\includegraphics[width=\textwidth]{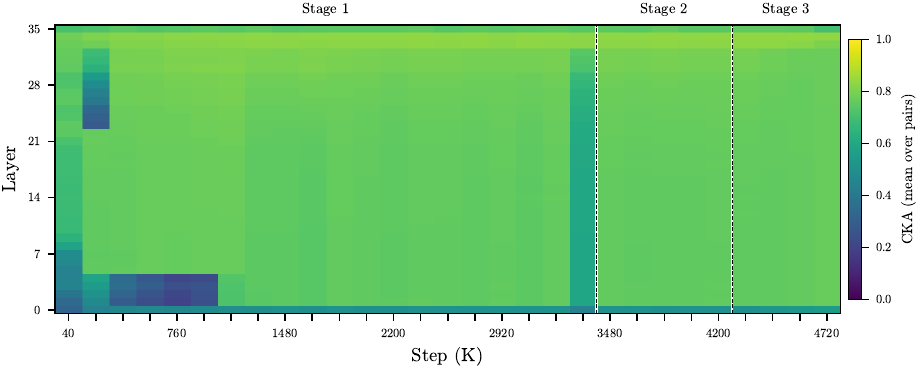}
\caption{\textbf{SmolLM3-3B training dynamics: CKA} averaged across language pairs. Layout as in Fig.~\ref{fig:app_td_smollm3_lambda}.}
\label{fig:app_td_smollm3_cka}
\end{figure*}

\begin{figure*}[!htbp]
\centering
\includegraphics[width=\textwidth]{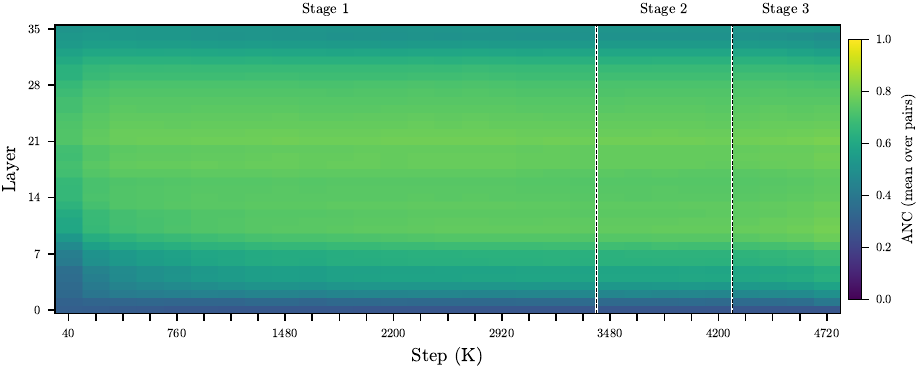}
\caption{\textbf{SmolLM3-3B training dynamics: ANC} averaged across language pairs. Layout as in Fig.~\ref{fig:app_td_smollm3_lambda}.}
\label{fig:app_td_smollm3_anc}
\end{figure*}

We trace each sharing metric across intermediate checkpoints for BLOOM (30 valid checkpoints across 5 sizes; duplicate weight uploads excluded as listed in Table~\ref{tab:app_checkpoints}) and SmolLM3-3B-Base (29 checkpoints spanning the three documented training stages). Figures~\ref{fig:app_td_bloom_lambda}--\ref{fig:app_td_smollm3_anc} render each (model size $\times$ metric) as a layer $\times$ checkpoint heatmap. For SmolLM3, vertical dashed lines mark the Stage~1 $\to$ Stage~2 boundary (around step 3.46M) and the Stage~2 $\to$ Stage~3 boundary (around step 4.22M).

The $\lambda$ heatmaps make the size threshold concrete: BLOOM-560M never develops a high-$\lambda$ band, BLOOM-1.1B undergoes a phase transition between step $10$K and $100$K, and the three larger BLOOM models already exhibit a mid-layer plateau at their earliest sampled checkpoint. The plateau's vertical extent grows with model size. SmolLM3 establishes a high-$\lambda$ band (layers $\approx$0--21) by its first checkpoint at step 40K; Stage 3 visibly extends the band into the upper layers (up to layer 29 in the final checkpoint). ILO follows a similar trajectory to $\lambda$ in BLOOM but rises more gradually. CKA and ANC by contrast already register substantial similarity in middle layers from the earliest checkpoints---including BLOOM-560M where $\lambda$ never crosses threshold---reflecting the anisotropy build-up reported in \S\ref{sec:why_disagree} rather than a sharing transition.

\FloatBarrier

\subsection{Instruction Tuning Effects}
\label{sec:app_instruction_effects}

\begin{figure*}[!htbp]
\centering
\includegraphics[width=\textwidth]{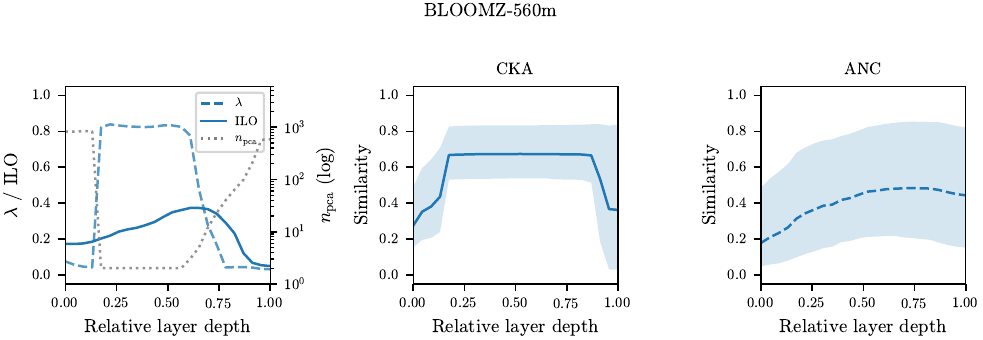}
\caption{\textbf{BLOOMZ-560m.} Panels as in Fig.~\ref{fig:app_metrics_BLOOM-560M}.}
\label{fig:app_it_bloomz-560m}
\end{figure*}

\begin{figure*}[!htbp]
\centering
\includegraphics[width=\textwidth]{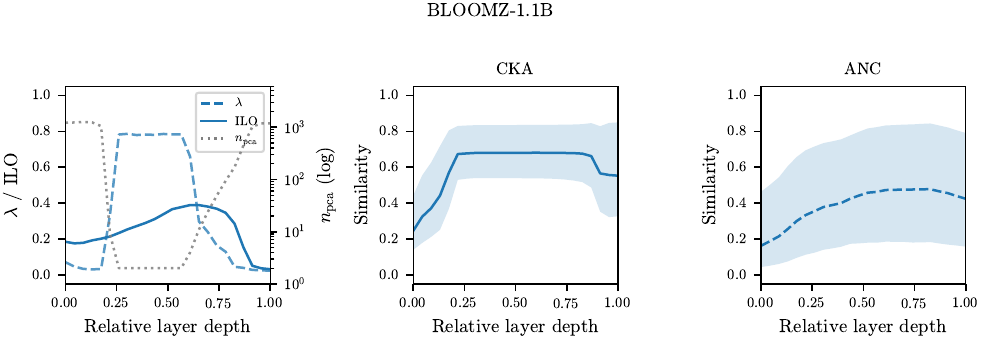}
\caption{\textbf{BLOOMZ-1.1B.} Panels as in Fig.~\ref{fig:app_metrics_BLOOM-560M}.}
\label{fig:app_it_bloomz-1-1b}
\end{figure*}

\begin{figure*}[!htbp]
\centering
\includegraphics[width=\textwidth]{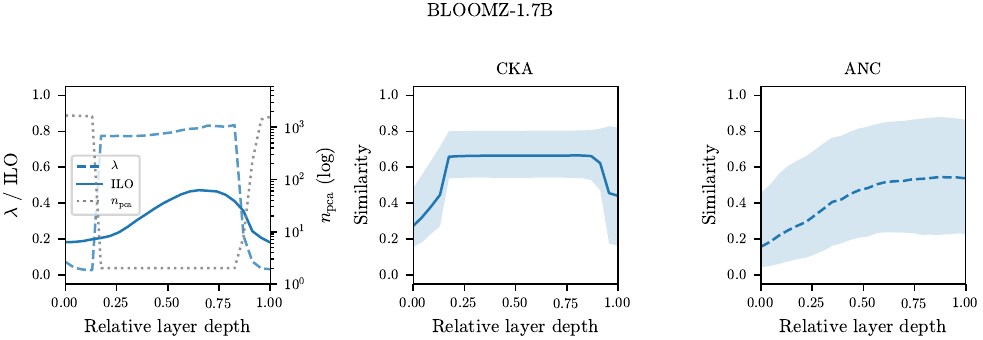}
\caption{\textbf{BLOOMZ-1.7B.} Panels as in Fig.~\ref{fig:app_metrics_BLOOM-560M}.}
\label{fig:app_it_bloomz-1-7b}
\end{figure*}

\begin{figure*}[!htbp]
\centering
\includegraphics[width=\textwidth]{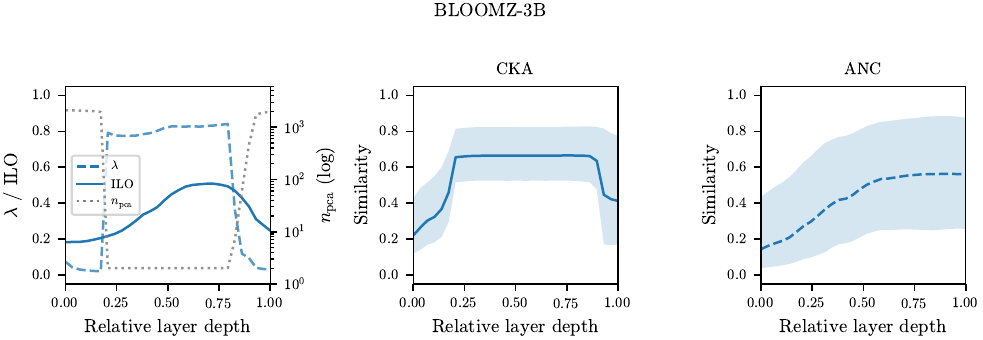}
\caption{\textbf{BLOOMZ-3B.} Panels as in Fig.~\ref{fig:app_metrics_BLOOM-560M}.}
\label{fig:app_it_bloomz-3b}
\end{figure*}

\begin{figure*}[!htbp]
\centering
\includegraphics[width=\textwidth]{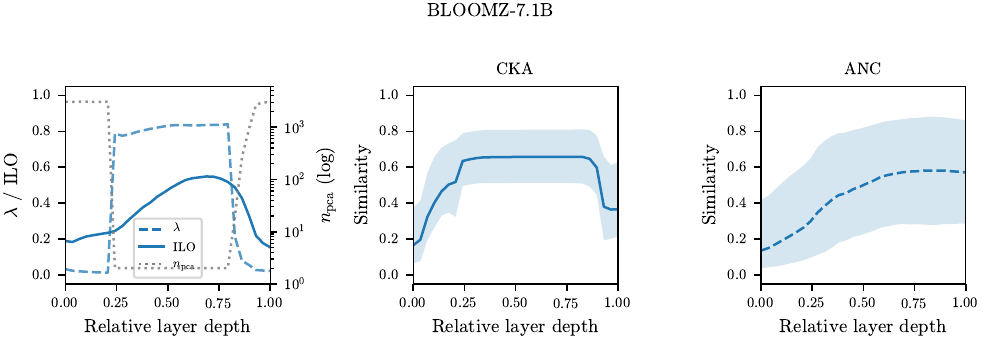}
\caption{\textbf{BLOOMZ-7.1B.} Panels as in Fig.~\ref{fig:app_metrics_BLOOM-560M}.}
\label{fig:app_it_bloomz-7-1b}
\end{figure*}

\begin{figure*}[!htbp]
\centering
\includegraphics[width=\textwidth]{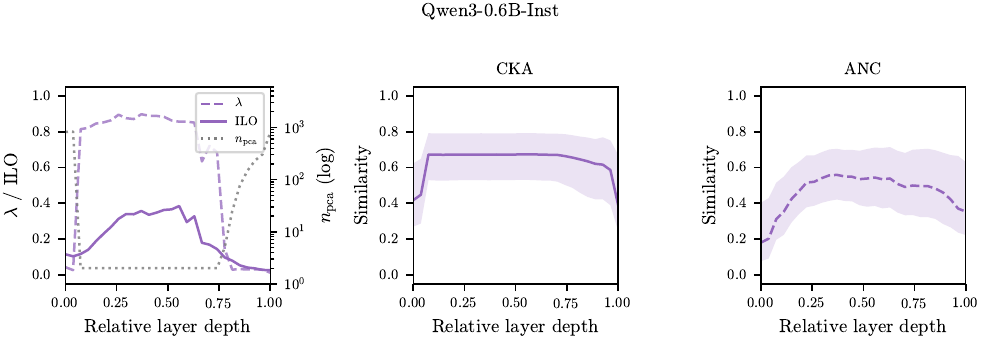}
\caption{\textbf{Qwen3-0.6B-Instruct.} Panels as in Fig.~\ref{fig:app_metrics_BLOOM-560M}.}
\label{fig:app_it_qwen3-0-6b-inst}
\end{figure*}

\begin{figure*}[!htbp]
\centering
\includegraphics[width=\textwidth]{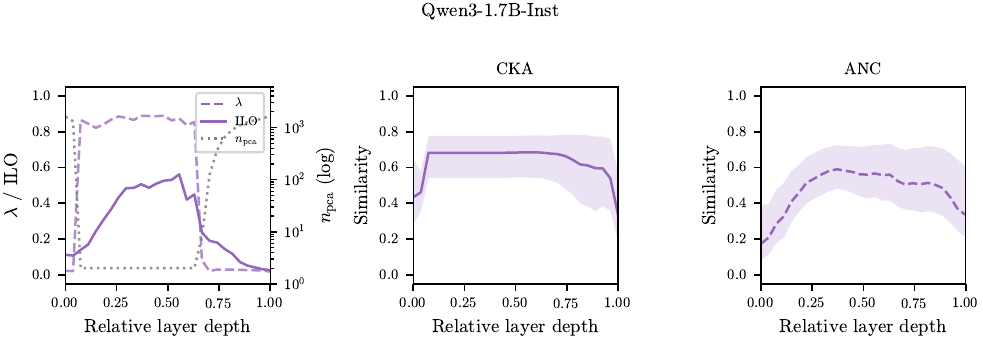}
\caption{\textbf{Qwen3-1.7B-Instruct.} Panels as in Fig.~\ref{fig:app_metrics_BLOOM-560M}.}
\label{fig:app_it_qwen3-1-7b-inst}
\end{figure*}

\begin{figure*}[!htbp]
\centering
\includegraphics[width=\textwidth]{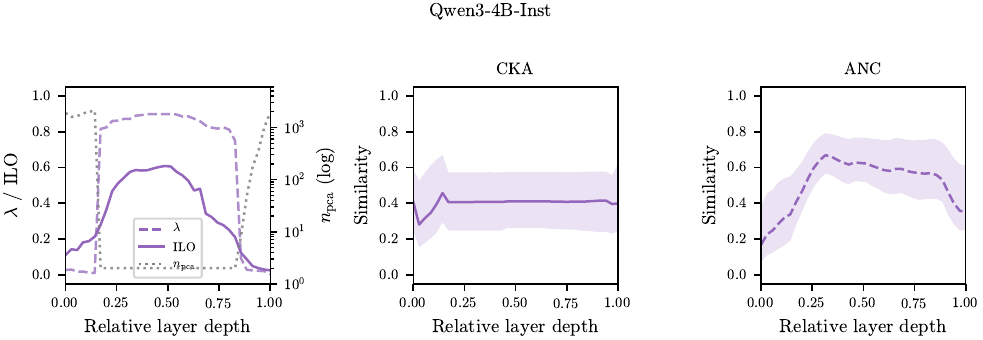}
\caption{\textbf{Qwen3-4B-Instruct.} Panels as in Fig.~\ref{fig:app_metrics_BLOOM-560M}.}
\label{fig:app_it_qwen3-4b-inst}
\end{figure*}

\begin{figure*}[!htbp]
\centering
\includegraphics[width=\textwidth]{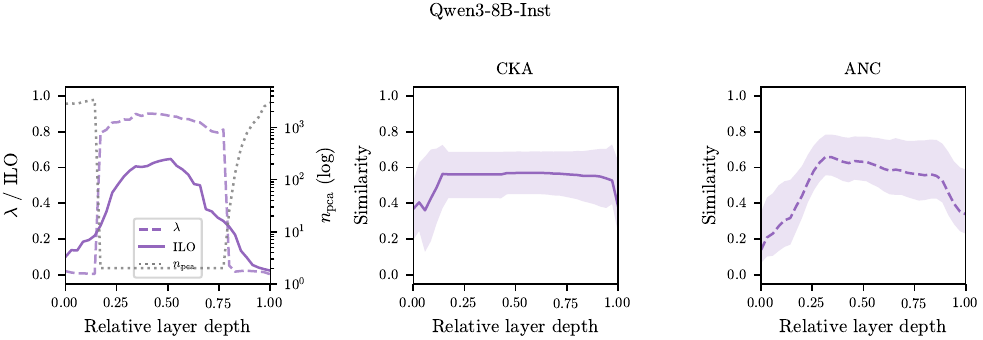}
\caption{\textbf{Qwen3-8B-Instruct.} Panels as in Fig.~\ref{fig:app_metrics_BLOOM-560M}.}
\label{fig:app_it_qwen3-8b-inst}
\end{figure*}

\begin{figure*}[!htbp]
\centering
\includegraphics[width=\textwidth]{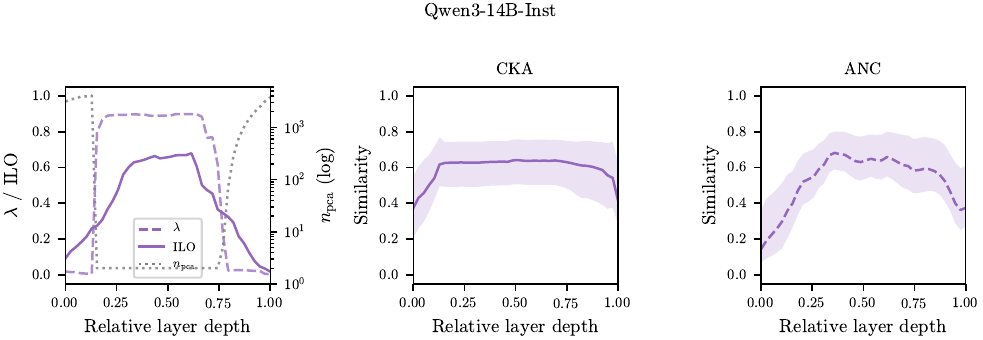}
\caption{\textbf{Qwen3-14B-Instruct.} Panels as in Fig.~\ref{fig:app_metrics_BLOOM-560M}.}
\label{fig:app_it_qwen3-14b-inst}
\end{figure*}

\begin{figure*}[!htbp]
\centering
\includegraphics[width=\textwidth]{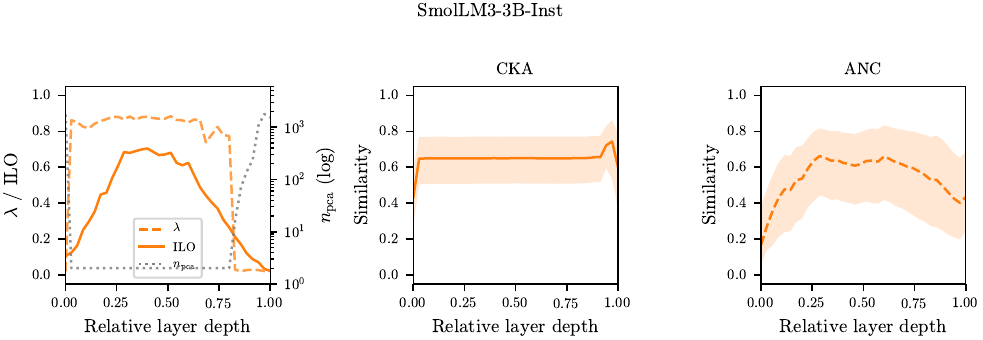}
\caption{\textbf{SmolLM3-3B-Instruct.} Panels as in Fig.~\ref{fig:app_metrics_BLOOM-560M}.}
\label{fig:app_it_smollm3-3b-inst}
\end{figure*}

Figures~\ref{fig:app_it_bloomz-560m}--\ref{fig:app_it_smollm3-3b-inst} replicate the 3-panel layout of \S\ref{sec:app_fullresults} for the instruction-tuned variants we evaluate: BLOOMZ (560m--7.1B), Qwen3-Instruct (0.6B--14B), and SmolLM3-3B-Instruct. The base counterparts are in Figures~\ref{fig:app_metrics_BLOOM-560M}--\ref{fig:app_metrics_qwen3-14b} for direct comparison.

Effect sizes vary sharply by family. In BLOOMZ the largest shift is at the smallest scale: BLOOMZ-560m moves from a weak mid-layer peak ($\lambda = 0.21$ in base) to a full sharing plateau ($\lambda = 0.84$), with ILO rising in parallel ($0.25 \to 0.37$). The remaining BLOOM sizes already exhibit sharing in the base model and gain only modest peak-$\lambda$ increments ($+0.02$ to $+0.07$); peak ILO rises more substantially across all BLOOM sizes ($+0.09$ to $+0.15$). Qwen3-Instruct and SmolLM3-3B-Instruct show negligible changes in either metric (peak-$\lambda$ within $\pm 0.03$, peak-ILO within $\pm 0.03$), consistent with their base models already operating near the upper end of the sharing range. CKA and ANC profiles change only marginally across all families.

\FloatBarrier

\subsection{Finetuning Effects}
\label{sec:app_finetuning_effects}

We re-applied the four sharing metrics to each finetuned model at the best learning rate from Appendix~\ref{sec:app_xlang_finetuning}, with peak layer fixed at the base-model peak. Table~\ref{tab:app_ft_summary} summarizes the resulting shifts and Tables~\ref{tab:app_ft_belebele}--\ref{tab:app_ft_xquad} report the per-task breakdown. The analysis covers all 21 models $\times$ 5 tasks = 105 (model, task) pairs.

\begin{table}[h]
\centering
\small
\begin{tabular}{lcccc}
\toprule
 & $\lambda$ & ILO & CKA & ANC \\
\midrule
Median $|\Delta|$            & 0.003 & 0.021 & 0.009 & 0.010 \\
P90 $|\Delta|$               & 0.035 & 0.119 & 0.171 & 0.113 \\
\# pairs $|\Delta|\!>\!0.05$ & 8     & 31    & 24    & 19    \\
\# pairs $|\Delta|\!>\!0.10$ & 4     & 14    & 16    & 12    \\
\bottomrule
\end{tabular}
\caption{Distribution of peak-metric shifts after cross-lingual finetuning across 105 (model, task) pairs.}
\label{tab:app_ft_summary}
\end{table}

\begin{table}[h]
\centering
\small
\begin{tabular}{lrrrr}
\toprule
Model & $\Delta\lambda$ & $\Delta$ILO & $\Delta$CKA & $\Delta$ANC \\
\midrule
\multicolumn{5}{l}{\emph{BLOOM}} \\
BLOOM-560M & +0.068 & +0.022 & +0.002 & +0.003 \\
BLOOM-1b1 & +0.023 & +0.035 & +0.001 & +0.007 \\
BLOOM-1b7 & +0.016 & +0.030 & $-$0.000 & +0.006 \\
BLOOM-3B & +0.025 & +0.049 & +0.001 & +0.016 \\
BLOOM-7b1 & +0.016 & +0.042 & $-$0.002 & +0.004 \\
\midrule
\multicolumn{5}{l}{\emph{OPT}} \\
OPT-125M & +0.000 & +0.000 & $-$0.000 & $-$0.002 \\
OPT-350M & $-$0.001 & +0.000 & +0.005 & $-$0.003 \\
OPT-1.3B & +0.004 & +0.016 & +0.009 & $-$0.009 \\
OPT-2.7B & $-$0.001 & +0.006 & +0.037 & $-$0.020 \\
OPT-6.7B & +0.001 & +0.011 & +0.014 & $-$0.008 \\
\midrule
\multicolumn{5}{l}{\emph{XGLM}} \\
XGLM-564M & +0.000 & $-$0.003 & $-$0.001 & +0.002 \\
XGLM-1.7B & +0.000 & +0.124 & $-$0.249 & +0.113 \\
XGLM-2.9B & +0.001 & +0.127 & $-$0.067 & +0.002 \\
XGLM-4.5B & +0.000 & $-$0.115 & $-$0.033 & $-$0.162 \\
XGLM-7.5B & $-$0.002 & +0.288 & $-$0.199 & $-$0.007 \\
\midrule
\multicolumn{5}{l}{\emph{Qwen3}} \\
Qwen3-0.6B & +0.006 & +0.102 & +0.000 & +0.011 \\
Qwen3-1.7B & $-$0.008 & +0.075 & +0.003 & +0.016 \\
Qwen3-4B & +0.001 & +0.011 & $-$0.015 & $-$0.004 \\
Qwen3-8B & +0.001 & +0.012 & $-$0.042 & +0.001 \\
Qwen3-14B & $-$0.001 & +0.009 & $-$0.000 & $-$0.006 \\
\midrule
\multicolumn{5}{l}{\emph{SmolLM3}} \\
SmolLM3-3B & +0.003 & +0.013 & $-$0.005 & +0.001 \\
\bottomrule
\end{tabular}
\caption{\textbf{Peak-metric shifts after finetuning on Belebele.} Positive values indicate the finetuned model's peak metric exceeds the base model's; the peak layer is read independently per metric on the base model.}
\label{tab:app_ft_belebele}
\end{table}

\begin{table}[h]
\centering
\small
\begin{tabular}{lrrrr}
\toprule
Model & $\Delta\lambda$ & $\Delta$ILO & $\Delta$CKA & $\Delta$ANC \\
\midrule
\multicolumn{5}{l}{\emph{BLOOM}} \\
BLOOM-560M & +0.073 & $-$0.021 & $-$0.343 & $-$0.034 \\
BLOOM-1b1 & +0.003 & +0.003 & $-$0.000 & +0.001 \\
BLOOM-1b7 & +0.001 & +0.001 & $-$0.001 & +0.006 \\
BLOOM-3B & +0.006 & +0.005 & +0.001 & $-$0.000 \\
BLOOM-7b1 & +0.006 & +0.010 & $-$0.009 & +0.015 \\
\midrule
\multicolumn{5}{l}{\emph{OPT}} \\
OPT-125M & $-$0.001 & $-$0.001 & $-$0.001 & $-$0.003 \\
OPT-350M & $-$0.000 & $-$0.000 & +0.001 & +0.006 \\
OPT-1.3B & $-$0.001 & $-$0.000 & +0.013 & $-$0.008 \\
OPT-2.7B & $-$0.002 & $-$0.005 & +0.018 & $-$0.003 \\
OPT-6.7B & $-$0.002 & $-$0.005 & +0.008 & +0.002 \\
\midrule
\multicolumn{5}{l}{\emph{XGLM}} \\
XGLM-564M & $-$0.001 & $-$0.012 & +0.001 & $-$0.042 \\
XGLM-1.7B & +0.000 & +0.058 & $-$0.022 & +0.134 \\
XGLM-2.9B & $-$0.001 & +0.042 & $-$0.214 & $-$0.023 \\
XGLM-4.5B & $-$0.000 & +0.048 & $-$0.009 & $-$0.041 \\
XGLM-7.5B & $-$0.001 & +0.208 & $-$0.284 & $-$0.016 \\
\midrule
\multicolumn{5}{l}{\emph{Qwen3}} \\
Qwen3-0.6B & +0.002 & +0.009 & +0.000 & +0.008 \\
Qwen3-1.7B & +0.000 & +0.010 & +0.000 & +0.003 \\
Qwen3-4B & $-$0.001 & +0.000 & $-$0.001 & +0.000 \\
Qwen3-8B & $-$0.002 & +0.004 & +0.027 & +0.005 \\
Qwen3-14B & $-$0.000 & $-$0.004 & +0.000 & $-$0.002 \\
\midrule
\multicolumn{5}{l}{\emph{SmolLM3}} \\
SmolLM3-3B & $-$0.000 & $-$0.001 & +0.001 & $-$0.000 \\
\bottomrule
\end{tabular}
\caption{\textbf{Peak-metric shifts after finetuning on SIB-200.} Columns and conventions as in Table~\ref{tab:app_ft_belebele}.}
\label{tab:app_ft_sib200}
\end{table}

\begin{table}[h]
\centering
\small
\begin{tabular}{lrrrr}
\toprule
Model & $\Delta\lambda$ & $\Delta$ILO & $\Delta$CKA & $\Delta$ANC \\
\midrule
\multicolumn{5}{l}{\emph{BLOOM}} \\
BLOOM-560M & +0.051 & +0.072 & $-$0.003 & +0.013 \\
BLOOM-1b1 & $-$0.308 & +0.106 & $-$0.015 & +0.033 \\
BLOOM-1b7 & +0.029 & +0.094 & $-$0.023 & $-$0.021 \\
BLOOM-3B & +0.033 & +0.066 & $-$0.004 & $-$0.005 \\
BLOOM-7b1 & +0.013 & +0.074 & $-$0.022 & $-$0.014 \\
\midrule
\multicolumn{5}{l}{\emph{OPT}} \\
OPT-125M & +0.001 & $-$0.000 & $-$0.001 & +0.004 \\
OPT-350M & $-$0.008 & +0.000 & +0.006 & +0.005 \\
OPT-1.3B & +0.006 & +0.004 & $-$0.061 & $-$0.091 \\
OPT-2.7B & +0.022 & $-$0.012 & $-$0.058 & $-$0.143 \\
OPT-6.7B & +0.002 & $-$0.007 & $-$0.001 & $-$0.052 \\
\midrule
\multicolumn{5}{l}{\emph{XGLM}} \\
XGLM-564M & $-$0.000 & +0.077 & $-$0.088 & $-$0.173 \\
XGLM-1.7B & +0.008 & +0.223 & $-$0.171 & +0.071 \\
XGLM-2.9B & +0.009 & +0.175 & $-$0.093 & $-$0.008 \\
XGLM-4.5B & +0.008 & +0.229 & $-$0.056 & +0.080 \\
XGLM-7.5B & $-$0.001 & +0.293 & $-$0.206 & +0.003 \\
\midrule
\multicolumn{5}{l}{\emph{Qwen3}} \\
Qwen3-0.6B & +0.012 & +0.119 & +0.000 & +0.012 \\
Qwen3-1.7B & +0.008 & +0.074 & +0.001 & +0.028 \\
Qwen3-4B & +0.004 & +0.072 & $-$0.117 & $-$0.003 \\
Qwen3-8B & $-$0.003 & +0.064 & $-$0.036 & $-$0.157 \\
Qwen3-14B & $-$0.006 & +0.019 & $-$0.532 & $-$0.437 \\
\midrule
\multicolumn{5}{l}{\emph{SmolLM3}} \\
SmolLM3-3B & +0.003 & +0.023 & $-$0.021 & $-$0.004 \\
\bottomrule
\end{tabular}
\caption{\textbf{Peak-metric shifts after finetuning on XNLI.} Columns and conventions as in Table~\ref{tab:app_ft_belebele}.}
\label{tab:app_ft_xnli}
\end{table}

\begin{table}[h]
\centering
\small
\begin{tabular}{lrrrr}
\toprule
Model & $\Delta\lambda$ & $\Delta$ILO & $\Delta$CKA & $\Delta$ANC \\
\midrule
\multicolumn{5}{l}{\emph{BLOOM}} \\
BLOOM-560M & +0.272 & +0.042 & +0.001 & +0.014 \\
BLOOM-1b1 & +0.048 & +0.055 & $-$0.041 & $-$0.004 \\
BLOOM-1b7 & +0.051 & +0.053 & $-$0.015 & +0.012 \\
BLOOM-3B & +0.047 & +0.041 & $-$0.018 & +0.001 \\
BLOOM-7b1 & +0.013 & +0.025 & $-$0.006 & $-$0.008 \\
\midrule
\multicolumn{5}{l}{\emph{OPT}} \\
OPT-125M & $-$0.002 & +0.000 & $-$0.003 & $-$0.044 \\
OPT-350M & $-$0.000 & +0.000 & $-$0.001 & $-$0.014 \\
OPT-1.3B & $-$0.000 & $-$0.000 & +0.010 & $-$0.024 \\
OPT-2.7B & $-$0.001 & $-$0.008 & +0.027 & $-$0.026 \\
OPT-6.7B & +0.006 & $-$0.014 & +0.007 & $-$0.026 \\
\midrule
\multicolumn{5}{l}{\emph{XGLM}} \\
XGLM-564M & $-$0.003 & +0.002 & $-$0.023 & $-$0.208 \\
XGLM-1.7B & $-$0.001 & +0.097 & $-$0.098 & +0.155 \\
XGLM-2.9B & $-$0.002 & +0.070 & $-$0.168 & $-$0.026 \\
XGLM-4.5B & +0.006 & +0.084 & $-$0.053 & +0.056 \\
XGLM-7.5B & $-$0.001 & +0.236 & $-$0.192 & $-$0.021 \\
\midrule
\multicolumn{5}{l}{\emph{Qwen3}} \\
Qwen3-0.6B & $-$0.003 & +0.064 & $-$0.001 & +0.001 \\
Qwen3-1.7B & $-$0.006 & +0.066 & +0.001 & $-$0.007 \\
Qwen3-4B & $-$0.007 & +0.009 & $-$0.115 & $-$0.005 \\
Qwen3-8B & $-$0.002 & +0.013 & $-$0.026 & $-$0.028 \\
Qwen3-14B & +0.001 & +0.009 & $-$0.009 & +0.009 \\
\midrule
\multicolumn{5}{l}{\emph{SmolLM3}} \\
SmolLM3-3B & $-$0.003 & $-$0.011 & $-$0.009 & $-$0.010 \\
\bottomrule
\end{tabular}
\caption{\textbf{Peak-metric shifts after finetuning on XCSR.} Columns and conventions as in Table~\ref{tab:app_ft_belebele}.}
\label{tab:app_ft_xcsr}
\end{table}

\begin{table}[h]
\centering
\small
\begin{tabular}{lrrrr}
\toprule
Model & $\Delta\lambda$ & $\Delta$ILO & $\Delta$CKA & $\Delta$ANC \\
\midrule
\multicolumn{5}{l}{\emph{BLOOM}} \\
BLOOM-560M & $-$0.139 & +0.021 & $-$0.027 & $-$0.004 \\
BLOOM-1b1 & $-$0.163 & $-$0.003 & $-$0.001 & $-$0.007 \\
BLOOM-1b7 & $-$0.001 & $-$0.012 & $-$0.002 & $-$0.014 \\
BLOOM-3B & +0.014 & $-$0.025 & +0.001 & $-$0.019 \\
BLOOM-7b1 & +0.003 & $-$0.015 & $-$0.001 & $-$0.011 \\
\midrule
\multicolumn{5}{l}{\emph{OPT}} \\
OPT-125M & +0.002 & +0.001 & $-$0.000 & $-$0.002 \\
OPT-350M & $-$0.006 & +0.001 & $-$0.021 & $-$0.020 \\
OPT-1.3B & +0.020 & $-$0.020 & $-$0.128 & $-$0.080 \\
OPT-2.7B & +0.014 & $-$0.044 & $-$0.262 & $-$0.125 \\
OPT-6.7B & +0.035 & $-$0.037 & $-$0.005 & $-$0.061 \\
\midrule
\multicolumn{5}{l}{\emph{XGLM}} \\
XGLM-564M & $-$0.001 & $-$0.027 & $-$0.006 & $-$0.183 \\
XGLM-1.7B & $-$0.000 & +0.047 & $-$0.151 & +0.103 \\
XGLM-2.9B & $-$0.001 & +0.031 & $-$0.009 & $-$0.032 \\
XGLM-4.5B & +0.005 & +0.031 & $-$0.049 & +0.037 \\
XGLM-7.5B & $-$0.002 & +0.178 & $-$0.195 & $-$0.026 \\
\midrule
\multicolumn{5}{l}{\emph{Qwen3}} \\
Qwen3-0.6B & $-$0.009 & $-$0.034 & $-$0.000 & +0.004 \\
Qwen3-1.7B & $-$0.006 & $-$0.023 & +0.000 & $-$0.001 \\
Qwen3-4B & +0.000 & $-$0.002 & +0.000 & +0.000 \\
Qwen3-8B & +0.001 & $-$0.017 & +0.017 & $-$0.004 \\
Qwen3-14B & +0.000 & $-$0.008 & +0.001 & $-$0.004 \\
\midrule
\multicolumn{5}{l}{\emph{SmolLM3}} \\
SmolLM3-3B & +0.001 & $-$0.005 & $-$0.001 & $-$0.002 \\
\bottomrule
\end{tabular}
\caption{\textbf{Peak-metric shifts after finetuning on XQuAD.} Columns and conventions as in Table~\ref{tab:app_ft_belebele}.}
\label{tab:app_ft_xquad}
\end{table}

\FloatBarrier
\section{Statistical Analysis}
\label{sec:app_regression}

All correlations in this section use Spearman's $\rho$ between a per-model
peak (or mean) sharing metric and transfer scores over the
five tasks reported in Appendix~\ref{sec:app_xlang_results} ($n = 21$
base models, 5 families).

\subsection{Primary Correlation Analysis}
\label{sec:app_primary_correlation}

Table~\ref{tab:app_stats_primary} reports the headline Spearman $\rho$ between each sharing metric and transfer scores with raw and Benjamini--Hochberg-adjusted $p$-values across the four-metric family. ILO and ANC are significant under both; $\lambda$ is significant but weak; CKA is not.

\begin{table}[!htbp]
\centering
\small
\begin{tabular}{lccc}
\toprule
\textbf{Metric} & $\rho$ & $p_\text{raw}$ & $p_\text{BH}$ \\
\midrule
ILO            & $0.901$ & ${<}.001$ & ${<}.001$ \\
ANC (mean)     & $0.839$ & ${<}.001$ & ${<}.001$ \\
$\lambda$      & $0.525$ & $.015$    & $.020$    \\
CKA (mean)     & $0.136$ & $.556$    & $.556$    \\
\bottomrule
\end{tabular}
\caption{\textbf{Primary Spearman correlations between sharing metrics
and mean cross-lingual transfer} (\(n=21\) base models across 5 families).
$p_\text{BH}$ is the Benjamini--Hochberg-adjusted $p$-value across the
four metric tests.}
\label{tab:app_stats_primary}
\end{table}

\paragraph{Without OPT.} OPT is our English-only baseline and the only family in our set without explicit multilingual pretraining. Table~\ref{tab:app_stats_primary_noopt} repeats the analysis on the remaining $n=16$ multilingual models. ILO remains the strongest ($\rho = 0.78$); ANC drops ($\rho = 0.64$), close to $\lambda$ ($\rho = 0.66$); CKA remains non-significant.

\begin{table}[!htbp]
\centering
\small
\begin{tabular}{lccc}
\toprule
\textbf{Metric} & $\rho$ & $p_\text{raw}$ & $p_\text{BH}$ \\
\midrule
ILO            & $0.777$ & ${<}.001$ & $.002$    \\
$\lambda$      & $0.656$ & $.006$    & $.011$    \\
ANC (mean)     & $0.635$ & $.008$    & $.011$    \\
CKA (mean)     & $0.112$ & $.680$    & $.680$    \\
\bottomrule
\end{tabular}
\caption{\textbf{Primary correlations after excluding OPT}
($n=16$ multilingual models across 4 families: BLOOM, XGLM, Qwen3,
SmolLM3). Removing the English-only contrast reduces every correlation; ILO remains the strongest ($0.78$), with $\lambda$ ($0.66$) and ANC ($0.64$) in a second tier.}
\label{tab:app_stats_primary_noopt}
\end{table}

\FloatBarrier

\subsection{Partial Spearman Correlations}
\label{sec:app_partial_spearman}

Larger models tend to share more and to transfer better, so the headline correlations could in principle reflect model scale rather than sharing. Table~\ref{tab:app_stats_partial} reports the partial Spearman $\rho$ holding $\log_{10}$(parameter count) fixed. ILO and ANC are largely unchanged; $\lambda$ is essentially unchanged; CKA increases modestly but remains weak.

\begin{table}[!htbp]
\centering
\small
\begin{tabular}{lcc}
\toprule
\textbf{Metric} & $\rho$ & $\rho_\text{partial}$ \\
\midrule
ILO            & $0.901$ & $0.845$ \\
ANC (mean)     & $0.839$ & $0.836$ \\
$\lambda$      & $0.525$ & $0.504$ \\
CKA (mean)     & $0.136$ & $0.286$ \\
\bottomrule
\end{tabular}
\caption{\textbf{Partial Spearman correlations controlling for
$\log_{10}$(parameter count)} ($n = 21$ base models). $\rho$ reproduces
the raw Spearman correlation; $\rho_\text{partial}$ is the same correlation
after partialling out model size.}
\label{tab:app_stats_partial}
\end{table}

\FloatBarrier

\subsection{Consistent Layer Aggregation}
\label{sec:app_aggregation}

The results in the main text summarize ILO and \GMMLAMBDA\ at their peak"-sharing layer but CKA and ANC as the mean over layers and language pairs.
Because a max"-of"-layers statistic and a mean"-of"-layers statistic have different sampling behaviour, differences between metrics could in principle reflect the summary rule rather than the metrics themselves.
We therefore compute all transfer correlations under two consistent rules.
For every metric we first form a per"-layer profile by averaging over languages or language pairs within each layer (exactly how the ILO profile is constructed), and then take either the profile maximum (all"-peak) or the profile mean over all layers (all"-mean).

Table~\ref{tab:app_aggregation} reports the result.
The ranking of metrics is identical under both consistent rules and identical to the mixed convention used in the main text: ILO strongest, ANC a clear second tier, \GMMLAMBDA\ moderate, CKA null.
ILO and ANC are stable to within $0.04$ across rules on every statistic.
\GMMLAMBDA\ is the only metric with a notable aggregation sensitivity: under all"-mean its cross"-model correlation weakens ($0.52 \to 0.46$) and its per"-family correlations turn negative in three of four families (OPT $-0.90$, Qwen3 $-0.50$, XGLM $-0.70$), so its already"-fragile within"-family behaviour (\S\ref{sec:app_per_family}) additionally depends on the summary rule.
CKA does not improve when given the same peak treatment as ILO ($\rho = 0.12$).
The comparison in the main text is therefore not an artifact of the mixed summary convention.

\begin{table}[!htbp]
\centering
\small
\setlength{\tabcolsep}{4.5pt}
\begin{tabular}{llcccc}
\toprule
\textbf{Rule} & \textbf{Metric} & $\rho$ & $p_\text{BH}$ & $\rho_\text{partial}$ & $p_\text{perm}^{\text{within-fam}}$ \\
\midrule
all-peak & ILO            & $0.901$ & ${<}.001$ & $0.845$ & $.014$ \\
         & ANC            & $0.840$ & ${<}.001$ & $0.846$ & $.43$  \\
         & \GMMLAMBDA     & $0.525$ & $.020$    & $0.504$ & $.056$ \\
         & CKA            & $0.116$ & $.618$    & $0.332$ & $.87$  \\
\midrule
all-mean & ILO            & $0.882$ & ${<}.001$ & $0.804$ & $.009$ \\
         & ANC            & $0.839$ & ${<}.001$ & $0.836$ & $.39$  \\
         & \GMMLAMBDA     & $0.458$ & $.049$    & $0.560$ & $.96$  \\
         & CKA            & $0.136$ & $.556$    & $0.286$ & $.93$  \\
\bottomrule
\end{tabular}
\caption{\textbf{Transfer correlations under consistent layer"-aggregation
rules} ($n = 21$ base models). Every metric is summarized by the same rule:
peak of the per"-layer profile (all"-peak) or mean over all layers (all"-mean);
profiles average over languages or language pairs within each layer.
$p_\text{BH}$ is Benjamini--Hochberg-adjusted across the four metric tests per
rule; $\rho_\text{partial}$ controls for $\log_{10}$(parameters);
$p_\text{perm}^{\text{within-fam}}$ is the within"-family permutation test of
\S\ref{sec:stats} ($10{,}000$ shuffles). ILO is the only metric that survives
the within"-family permutation under either rule, and the metric ranking is
identical under both.}
\label{tab:app_aggregation}
\end{table}

Table~\ref{tab:app_peak_layers} reports where each metric peaks.
ILO peaks in the middle third of the network in nearly every model (relative depth $0.23$--$0.65$), consistent with the inverted"-U profiles of Figure~\ref{fig:layer_profiles}.
CKA's peak location is unstable across models: it falls on the embedding layer (layer~0) in the smallest OPT and XGLM models but on the second"-to"-last layer in Qwen3"-4B/8B and SmolLM3.
Peak CKA therefore selects qualitatively different computational stages in different models, shared lexical statistics near the embeddings in some and output"-adjacent similarity in others, rather than the mid"-network stage that peak ILO consistently identifies, which is a further indication that CKA does not track the sharing phenomenon the other metrics measure.
For \GMMLAMBDA\ in XGLM, the reported peak at layer~0 is not meaningful: the profile is near"-zero and flat, so no peak exists to locate.

\begin{table}[!htbp]
\centering
\small
\setlength{\tabcolsep}{5pt}
\begin{tabular}{lccccc}
\toprule
\textbf{Model} & $L$ & \textbf{ILO} & \GMMLAMBDA & \textbf{CKA} & \textbf{ANC} \\
\midrule
BLOOM-560m & 23 & 12 & 10 & 12 & 16 \\
BLOOM-1b1 & 23 & 14 & 11 & 7 & 14 \\
BLOOM-1b7 & 23 & 15 & 11 & 11 & 17 \\
BLOOM-3b & 29 & 17 & 18 & 15 & 20 \\
BLOOM-7b1 & 29 & 17 & 16 & 17 & 21 \\
OPT-125m & 11 & 0 & 0 & 0 & 8 \\
OPT-350m & 23 & 0 & 0 & 0 & 21 \\
OPT-1.3b & 23 & 15 & 0 & 3 & 19 \\
OPT-2.7b & 31 & 19 & 3 & 5 & 25 \\
OPT-6.7b & 31 & 19 & 3 & 24 & 25 \\
XGLM-564M & 23 & 11 & 0 & 0 & 3 \\
XGLM-1.7B & 23 & 11 & 0 & 0 & 15 \\
XGLM-2.9B & 47 & 11 & 0 & 6 & 7 \\
XGLM-4.5B & 47 & 19 & 0 & 5 & 21 \\
XGLM-7.5B & 31 & 17 & 0 & 3 & 3 \\
Qwen3-0.6B & 27 & 15 & 8 & 15 & 10 \\
Qwen3-1.7B & 27 & 15 & 12 & 15 & 10 \\
Qwen3-4B & 35 & 11 & 15 & 34 & 11 \\
Qwen3-8B & 35 & 18 & 14 & 34 & 11 \\
Qwen3-14B & 39 & 24 & 15 & 20 & 14 \\
SmolLM3-3B & 35 & 10 & 10 & 34 & 10 \\
\bottomrule
\end{tabular}
\caption{\textbf{Peak"-layer locations per metric} (peak of the per"-layer
profile; $L$ = number of layers). ILO peaks mid"-network in all models.
CKA peaks at the embedding layer in the smallest OPT/XGLM models but at the
second"-to"-last layer in Qwen3"-4B/8B and SmolLM3, so its peak does not
correspond to a comparable computational stage across models. \GMMLAMBDA's
layer"-0 peaks in XGLM reflect a near"-zero flat profile rather than a
genuine maximum.}
\label{tab:app_peak_layers}
\end{table}

\subsection{Per"-Language Correlations}
\label{sec:app_perlanguage}

The correlations in the main text aggregate each metric and the transfer score to one value per model.
To test whether the model"-level association merely averages away per"-language structure, we repeat the analysis at the (model, language) level: 21 base models $\times$ 9 non"-English target languages $=$ 189 points, with per"-language transfer defined as the mean over each task's per"-language scores (Appendix~\ref{sec:app_xlang_results}).

Per"-language metric values use the same profile"-peak rule as Appendix~\ref{sec:app_aggregation}.
For ILO, we take each language's own ILO profile peak; ILO counts neighbours from any other language, so this measures how integrated a language is in the shared space.
CKA and ANC are pairwise, which permits two constructs: alignment of the target language with English, the finetuning source (vs.\ English), and the mean over all nine pairs involving the target (all pairs).
\GMMLAMBDA\ is defined jointly over all languages and has no per"-language form, so it does not participate in this analysis.

We report the median within"-model Spearman $\rho$ (each model contributes one correlation across its nine target languages) with the number of models where it is positive, and a pooled partial Spearman $\rho$ over all 189 points controlling for model"-level transfer, computed once more with English"-target subword overlap as an additional control, which addresses the possibility that per"-language associations are lexical rather than representational.
A permutation test in the within"-family design of \S\ref{sec:stats}, applied at the language level ($10{,}000$ shuffles of per"-language transfer within each model, statistic: the model"-controlled partial $\rho$), confirms the pooled association for every metric variant ($p < .001$), so the table distinguishes the variants by effect size rather than by significance.

\begin{table}[!htbp]
\centering
\small
\setlength{\tabcolsep}{4pt}
\begin{tabular}{lcccc}
\toprule
\textbf{Metric} & $\tilde{\rho}_\text{within}$ & $+$ & $\rho_\text{ctrl}$ & $\rho_\text{ctrl+ov}$ \\
\midrule
ILO                & $0.68$ & 20/21 & $0.73$ & $0.70$ \\
ANC (vs.\ English) & $0.77$ & 20/21 & $0.69$ & $0.67$ \\
ANC (all pairs)    & $0.77$ & 20/21 & $0.49$ & $0.46$ \\
CKA (vs.\ English) & $0.47$ & 18/21 & $0.43$ & $0.36$ \\
CKA (all pairs)    & $0.32$ & 17/21 & $0.18$ & $0.11$ \\
\bottomrule
\end{tabular}
\caption{\textbf{Per"-language metric vs.\ per"-language transfer}
(21 models $\times$ 9 target languages $=$ 189 points).
$\tilde{\rho}_\text{within}$: median within"-model Spearman $\rho$
($n = 9$ languages per model); $+$: models with a positive within"-model
correlation; $\rho_\text{ctrl}$: pooled partial $\rho$ controlling for
model"-level transfer; $\rho_\text{ctrl+ov}$: additionally controlling for
English"-target subword overlap. All variants pass the within"-model
permutation test ($p < .001$; see text).}
\label{tab:app_perlanguage}
\end{table}

Three observations follow from Table~\ref{tab:app_perlanguage}.
First, the model"-level conclusion is not an artifact of aggregation: per"-language ILO correlates positively with per"-language transfer within 20 of the 21 models (the exception, Qwen3"-14B, is a flat $\rho = -0.12$, $p = .77$), with a median within"-model $\rho$ of $0.68$.
Second, subword overlap does not account for the association: controlling for English"-target overlap changes the ILO and ANC (vs.\ English) partial correlations by no more than $0.04$, so the per"-language signal is representational rather than lexical, although the overlap channel exists.
Third, for the pairwise metrics, what predicts a language's transfer is its alignment with English, the finetuning source, rather than its average alignment with the other languages: CKA between the target and English tracks transfer ($\tilde{\rho}_\text{within} = 0.47$), whereas CKA averaged over the target's other pairs is weaker ($0.32$) and near zero once overlap is controlled ($0.11$), and ANC shows the same ordering under model"-level controls ($0.69$ vs.\ $0.49$).
ILO is the exception: its any"-language neighbourhood construct predicts transfer without referencing the source language at all.

\subsection{Permutation Tests}
\label{sec:app_permutation_tests}

Table~\ref{tab:app_stats_permutation} reports permutation $p$-values from two null distributions. $p_\text{perm}^{\text{full}}$ permutes transfer scores freely across all 21 models. $p_\text{perm}^{\text{within-fam}}$ permutes only \emph{within} each family, retaining all between-family contrasts; this asks whether the metric--transfer association exists inside families rather than as a between-family ordering alone. ILO survives both tests (within-family $p = .014$); $\lambda$ is borderline under the within-family permutation ($p=.056$) and ANC and CKA fail it outright.

\begin{table}[!htbp]
\centering
\small
\begin{tabular}{lcccc}
\toprule
\textbf{Metric} & $\rho$ & $p_\text{naive}$ & $p_\text{perm}^{\text{full}}$ & $p_\text{perm}^{\text{within-fam}}$ \\
\midrule
ILO            & $0.901$ & ${<}.001$ & $.0001$ & $.014$  \\
ANC (mean)     & $0.839$ & ${<}.001$ & $.0001$ & $.405$  \\
$\lambda$      & $0.525$ & $.015$    & $.018$  & $.056$  \\
CKA (mean)     & $0.136$ & $.556$    & $.558$  & $.931$  \\
\bottomrule
\end{tabular}
\caption{\textbf{Permutation tests} ($n = 21$, four families with
$\geq 2$ members eligible for within-family permutation). The
within-family permutation shuffles transfer scores only within each
family, retaining all between-family contrasts; it is the most
stringent test of whether the metric--transfer association exists
inside a family rather than as a between-family ordering only.}
\label{tab:app_stats_permutation}
\end{table}

\FloatBarrier

\subsection{Per-Task Heterogeneity}
\label{sec:app_per_task}

Table~\ref{tab:app_stats_per_task} reports per-task Spearman $\rho$ to test whether the headline average masks task-level differences. ILO is significant on all five tasks ($\rho \in [0.73, 0.88]$); ANC is significant on all five, though weakest on SIB-200; $\lambda$ is significant only on Belebele and XQuAD, dropping to non-significant on XNLI, XCSR, and SIB-200; CKA reaches no task.

\begin{table*}[!htbp]
\centering
\small
\setlength{\tabcolsep}{4pt}
\begin{tabular}{lrrrrr}
\toprule
\textbf{Metric} & \textbf{Belebele} & \textbf{XNLI} & \textbf{XCSR} & \textbf{SIB-200} & \textbf{XQuAD} \\
\midrule
ILO            & $0.866^{***}$ & $0.877^{***}$ & $0.836^{***}$ & $0.726^{***}$ & $0.840^{***}$ \\
ANC (mean)     & $0.761^{***}$ & $0.812^{***}$ & $0.735^{***}$ & $0.614^{**\hphantom{*}}$  & $0.878^{***}$ \\
$\lambda$      & $0.668^{**\hphantom{*}}$  & $0.425^{\hphantom{***}}$       & $0.271^{\hphantom{***}}$       & $0.148^{\hphantom{***}}$       & $0.636^{**\hphantom{*}}$  \\
CKA (mean)     & $0.265^{\hphantom{***}}$       & $0.068^{\hphantom{***}}$       & $0.073^{\hphantom{***}}$       & $-0.231^{\hphantom{***}}$      & $0.251^{\hphantom{***}}$       \\
\bottomrule
\end{tabular}
\caption{\textbf{Per-task Spearman $\rho$} between each sharing metric
and per-task cross-lingual transfer ($n = 21$). Significance under
BH-FDR over the 20-test pool: $^{*}: p < .05$, $^{**}: p < .01$,
$^{***}: p < .001$. ILO is the only metric significant across all five
tasks; $\lambda$ drops to non-significant on XNLI, XCSR, and SIB-200.}
\label{tab:app_stats_per_task}
\end{table*}

\subsection{Within-Family Breakdown}
\label{sec:app_per_family}

Per-family Spearman $\rho$ uses $n = 5$ models per family and therefore has low individual-test power, but the sign and direction of the four family-level correlations is informative in itself. Table~\ref{tab:app_stats_per_family} reports per-family $\rho$. ILO is positive in three of four families and flat on XGLM ($-0.10$); with $n = 5$ per family these individual coefficients are individually uninformative, which is why our within-family claims rest on the within-family permutation test ($p = .014$), which ILO alone survives. $\lambda$ flips sign on OPT ($-0.90$): more sharing predicts \emph{less} transfer inside the OPT family. ANC flips sign on XGLM ($-0.70$). CKA is non-positive in 3 of 4 families, yet its cross-model $\rho$ from Table~\ref{tab:app_stats_primary} is positive: the weak positive cross-model CKA--transfer ordering is driven by between-family means, not by any within-family relationship.

\begin{table*}[!htbp]
\centering
\small
\begin{tabular}{lcccc}
\toprule
\textbf{Metric} & \textbf{BLOOM} & \textbf{OPT} & \textbf{Qwen3} & \textbf{XGLM} \\
\midrule
ILO            & $+0.90$ & $+1.00$ & $+1.00$ & $-0.10$ \\
ANC (mean)     & $+0.90$ & $+1.00$ & $+0.70$ & $-0.70$ \\
$\lambda$      & $+0.60$ & $-0.90$ & $+0.70$ & $+0.30$ \\
CKA (mean)     & $0.00$  & $-0.40$ & $-0.50$ & $-0.10$ \\
\bottomrule
\end{tabular}
\caption{\textbf{Within-family Spearman $\rho$ per family}
($n = 5$ per family). SmolLM3 is a singleton and is omitted.
ILO is positive in three of four families and flat on XGLM ($-0.10$).
ANC shows a sign reversal
on XGLM ($-0.70$); $\lambda$ flips sign on OPT ($-0.90$); CKA is
non-positive in 3 of 4 families.}
\label{tab:app_stats_per_family}
\end{table*}

\end{document}